\documentclass[11pt]{article}
\usepackage{graphicx}

\usepackage{amsmath,amssymb,amsthm}
\usepackage{accents}
\usepackage{hyperref}
\usepackage{algorithm}
\usepackage{algorithmic}
\usepackage{multicol}
\usepackage{wrapfig}

\usepackage[dvipsnames]{xcolor}
\usepackage{flushend}

\usepackage{subcaption}
\usepackage[none]{hyphenat}
\usepackage{epstopdf}

\newcommand{\black}{\color{black}}

\newcommand{\overbar}[1]{\mkern 1.5mu\overline{\mkern-1.5mu#1\mkern-1.5mu}\mkern 1.5mu}
 \renewcommand{\aa}{\mathbf{a}}
  \providecommand{\bb}{\mathbf{b}}

  \providecommand{\ee}{\mathbf{e}}

  \providecommand{\vv}{\mathbf{v}}
  
  \providecommand{\xx}{\mathbf{x}}
  \providecommand{\yy}{\mathbf{y}}

  \providecommand{\mD}{\mathbf{D}}

  \providecommand{\cD}{\mathcal{D}}

  \providecommand{\cL}{\mathcal{L}}

  \providecommand{\cO}{\mathcal{O}}
  \providecommand{\cP}{\mathcal{P}}

\def\leref#1{Lemma~\ref{#1}}

\newtheorem{lemma}{Lemma}
\newtheorem{theorem}{Theorem}
\newtheorem{claim}{Claim}[section]

\newcommand{\T}{\scriptscriptstyle T}

\def\leref#1{Lemma~\ref{#1}}

\def\thref#1{Theorem~\ref{#1}}

\def\algref#1{Algorithm~\ref{#1}}
\def\appref#1{Appendix~\ref{#1}}
\def\asref#1{A\ref{#1}}
\def\bydef{\triangleq}

\newcommand{\R}{\mathbb{R}}
\newcommand{\abs}[1]{\left\lvert #1\right\rvert}
\newcommand{\norm}[1]{\left\lVert #1\right\rVert}

\DeclareMathOperator{\E}{\mathbb{E}}
\DeclareMathOperator{\argmin}{arg\,min}
\newcommand{\lin}[1]{\ensuremath \left\langle #1 \right\rangle}

\def\remark{\addtocounter{remark}{1}\def\@currentlabel{\theremark}%
\emph{Remark~\theremark}. } \makeatother
\newcounter{remark}

\newtheorem{assumption}{A}
 \usepackage[]{color-edits}

\begin{document}

\title{FedPD: A Federated Learning Framework with Optimal Rates and Adaptivity to Non-IID Data}

\author{Xinwei Zhang$^\dag$, Mingyi Hong$^\dag$, Sairaj Dhople$^\dag$, Wotao Yin$^{\ddag}$ and Yang Liu$^{\#}$ \thanks{
		$^\dag$ University of Minnesota, email: \{zhan6234,mhong,sdhople\}@umn.edu;
		$^{\ddag}$ University of California, Los Angeles, email: wotaoyin@math.ucla.edu;
		$^{\#}$ Webank, Co. Ltd, email: yangliu@webank.com.}}
\maketitle
\begin{abstract}

	
	Federated Learning (FL) is popular for communication-efficient learning from distributed data. 
	To utilize data at different clients without moving them to the cloud, algorithms such as the Federated Averaging (FedAvg) have adopted a ``computation then aggregation" (CTA) model, in which multiple local updates are performed using local data, before sending the local models to the cloud for aggregation. These algorithms fail to work when facing practical challenges, e.g., the local data being non-identically independent distributed. In this paper, we first characterize the behavior of the FedAvg algorithm, and show that without strong and unrealistic assumptions on the problem structure, it can behave erratically (e.g., diverge to infinity). Aiming at designing FL algorithms that are provably fast and require as few assumptions as possible, we propose a new algorithm design strategy from the primal-dual optimization perspective. Our strategy yields algorithms that  can deal with non-convex objective functions, achieve the best possible optimization and communication complexity (in certain sense), and deal with full-batch and mini-batch local computation models. Importantly, the proposed algorithms are {\it communication efficient}, in that the communication effort can be reduced when the level of heterogeneity among the local data also reduces. To our knowledge, this is the first algorithmic framework for FL that achieves all the above properties.
\end{abstract}
	\section{Introduction}\label{sec:intro}

Federated learning (FL)---a distributed machine learning approach proposed in \cite{konevcny2016federated}---has gained popularity for applications involving learning from  distributed  data. In FL, a cloud server  (the ``server")  can communicate with distributed data sources (the ``agents").  The goal is to train a global model that works well for all the distributed data, but without requiring the agents to reveal too much local information. Since its inception, the broad consensus on FL's implementation appears to involve a generic ``computation then aggregation" (CTA) protocol. This involves the following steps:  S1) the server sends the model $\xx$ to the agents; S2) the agents update their local  models $\xx_i$'s for several iterations based on their local data; S3) the server aggregates $\xx_i$'s to obtain a new global model $\xx$.
It is widely acknowledged that multiple local steps save communication efforts, while only transmitting local models protects data privacy {\cite{li2019federated}}.
 
Even though the FL paradigm has attracted significant research from both academia and industry, and many algorithms such as Federated Averaging (FedAvg) have been proposed, several attributes are not clearly established. In particular, the commonly adopted CTA protocol poses significant theoretical and practical challenges to designing effective FL algorithms. This work attempts to provide a deeper understanding of FL, by raising and resolving a few theoretical questions, as well as by developing an effective algorithmic framework with several desirable features.

\noindent{\bf \underline{Problem Formulation.}} Consider the vanilla FL solving the following problem: 
\begin{align}\label{eq:problem}
    \min_{\xx\in \R^d}f(\xx)\triangleq\frac{1}{N}\sum^{N}_{i=1}f_i(\xx),\;~\mbox{with}\;  f_i(\xx)\triangleq w_i\sum_{\xi_i\in\cD_i}F(\xx;\xi_i),
\end{align}
where $N$ is the number of agents; $\xi_i$ denotes one sample in data set $\cD_i$ stored on the $i$-th agent; $F:\mathbb{R}^{d}\to \mathbb{R}$ is the ``loss function''
for the $i$-th data point; and $w_i>0$ is a ``weight coefficient'' (a common choice is $w_i=1/{\vert\cD_i\vert}$~\cite{li2019federated}). We assume that the loss function takes the same form across different agents, and furthermore, {we denote $M:=\sum_{i=1}^{N}|\cD_i|$ to be the total number of samples}. 
One can also consider a related setting, where each $f_i(\xx)$ represents the expected loss  ~\cite{pmlr-v97-yu19d} 
\begin{equation}\label{eq:subproblem} 
    f_i(\xx)\bydef\E_{\xi_i\in\cP_i}F(\xx;\xi_i),
\end{equation}
where $\cP_i$ denotes the data distribution on the $i$-th agent.
Throughout the paper, we will make the following blanket assumptions for problem \eqref{eq:problem}: 
\begin{assumption}\label{as:smooth}
	Each $f_i(\cdot)$, as well as $f$ in \eqref{eq:problem} is $L$--smooth:
	\small{\begin{align*}
		&\norm{\nabla f_i(\xx)-\nabla f_i(\yy)}\leq L\norm{ \xx-\yy}, \;\;\; \norm{\nabla f(\xx)-\nabla f(\yy)}\leq L\norm {\xx-\yy},~\forall\; \xx,\yy \in \R^d,i=1,\dots,N.
		\end{align*}}
\end{assumption}

\begin{assumption}\label{as:lower-bounded}	The objective of problem \eqref{eq:problem} is lower bounded: $f({\xx})\ge c >-\infty$, $\forall~\xx\in\mathbb{R}^d$.
\end{assumption}
In addition to these standard assumptions, state-of-the-art efforts on analysis of FL algorithms oftentimes invoke a number of more {\it restrictive} assumptions.
\begin{assumption}\label{as:bounded}
	{\rm \textsf{(Bounded Gradient (BG))}} The gradients $\nabla f_i$'s are upper bounded (by a constant $G>0$)
	\begin{align}\label{eq:bb}
	\norm{\nabla f_i(\xx)}^2\leq G^2,\quad \forall~\xx\in \R^d, {\forall~i = 1,\dots,N.}
	\end{align}
\end{assumption}

{\black\begin{assumption}\label{as:homogeneous}
	{\rm \textsf{(I.I.D. Data)}} Either one of the following holds: 
	\begin{align}
	{\E\nabla [f_i (\xx)] =\nabla f(\xx)}, &\quad\forall~\xx\in \R^d, \; \forall~i =1,\dots,N,\label{eq:iid}\\
	{\sum^N_{i=1}\norm{\nabla f_i(\xx)}^2}\leq B^2\norm{\nabla f(\xx)}^2, &\quad\forall\;\xx~\in~\{\xx\in \R^d~\vert\norm{\nabla f(\xx)}^2>\epsilon\}.\label{eq:iid:2}
	\end{align}
	\vspace{-0.5cm}
\end{assumption}}
Let us comment on the above assumptions. 

First, the BG assumption typically does not hold for ~\eqref{eq:problem},
in particular, $f_i(\xx) = \|\mathbf{A}_i \xx -\bb_i\|^2$ (where $\mathbf{A}_i$ and $\bb_i$ are related to data). However, the BG assumption is critical for analyzing FedAvg-type algorithms because it bounds the distance traveled after multiple local iterates. Second,~\eqref{eq:iid} is typically used in FL to characterize homogeneity about local data~\cite{Stich:265641,wang2018cooperative}. 
However, an assumption of this type does not hold for FL applications where the data (such as medical records, keyboard input data) are generated by the individual agents~\cite{konevcny2016federated,pmlr-v97-yu19d,8770530,NIPS2017_7029,bonawitz2019towards,li2019convergence}. A reasonable relaxation to this i.i.d. assumption is the following notion of $\delta$-non-i.i.d.-ness of the data distribution. 
\begin{assumption}\label{as:heterogeneous}
	{\rm \textsf{($\delta$-Non-I.I.D. Data)}} 
	The local functions are called $\delta$-non-i.i.d. if either one of the equivalent conditions below holds:
	\begin{align}\label{eq:non-iid}
	\norm{\nabla f_i (\xx) -\nabla f_j(\xx)}\leq \delta, \;\forall~\xx\in \R^d, \; \forall~i\ne j, \;\mbox{\rm or}\; 
	\|\nabla f_i (\xx) - \nabla f(\xx)\|\le \delta\;\forall~\xx\in \R^d, \; \forall~i.\ 
\end{align}
\end{assumption}
By varying $\delta$ from $0$ to $\infty$, \eqref{eq:non-iid} provides a characterization of data non-i.i.d.-ness. {In Appendix \ref{app:example}, we give a few examples of loss functions with different values of $\delta$.} Note that the second inequality in~\eqref{eq:non-iid} is often used in decentralized optimization to quantify the {\it similarity} of local problems~\cite{Lian17decentralized,wang2018edge}. Third,~\eqref{eq:iid:2} does not hold for many practical problems. To see this, note that this condition is parameterized by $\epsilon$, which is typically the desired optimization accuracy~\cite{sahu2018convergence}. Since $\epsilon$ can be chosen arbitrarily small, \eqref{eq:iid:2} essentially requires that the problem is {\it realizable}, that is, $\|\nabla f(\xx)\|$ approaches zero only when all the local gradients approach zero at $\xx$, that is, when the local data are ``similar". 

{\black Finally,  we mention that our objective is to understand FL algorithm from an optimization perspective. So we say that a solution $\xx$ is an $\epsilon$-stationary solution if the following holds:
\begin{align}\label{eq:optimality}
\|\nabla f(\xx)\|^2\le \epsilon.
\end{align}
We are interested in finding the {\it minimum} system resources required, such as the number of local updates, the number of times local variables are transmitted to the server, and the number of times local samples $F(\xx; \xi_i)$'s are accessed, before computing an $\epsilon$-solution \eqref{eq:optimality}. These quantities are referred to as {\it local computation}, {\it communication} complexity, and {\it sample} complexities, respectively.}

\noindent{\bf \underline{Questions to address}} Despite extensive recent research, the FL framework, and in particular, the CTA  protocol is not yet well understood. Below, we list four questions regarding the CTA protocol. 

\noindent{\rm \textsf{Q1 (local updates)}.} What are the best local update directions for the agents to take so as to achieve the best overall system performance (stability, sample complexity, etc.)? 

\noindent{\rm \textsf{Q2 (global aggregation)}.} Can we use more sophisticated processing in the aggregation step to help improve the system performance (sample or communication complexity)? 

\noindent{\rm \textsf{Q3 (communication efficiency)}.} If multiple local updates are preformed between two aggregation steps, will it reduce the communication overhead? 

\noindent{\rm \textsf{Q4 (assumptions)}.} What is the best performance that the CTA type algorithms can achieve while relying on a minimum set of assumptions about the problem? 

Although these questions are not directly related to data privacy, another important aspect of FL, we argue that answering these fundamental questions can provide much needed understanding on algorithms following the CTA, and thus the FL approach. A few recent works (to be reviewed shortly) have touched upon those questions,  but to our knowledge, none of them has conducted a thorough investigation of the questions listed above. 

\noindent{\bf \underline{Related Works.}} We start with a popular method following the CTA protocol, the FedAvg in Algorithm \ref{alg:FedAvg}, which covers the original FedAvg \cite{konevcny2016federated}, the Local SGD~\cite{Stich:265641}, PR-SGD~\cite{yu2019parallel,pmlr-v97-yu19d} and the RI-SGD~\cite{pmlr-v97-haddadpour19a} among others.

\begin{wrapfigure}{r}{0.47\textwidth}
	\begin{center}\small
\begin{minipage}{1\linewidth}
	\vspace{-1.5cm}
	\begin{algorithm}[H]
		\small
		\begin{algorithmic}
			\STATE {Initialize: $\xx^{0}_i\bydef\xx^{0}, i=1,\dots,N$}\\
			\FOR{$r=0,\dots,T-1$ {\it (stage)}}
			\FOR{$q=0,\dots,Q-1$ {\it (iteration)}}
			\STATE {\black either {\bf Option 1} (Local SGD), for all $i$} \\
			\STATE {\black or {\bf Option 2} (Local GD), for all $i$}
			\ENDFOR
			\STATE {Global averaging:} { $\xx^{r+1} =\frac{1}{N}\sum^N_{i=1}\xx^{r,Q}_i$}
			\STATE { Update agents' $\xx^{r+1,0}_i =\xx^{r+1},\; i=1,\dots,N$}
			\ENDFOR
		\end{algorithmic}
		\caption{FedAvg Algorithm}\label{alg:FedAvg}
	\end{algorithm}
\end{minipage}
	\end{center}
	\vspace{-1.5cm}
\end{wrapfigure}
In FedAvg, $T$ is the total {\it stage} number, $Q$ the number of local updates, $r$  the index of the {\it stage}, $q$ the index of the {\it inner iteration}, and $\eta^{r,q}$'s are the stepsizes. It has two options for local updates:
\vspace{-0.1cm}
\begin{align}
& \mbox{\rm {\bf Option 1}:	 Sample $\xi^{r,q}_{i}$ form $\cD_i$},\nonumber\\ & \mbox{Set}~\xx^{r,q+1}_i\bydef\xx^{r,q}_i-\eta^{r,q}\nabla F(\xx^{r,q}_i;\xi^{r,q}_i).\label{eq:local:sgd}\\
\vspace{-0.1cm}
& \mbox{\rm\bf  Option 2}:	\xx^{r,q+1}_i\bydef\xx^{r,q}_i-\eta^{r,q}\nabla f_i(\xx^{r,q}_i).\label{eq:local:gd}
\end{align}

Many recent works are extensions of FedAvg. The algorithm proposed in \cite{pmlr-v97-yu19d} adds momentum to the algorithm. In \cite{pmlr-v97-haddadpour19a}, the data on the local agents are separated into blocks and shared with other agents. In \cite{khaled2019first} the {\it local GD} version \eqref{eq:local:gd} is studied. In~\cite{wang2018cooperative}, a cooperative-SGD is considered; it includes virtual agents, extra variables, and relaxes the parameter server topology. 

It is pertinent to consider how these algorithms address questions Q1--Q4. For Q1, most FedAvg-type algorithms perform multiple local (stochastic) GD steps to minimize the local loss function. However, we will see shortly that in many cases, successive local GD steps lead to algorithm divergence. For Q2, most algorithms use simple averaging, and there is little discussion on whether other types of (linear) processing will be helpful. For Q3, a number of recent works such as~\cite{pmlr-v97-yu19d,khaled2019first} show that, for non-convex problems, to achieve $\epsilon$-solution~\eqref{eq:optimality}, 
 a total of  $O(1/\epsilon^{3/2})$ aggregation steps are needed. However, it is not clear if this achieves the best communication complexity. As for Q4, the FedAvg-type algorithm typically requires either some variant of the BG assumption, or some i.i.d. assumption, or both; See Table \ref{tab:convergence} rows 1--5 for detailed discussions.

A number of more recent works have improved upon FedAvg  in various aspects. FedProx~\cite{sahu2018convergence} addresses Q1 and Q4 by perturbing the update direction. 
This algorithm does not need the BG, but it still requires the i.i.d. assumption~\eqref{eq:iid:2}. The VRL-SGD proposed in \cite{liang2019variance} addresses Q1 and Q4 by using the variance reduction (VR) technique to update the directions for local agents and achieves $\cO(1/\epsilon)$ communication complexity without {the} i.i.d. assumption.  F-SVRG~\cite{cen2019convergence} is another recent algorithm that uses VR. This algorithm {\it does not} follow the CTA protocol as the agents have to transmit the local gradients, but it does not require \asref{as:bounded} and~\asref{as:homogeneous}.  The PR-SPIDER {\cite{sharma2019parallel}} further improves upon FSVRG by reducing sample complexity (SC) from $\cO(M/\epsilon)$ to $\cO(\sqrt{M}/\epsilon)$ (where $M$ is typically larger than $1/\epsilon$). Although {FSVRG and PR-SPIDER} neither require the BG or the i.i.d. assumptions, they require the agents to transmit local gradients to the server and thus do not follow the CTA protocol. This is undesirable, as it has been shown that local gradient information can leak private data~\cite{zhao2020idlg}. Additionally, questions Q2-Q3 are {\it not} addressed in these works.

\begin{table*}[tb!]
	\small
	\centering
	\caption{\small Convergence rates of FL algorithms, measured by total rounds of communication  (RC), number of local updates (LC), and {number of accessed sample (AS), before reaching $\epsilon$-stationary solution}. DN refers to degree of non-i.i.d, BG refers to bounded gradient, NC is non-convex, $\mu$SC means $\mu$-Strongly Convex. {$p$ is the function of $\cO(\frac{\epsilon}{\delta^2})$ illustrated in Fig. \ref{fig:p_delta}}. $^\star$The i.i.d. assumption of FedProx is described in {\black \eqref{eq:iid:2}}; VRL-SGD needs assumption of bounded variance of the stochastic gradient, { which in our finite sum setting implies the BG.} \vspace{0.2cm}}
	\label{tab:convergence}
	\vspace{-0.3cm}
	\begin{tabular}{l||c|l|l|l|l|l}
		\hline\small
		Algorithm                           &Convexity      &DN         &BG         &RC ($T$)                   & LC ($QT$)             & AS \\
		\hline
		FedAvg~\cite{Stich:265641}          &$\mu$SC        &0          &No         & $\cO\left(1/\epsilon^{1/2}\right)$        & $\cO(1/\epsilon)$     &$\cO(1/\epsilon)$ \\
		FedAvg~\cite{li2019convergence}     &$\mu$SC        &-          &Yes        & $\cO\left(1/\epsilon\right)$              & $\cO(1/\epsilon)$     &$\cO(1/\epsilon)$ \\
		Coop-SGD~\cite{wang2018cooperative} &NC             &0          &No         & $\cO(1/\epsilon)$         & $\cO(1/\epsilon^2)$   &$\cO(1/\epsilon^2)$\\
		MPR-SGD~\cite{pmlr-v97-yu19d}       &NC             &-          &Yes        & $\cO(1/\epsilon^{3/2})$   & $\cO(1/\epsilon^2)$   &$\cO(1/\epsilon^2)$\\
		Local-GD \cite{khaled2019first}     &C              &-          &No         & $\cO(1/\epsilon^{3/2})$   & $\cO(1/\epsilon^2)$   &$\cO(M/\epsilon^2)$ \\
		\hline
		FedProx \cite{sahu2018convergence}  &NC             &-$\star$   &No         & $\cO(1/\epsilon)$         & $\cO(1/\epsilon^2)$   &$\cO(1/\epsilon^2)$\\
		F-SVRG\cite{cen2019convergence}     & NC            &-          &No         & $\cO(1/\epsilon)$         & $\cO(Q/\epsilon)$     &$\cO(M/\epsilon+Q/\epsilon)$\\
		VRL-SGD\cite{liang2019variance}     & NC            &-          &{Yes}$^\star$ & $\cO(1/\epsilon)$& $\cO(1/\epsilon^2)$   &$\cO(1/\epsilon^2)$\\
		\hline
		{\bf Fed-PD-GD}                & NC            & $\delta>0$&No         & { $\cO\left((1-p)/\epsilon \right)$}  & $\cO\left(\log(1/\epsilon)/\epsilon\right)$     &$\cO(M\log(1/\epsilon)/\epsilon)$\\
		{\bf Fed-PD-SGD}               & NC            &$\delta>0$ &No         & {$\cO((1-p)/\epsilon)$}& $\cO(1/\epsilon^2)$   &$\cO(1/\epsilon^2)$\\
		{\bf Fed-PD-VR}                & NC            &-          &No         & $\cO(1/\epsilon)$         & $\cO(Q/\epsilon)$     &$\cO(M+\sqrt{M}/\epsilon)$\\
		\hline
	\end{tabular}
\vspace{-0.5cm}
\end{table*}


\noindent{\bf \underline{Our Main Contributions.}} First, we address Q1-Q4 and provide an in-depth examination of the CTA protocol. We show that algorithms following the CTA protocol that are based on successive local gradient updates, the best possible communication efficiency is $\mathcal{O}(1/\epsilon)$; neither additional local processing nor general linear processing can improve this rate. {\black We then show that the BG and/or i.i.d. data assumption is important for the popular FedAvg to work as intended.}  

Our investigation suggests that the existing FedAvg-based algorithms are (provably) insufficient in dealing with many practical problems, calling for a new design strategy. We then propose a meta-algorithm called Federated Primal-Dual (FedPD), which also follows the CTA protocol and can be implemented in several different forms with desirable properties. In particular, it {\it i)}~can deal with the general non-convex problem, {\it ii)}~achieve the best possible optimization and communication complexity when data is non-i.i.d., {\it iii)}~achieve convergence under  {\black only Assumptions A1-- A2.} 
\begin{wrapfigure}{R}{0.4\textwidth}
	\vspace{-0.4cm}
	\centering
	\includegraphics[width =0.8\linewidth]{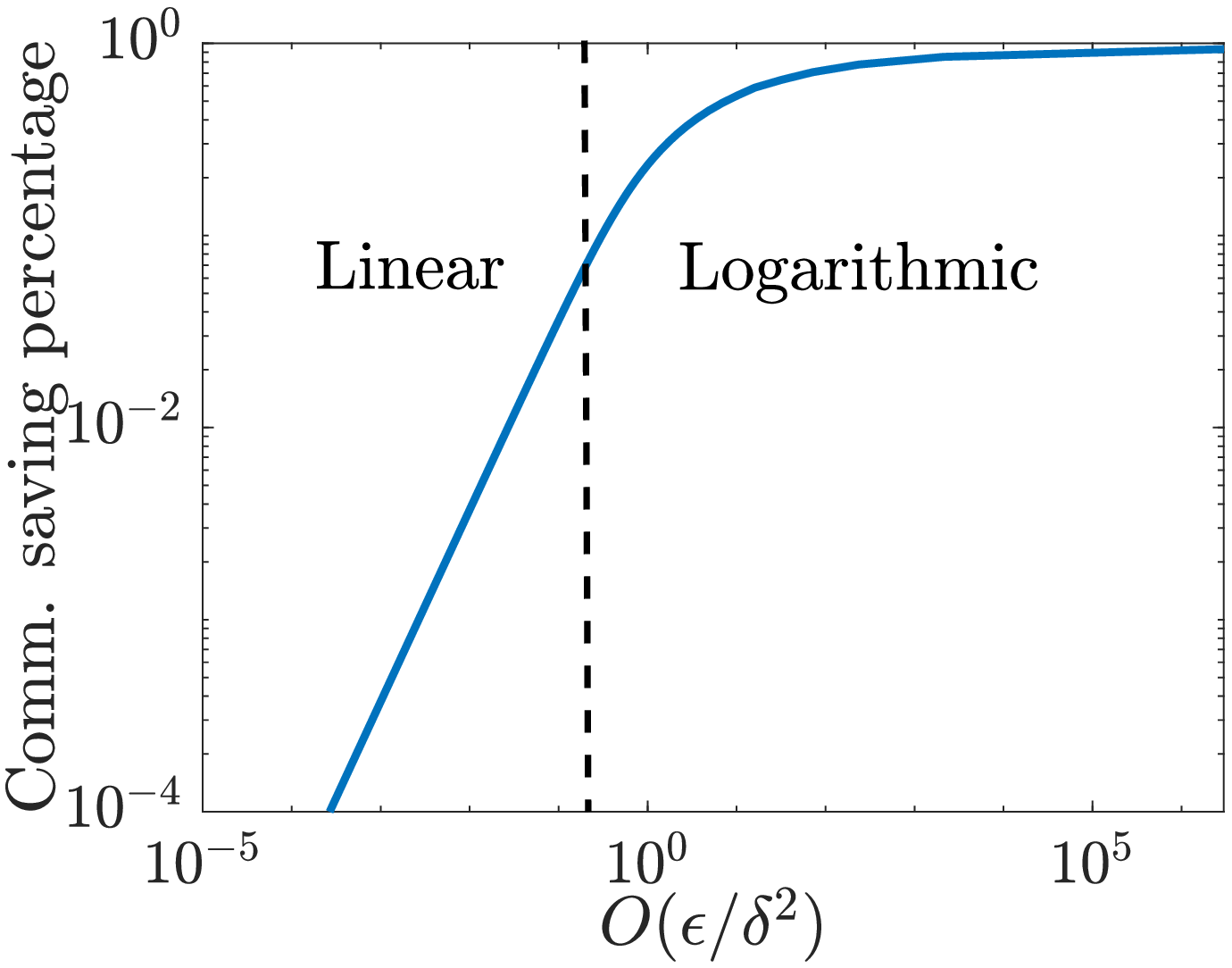}
	\caption{\small Relation of the percentage of comm. savings,  accuracy $\epsilon$,  heterogeneity $\delta$; Fixing $\epsilon$, when the data is heterogeneous (left) the curve is linear, while the data is homogeneous (right) the curve is logarithmic. Details in Sec. \ref{sub:convergence}.}
	\label{fig:p_delta}
	\vspace{-1.4cm}
\end{wrapfigure}

Most importantly, the communication pattern of the proposed algorithm can be adapted to the degree of non-i.i.d.-ness of the local data. {That is, we show that under the $\delta$-non-i.i.d. condition \eqref{eq:non-iid}, communication saving and data heterogeneity interestingly exhibit a linear-logarithmic relationship;  
see Fig.~\ref{fig:p_delta} for an illustration}. To our knowledge, this is the first algorithm for FL that achieves all the above properties. 
\vspace{-0.2cm}

\section{Addressing Open Questions}\label{sec:rethink}
We first address Q2--Q3. Specifically, for  problems satisfying  \asref{as:smooth}--\asref{as:lower-bounded}, does performing multiple local updates or using different ways to combine local models reduce communication complexity? 
We show that such a saving is impossible; there exist problems satisfying \asref{as:smooth}--\asref{as:lower-bounded}, yet no matter what types of linear combinations the server performs, as long as the agents use local gradients to update the model, it takes at least $\mathcal{O}(1/\epsilon)$ communication rounds to achieve an $\epsilon$-stationary solution \eqref{eq:optimality}. Consider the following generic CTA protocol. Let $t$ denote the index for communication rounds. Between two rounds $t-1$ and $t$, each agent performs $Q$ local updates. Denote $x^{t-1,q}_i$ to be the $q$-th local update. Then, $x^{t-1,Q}_i$'s are sent to the server, combined through a (possibly time-varying) function $V^t(\cdot):\mathbb{R}^{Nd}\to \mathbb{R}^{d}$, and sent back. The agents then generate a new iterate, by combining the received message with past gradients using a (possibly time-varying) function $W^t_i(\cdot)$: 
\begin{subequations}\label{eq:span}
		\begin{align}
	x^{t} & ={V^t(\{x^{t-1,Q}_i\}_{i=1}^{N})}, \; x^{t,0}_i = x^t, \;\; \forall~i\in[N], \label{eq:agg}\\
	x^{t,q}_i  & \in \hspace{-0.1cm} {W^t_i\left(\{x_i^{r,k}, \left\{\nabla F(x^{r,q}_i;\xi_i)\right\}_{\xi_i\in D_i}\}_{k\in[q-1], r\in[t]}\right)}, \; \forall\; q\in [Q], \; \forall~i\in [N] \label{eq:comp}.
	\end{align} 
\end{subequations}
We focus on the case where the $V^t(\cdot)$'s and $W^t_i(\cdot)$'s are {\it linear} operators, which implies that $x^{t,q}_i$ can use all past iterates and (sample) gradients for its update. Clearly, \eqref{eq:span} covers both the local-GD and local-SGD versions of FedAvg as special cases.
In the following, we provide an informal statement of the result. The formal statement  and the full proof are given in Appendix \ref{app:lower-bound} and Theorem \ref{thm:1}.
	
	\vspace{-0.1cm}
	\begin{claim}\label{claim:lower-bound}
		{\bf (Informal)} Consider any algorithm $A$ that belongs to the class described in \eqref{eq:span}, with $V^t(\cdot)$ and $W^t_i(\cdot)$'s being {\it linear} and possibly time-varying operators. Then, there exists a non-convex problem instance satisfying Assumptions \ref{as:smooth}--\ref{as:lower-bounded} such that for any $Q>0$, algorithm $A$ takes at least $\mathcal{O}(1/\epsilon)$ communication rounds to reach an $\epsilon$-stationary solution satisfying \eqref{eq:optimality}.
	\end{claim}
\vspace{-0.1cm}
\begin{remark}\label{rmk:iid}
{\black The proof technique is related to those developed from both classical and recent works that characterize lower bounds for first-order methods, in both centralized \cite{Nesterov04,carmon2017lower} and decentralized \cite{scaman2017optimal,sun18optimal} settings. The main technical difference is that our processing model \eqref{eq:span} additionally allows  local processing iterations, and there is a central aggregator. Our goal is {\it not} to establish lower bounds on  the number of (centralized) gradient access, nor to show the optimal graph dependency, but to characterize (potential) communication savings when allowing multiple steps of local processing.} 

In the proof, we construct difficult problem instances in which $f_i$'s are not i.i.d. (more precisely, $\delta$ in assumption \eqref{eq:iid}  grows with  the total number of iterations $T$). Then we show that it is necessary to aggregate (thus communicate) to make any progress. On the other hand, it is obvious that in another extreme case where the data are $0$-non-i.i.d., only $\mathcal{O}(1)$ communication rounds are needed. {\black An {\it open question} is}: when the local data are {\it related} to each other, i.e.,  $\delta$ lies between $0$ and infinity, is it possible to reduce the total communication rounds? This question is addressed below in Sec. \ref{sec:fedpd}. 
\hfill$\blacksquare$
\end{remark}

We {now} address Q1 and Q4. We consider the  FedAvg Algorithm \ref{alg:FedAvg}, {\black and show that BN and/or i.i.d. assumptions are critical for them to perform well.} Our result suggests that, despite its popularity, components in FedAvg, such as the pure local (stochastic) gradient directions and linear aggregation are not compatible with each other. The proof of the results below are given in Appendix~\ref{app:bn}.


\vspace{-0.1cm}
\begin{claim}\label{claim:bn}
	{Fix any constant $\eta>0$, $Q>1$ for Algorithm \ref{alg:FedAvg}. There exists a problem that satisfies \asref{as:smooth} and \asref{as:lower-bounded} but fails to satisfy  \asref{as:bounded} and \asref{as:homogeneous}, on which FedAvg diverges to infinity.} 
\end{claim}
\vspace{-0.1cm}

\begin{remark}\label{rmk:bn}
A recent work \cite{khaled2019first} has shown that FedAvg with {\it constant} stepsize $\eta>0$ can only converge to a neighborhood of the global minimizer for {\it convex} problems. 
Moreover, the error to the global optima is related to $Q$ and the degree of non-i.i.d.-ness as measured by the size of $\sum^N_{i=1}\norm{\nabla f_i(\xx^\star)}^2$ where $\xx^\star$ is the global optimal solution. 
On the other hand, our result indicates that when $f_i$'s are \emph{non-convex}, FedAvg can perform much worse without the BN and the i.i.d. assumption. {\black Even if $Q=2$ and there exists a solution such that $\sum_{i=1}^{N}\|f_i(\hat{\xx})\|^2 =0$, FedAvg (with constant stepsize $\eta$) diverges and the iteration can go to $\infty$.}
\hfill$\blacksquare$
\end{remark}
	
{\black One may think that using a constant stepsize is the culprit for the divergence in Claim \ref{claim:bn}. In fact, we can show that having BG or not can still impact the performance of FedAvg, even when {\it diminishing} stepsize is used. In particular, we show in Appendix \ref{sub:FDGD}, that  FedAvg converges under the BG assumption for any {\it diminishing} stepsize, but without it, the choice of the stepsize can be \emph{significantly restricted.}} 

	\vspace{-0.2cm}
\section{The FedPD Framework}\label{sec:fedpd}
\vspace{-0.2cm}
{\black The previous section reveals a number of properties about FedAvg and, broadly speaking, the CTA protocol. But  {\it why} does FedAvg only work under very restrictive conditions? Is it because the local update directions are not chosen correctly? Is it possible to make it work without any additional assumptions? Can we  reduce communication effort when the local data becomes more i.i.d.?}

In this section, we propose a meta-algorithm called Federated Primal-Dual  (FedPD), which can be specialized into different sub-variants to address the above questions. These algorithms possess a few desirable features: They can achieve the best optimization and communication complexity when data is non-i.i.d. {\black (i.e., achieving the bounds in Claim \ref{claim:lower-bound})}; they only require \asref{as:smooth} --\asref{as:lower-bounded}, while being able to utilize both full or sampled local gradients. Most importantly, the communication pattern of the proposed algorithm can be made adaptive to the degree of data non-i.i.d.-ness across the agents. 

\subsection{The Proposed Algorithm}
Our algorithm is based on the following {\it global consensus} reformulation of the original problem \eqref{eq:problem}: 
\begin{equation}
    \begin{aligned}\label{eq:consensus}
        \min_{\xx_0,\xx_i} &\frac{1}{N}\sum^N_{i=1}f_i(\xx_i), \quad \; \mbox{s.t.}\,\,~\xx_i=\xx_0,\;\forall i\in[N].
    \end{aligned}
\end{equation}
Similar to traditional primal-dual based algorithms ~\cite{BoydADMMsurvey2011}, the idea is that, when relaxing the equality constraints, the resulting problem is {\it separable} across different nodes. However, different from ADMM,  the agents can now perform either a {\it single} (or {\it multiple}) local update(s) between two communication rounds. Importantly, such flexibility makes it possible to adapt the communication frequency to the degree of $\delta$-non-i.i.d.-ness  of the local data.  {In particular, we identify that under $\delta$-non-i.i.d. \eqref{eq:iid}, the fraction $\epsilon/\delta^2$ is the key quantity that determines communication saving; see Fig.~\ref{fig:p_delta}. Intuitively, significant reduction can be achieved when $\delta$ is smaller than $\epsilon$; otherwise, the reduction goes to zero linearly as $\delta$ increases.} 
To  our knowledge, none of the existing ADMM based algorithms, nor FL based algorithms, are able to provably achieve such a reduction.

To present our algorithm, let us define the augmented Lagrangian (AL) function of \eqref{eq:consensus} as
\begin{align*}
    \begin{aligned}
    &{\black \cL(\xx_{0:N},\lambda)}\bydef~\frac{1}{N}\sum^N_{i=1}\cL_i(\xx_0,\xx_i,\lambda_i), \; \cL_i(\xx_i,\xx_0,\lambda_i)\bydef~f_i(\xx_i)+\lin{\lambda_i,\xx_i-\xx_0}+\frac{1}{2\eta}\norm{\xx_i-\xx_0}^2.
    \end{aligned}
\end{align*}
Fixing $\xx_0$, the AL is separable over all local pairs $\{(\xx_i,\lambda_i)\}$. The key technique in the design is to  specify {\it how} each {local} AL $\cL_i(\cdot)$ should be optimized, and {\it when} to perform model aggregation.

Federated primal-dual algorithm (FedPD) captures the main idea of the classical primal-dual based  algorithm while meeting the flexibility need of FL; see \algref{alg:FedPD}. In particular, its update rules share a similar pattern as ADMM, but it does not specify how the local models are updated. Instead, an {\it oracle} $\mbox{\rm Oracle}_i(\cdot)$ is used as a placeholder for local processing, and we will see that careful instantiations of these oracles lead to algorithms with different properties. Importantly, we introduce a {\black critical constant $p\in[0,1)$}, which determines the frequency at which the aggregation and communication steps are {\black skipped}. In \algref{alg:FedPD_v1} and \algref{alg:FedPD_v2}, we provide two useful examples of the local oracles.


\begin{algorithm}[bt!]
	\begin{algorithmic}
		\small
		\STATE {{\bfseries Input:} $\xx^0, \eta, p, T, Q_1, \dots, Q_N$}\\
		\STATE {{\bfseries Initialize:} $\xx^0_0= \xx^0$,}\\
		\FOR{$r=0,\dots,T-1$}
		\FOR{$i=1,\dots,N$ in parallel}
		\STATE \mbox{\sf{Local Updates}}:\\
		\STATE $\xx^{r+1}_i= \mbox{\rm Oracle}_i(\cL_i(\xx^r_i,\xx^r_{0,i},\lambda^r_i),Q_i)$\\
		\STATE $\lambda^{r+1}_i= \lambda^{r}_i+\frac{1}{\eta}(\xx^{r+1}_i-\xx^r_{0,i})$\\
		\STATE $\xx^{r+}_{0,i}= \xx^{r+1}_i+\eta\lambda^{r+1}_i$\\
		\ENDFOR
		{\black\STATE With probability $1-p$:\\
			\STATE \qquad\mbox{\sf{Global Communicate}}:\\
			\STATE \qquad$\xx^{r+1}_{0}= \frac{1}{N}\sum^N_{i=1}\xx^{r+}_{0,i}$\\
			\STATE \qquad$\xx^{r+1}_{0,i}= \xx^{r+1}_0,\;i=1,\dots,N$\\
			\STATE With probability $p$:\\
			\STATE \qquad\mbox{\sf{Local Update}}: $\xx^{r+1}_{0,i}\bydef \xx^{r+}_{0,i}$
		}
		\ENDFOR
	\end{algorithmic}
	\caption{Federated Primal-Dual Algorithm}\label{alg:FedPD}
\end{algorithm}

\begin{algorithm}[bt!]
\begin{algorithmic}
    	\small
		\STATE \bfseries Input: $\cL_i(\xx^r_i,\xx^r_{0,i},\lambda^r_i),Q_i$\\
		\STATE Initialize: $\xx^{r}_{i,0}= \xx^r_i$, \\
		\STATE {\bf Option I} (GD)\\
		\FOR{$q=0,\dots,Q_i-1$}
			\STATE \hspace{-0.4cm} $\xx^{r,q+1}_i= \xx^{r,q}_i - \eta_1 \nabla_{\xx_i}\cL(\xx^{r,q}_i,\xx^r_{0,i},\lambda^r_i)$ 
		\ENDFOR 	\STATE {\bf Option II} (SGD)\\
		\FOR{$q=0,\dots,Q_i-1$}
			\STATE 	 $\xx^{r,q+1}_i\hspace{-0.2cm} = \xx^{r,q}_i - \eta_1 (h_i(\xx^{r,q}_i;\xi^{r,q}_i)+\lambda^r_i +\frac{1}{\eta}(\xx^{r,q}_i-\xx^{r}_{0,i}))$
		\ENDFOR
		\STATE{\bfseries Output:} $\xx^{r+1}_i\bydef \xx^{r,Q_i}_i$
	\end{algorithmic}
	\caption{Oracle Choice I}
	\label{alg:FedPD_v1}
	\end{algorithm}

In \algref{alg:FedPD_v1}, $Q_i$'s are chosen so that the local problems are solved accurately enough to satisfy:
\begin{align}\label{eq:stopping}
\norm{\nabla_{\xx_i}\cL(\xx^{r+1}_i,\xx^{r}_{0,i},\lambda^r_i)}^2\leq \epsilon_1.
\end{align}
 We provide two ways for solving this subproblem by using GD and SGD, but any other solver that achieves \eqref{eq:stopping} can be used. For the SGD version, the stochastic gradient is defined as 
 \begin{align}\label{eq:sample}
 h_i(\xx^{r,q}_i;\xi^{r,q}_i) \bydef \nabla F(\xx^{r,q}_i;\xi^{r,q}_i),~\mbox{with}~\xi^{r,q}_i\sim \mathcal{D}_i,
 \end{align}
 where $\sim$ denotes uniform sampling. {\black Despite the simplicity of the local updates, we will show that using Oracle I makes FedPD adaptive to the non-i.i.d. parameter $\delta$.}
 
In \algref{alg:FedPD_v2}, the oracle applies the variance reduction technique to reduce the sample complexity. The detailed descriptions and the analyses are given in \appref{app:o2} due to the space limitation.

\subsection{Convergence and Complexity Analysis}\label{sub:convergence}

We analyze the convergence of FedPD with Oracle I. The detailed proofs are given in \appref{app:convergence}. The convergence analysis of FedPD with an alternative Oracle is given in \appref{app:o2}

\begin{theorem}\label{th:FedPD_Comm}
	Suppose  \asref{as:smooth} --\asref{as:lower-bounded} hold. Consider FedPD with Oracle I, where $Q_i$ are selected by \eqref{eq:stopping}. 
	
	 	\vspace{-0.2cm} 
    \noindent{\bf Case I)} Suppose \asref{as:heterogeneous} holds with $\delta=\infty$. {\black Set $0<\eta<\frac{\sqrt{5}-1}{4L}$, $p = 0$. Then we have:}
    \begin{align*}
    & \frac{1}{T}\sum^{T}_{r=0}\norm{\nabla f(\xx^r_0)}^2 \leq \frac{C_2}{T} D_0 + C_4\epsilon_1, \;\; \mbox{\rm with}\;\; D_0:= f(\xx^0_0)-f(\xx^\star).
    \end{align*}
    
	\vspace{-0.2cm}
	\noindent{\bf Case II)} Suppose $0<\eta<\frac{\sqrt{5}-1}{4L}$, {\black $0\leq p<1$}, and \asref{as:heterogeneous} holds with a finite $\delta$. Then we have:{\small
    \begin{align}\label{eq:rate:delta}
    \hspace{-0.4cm} \frac{1}{T}\sum^{T}_{r=0}\E\norm{\nabla f(\xx^r_0)}^2 \leq \frac{C_2}{T}D_0 +{\black \frac{\eta (N-1)C_5(1-C_3^{1/(1-p)})^2p(p^2(3+L\eta)^2+4)}{N(1-2L\eta-p(1+L\eta))^2}}(\delta^2+\epsilon_1)+C_4\epsilon_1.
    \end{align}}
  
  Here $C_2, C_4, C_5>0$ are constants independent of $T, \delta, p$; $C_3: = {\black \frac{{p(1+L\eta)+L\eta}}{1-L\eta}}\geq 0$.
\end{theorem}

{\remark{({\bf Communication complexity}) {\black Case I says if one does not skip communication ($p=0$),  then to achieve $\epsilon$-stationarity (i.e., $\norm{\nabla f(\xx^t_0)}^2\leq\epsilon$ for some $r\in(1,T)$), they need to set $T=1/(2C_2 D_0 \epsilon)$, $\epsilon_1=\epsilon/(2C_4)$, and the total communication rounds is $T$. 

In Case II, the second term on the right hand side of \eqref{eq:rate:delta} involves both $\delta$ and $p$, so we can select them appropriately to reduce communication while maintaining the same accuracy. Specifically, we choose $\epsilon_1=\min\{\epsilon/(4 C_4),\delta^2\}$ and $T=1/(2C_2 D_0 \epsilon)$. Then we have
\[C(p) \triangleq \eta \frac{(1-C_3^{1/(1-p)})^2 p(p^2(3+L\eta)^2+4)}{(1-2L\eta-p(1+L\eta))^2} \le \frac{ \epsilon}{3\delta^2}.\]
Then we can achieve the same $\epsilon$-accuracy as Case I. The communication rounds here is $T(1-p)$. 

The relation between the saving $p$ and $\frac{\epsilon}{\delta^2}$ is showed in Table~\ref{tab:saving}. Roughly speaking, $p$ is inversely proportional to degree of non-i.i.d-ness $\delta^2$ when $\delta^2\in (\cO(\epsilon), \infty)$;  further, $p \rightarrow 1$ at a log-rate when $\delta^2 \rightarrow 0$.  Our result also indicates that, when using stage-wise training for neural networks, the algorithm can communicate less at the early stages since they typically have lower accuracy target to enable larger stepsizes~\cite{senior2013empirical}.
\begin{table*}[h!]
	\small
	\centering
	\vspace{-0.3cm}
	\caption{\small The relation between $p$ and $\frac{\epsilon}{\delta^2}$ with fixed $\eta = \frac{\sqrt{5}-1}{8L}$.\vspace{0.2cm}}
	\label{tab:saving}
	\vspace{-0.3cm}
	\begin{tabular}{crllr}
		\hline\small
		Range of $p$ & $C_3$& $C(p)$  &  $p$ as function of $\frac{\epsilon}{\delta^2}$& Relation\\
		\hline
		$[0,\frac{1-2L\eta}{1+L\eta})$&$<1$&$\approx 12\eta p$&$\frac{1}{36\eta}\frac{\epsilon}{\delta^2}$ &{\bf Linear}\\
		$[\frac{1-2L\eta}{1+L\eta},1)$&$\geq 1$&$\approx 14\eta C_3^{2/(1-p)}$&$1 - 2/\log(\frac{1}{42\eta}\frac{\epsilon}{\delta^2})$&{\bf Log}\\
		\hline
	\end{tabular}
\vspace{-0.2cm}
\end{table*}
}}}

{\remark{({\bf Computation complexity}) To achieve $\epsilon$ accuracy, we need both $T=\cO(1/\epsilon)$ and $\epsilon_1=\cO(\epsilon)$. As the local AL is strongly convex with respect to $\xx_i$, optimizing it to $\epsilon$ accuracy requires $\cO(\log(\epsilon))$ iterations for GD and $\cO(1/\epsilon)$ for SGD~\cite{ESGD2016}. So the total number of times that the local gradients (respectively, stochastic gradients) are accessed is given by $\mathcal{\cO(\text{1}/\epsilon \times \log( \text{1} /\epsilon))}$ (respectively, $\mathcal{O}(1/\epsilon^2)$).}}

We conclude this section by noting that the above communication and computation complexity results we have obtained are the best so far among all FL algorithms for non-convex problems satisfying \asref{as:smooth} -- \asref{as:lower-bounded}. Please see the last three rows of Table \ref{tab:convergence} for a summary of the results.

\subsection{Connection with Other Algorithms}
Before we close this section, we discuss the relation of FedPD with a few existing algorithms.

\noindent{\bf The FedProx/FedDANE}
In FedProx~\cite{sahu2018convergence} the agents optimize the following local objective: 
$f_i(\xx_i)+\frac{\rho}{2}\norm{\xx_i-\xx^r_0}^2$. This fails to converge to the global stationary solution. In contrast, FedPD introduces extra local dual variables $\{\lambda_i\}$ that record the gap between the local model $\xx_i$ and the global model $\xx_0$ which help the global convergence.
FedDANE~\cite{li2020feddane} also proposes a way of designing the subproblem by using the global gradient, but this violates the CTA protocol. Compared with these two algorithms, the proposed FedPD has weaker assumptions, and it achieves better sample and/or communication complexity. 

 \noindent{\bf Event Triggering Algorithms.} A number of recent works such as Lazily Aggregated Gradient (LAG) {\cite{NIPS2018_7752}} and COLA~\cite{2019COLA}  have been proposed to occasionally skip message exchanges among the agents to save communication. In LAG, each agent receives the global model every iteration, and decides whether to send some local {\it gradient differences} by checking certain conditions. Since gradient information is transmitted, LAG does not belong to the algorithm class \eqref{eq:span}. When the local problems are {\it unbalanced}, in the sense that the discrepancy between the local Lipschitz gradients $L_i$'s is large, then the agents with smaller $L_i$'s can benefit from {the lazy aggregation}. Meanwhile, instead of measuring whether the {local} problems are balanced, the $\delta$-non-i.i.d. criteria characterizes if local problems are {\it similar} by measuring {the uniform difference between arbitrary pairs of the local problems}. If the data is i.i.d., then the agents benefit equally from the communication reduction.
	\vspace{-0.2cm}
\section{Numerical Experiments}\label{sec:numerical}
\vspace{-0.2cm}

In the first experiment, we show the convergence of the proposed algorithms on synthetic data with FedAvg and FedProx as baselines. We use the non-convex penalized logistic regression~\cite{antoniadis2011penalized} as the loss function. We use two ways to generate the dataset, in the first case (referred to as the ``weakly non-i.i.d" case), the data is generated in an i.i.d. way. In the second case (referred to as the ``strongly non-i.i.d" case), we generate the data using non-i.i.d. distribution. In both cases there are $400$ samples on each agent with total $100$ agents.

We run FedPD with Oracle I (FedPD-SGD and FedPD-GD) and Oracle II (FedPD-VR). For FedPD-SGD, we set $Q=600$, and for FedPD-GD and FedPD-VR we set $Q=8$. For FedPD-GD we set $p=0$ and $p=0.5$, wherein the later case the agents skips half of the communication rounds. For FedPD-VR, we set mini-batch size $B=1$ and gradient computation frequency $I=20$. For comparison, we also run FedAvg with local GD/SGD and FedProx. For FedAvg with GD, $Q=8$, and for FedAvg with SGD, $Q=600$. For FedProx, we solve the local problem using variance reduction for $Q=8$ iterations. The total number of iterations $T$ is set as $600$ for all algorithms.

\begin{figure}[bt!]
    \vspace{-0.3cm}
    \centering
    \begin{subfigure}[t]{0.4\linewidth}
        \centering
        \includegraphics[width=\linewidth]{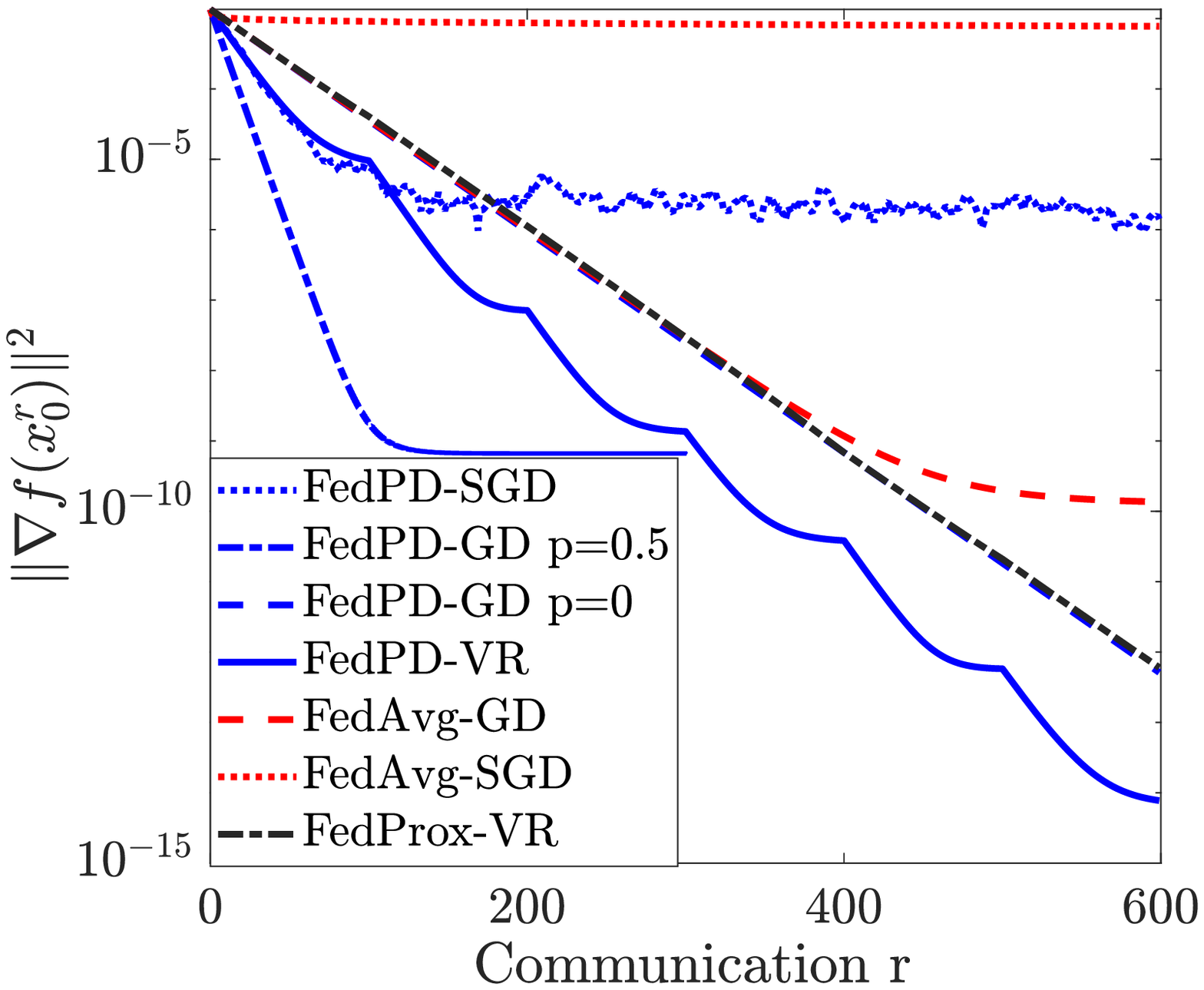}
        \caption{Stationary gap of FedAvg, FedProx and FedPD; weakly non-i.i.d. data.}
    \end{subfigure}
    \hfill
    \begin{subfigure}[t]{0.4\linewidth}
        \centering
        \includegraphics[width=\linewidth]{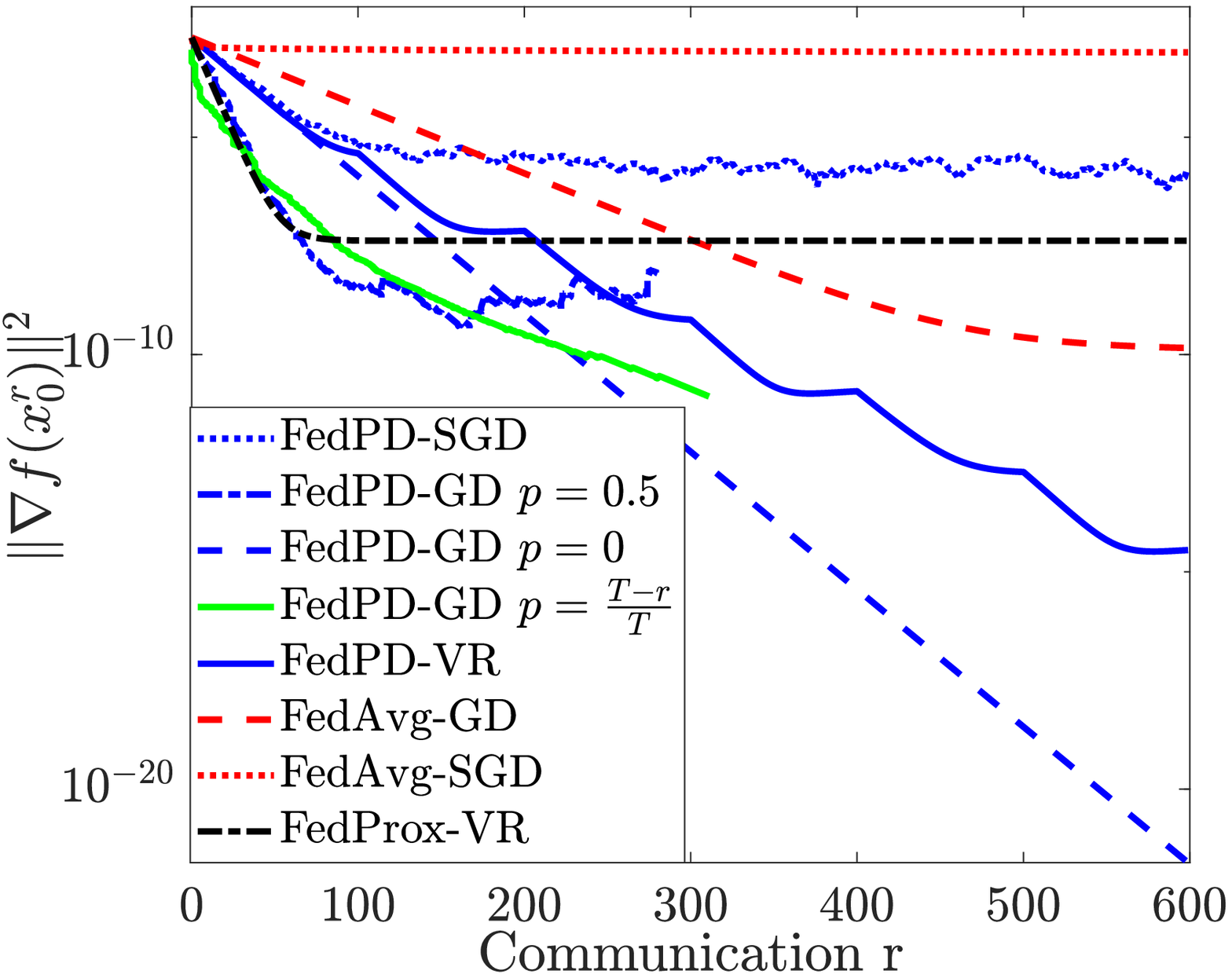}
        \caption{Stationary gap of FedAvg, FedProx and FedPD; strongly non-i.i.d. data.}
    \end{subfigure}
    \caption{\small The convergence result of the algorithms on penalized logistic regression with weakly and strongly non-i.i.d. data with respect to the number of communication rounds.}
    \label{fig:exp_logit_app_0}
    \vspace{-0.2cm}
\end{figure}
Fig.~\ref{fig:exp_logit_app_0} shows the results with respect to the number of communication rounds. In Fig.~\ref{fig:exp_logit_app_0}(a), we compare the convergence of the tested algorithms on weakly non-i.i.d. data set. It is clear that FedProx and  FedPD with $p=0$ (i.e., no communication skipping) are comparable. 
Meanwhile, FedAvg with local GD will not converge to the stationary point with a constant stepsize when local update step $Q>1$. By skipping half of the communication, FedPD with local GD can still achieve a similar error as FedAvg, but using fewer communication rounds. In Fig.~\ref{fig:exp_logit_app_0}(b), we compare the convergence results of different algorithms with the strongly non-i.i.d. data set. We can see that the algorithms using stochastic solvers become less stable compared with the case when the data sets are weakly non-i.i.d. Further, FedPD-VR and FedPD-GD with $p=0$ are able still to converge to the global stationary point while FedProx will achieve a similar error as the FedAvg with local GD.

We included more details on the experimental results and additional experiments in \appref{app:numerical}. 

	\newpage

\vspace{-0.2cm}
\section{Conclusion}
\vspace{-0.2cm}
We study federated learning under the CTA protocol. We explore a number of theoretical properties of this protocol, and design a meta-algorithm called FedPD, which contains various algorithms with desirable properties, such achieving the best communication/computation complexity, as being able to adapt its communication pattern with data heterogeneity. 


	\bibliographystyle{IEEEbib}
	\bibliography{references,ref}
	\appendix
	\onecolumn
\pretolerance=150
\section{Examples of Cost Functions Satisfy \asref{as:heterogeneous}}\label{app:example}
In this part, we provide a commonly used function that satisfies \asref{as:heterogeneous}. 

{\bf{Logistic Regression}} 

Consider the case where the $k^{th}$ sample $\xi_{i,k}$ in data set $\cD_i$ consist of a feature vector $\aa_k$ and a scalar label $b_k$. The feature vector $\aa_k$ has the same length as $\xx$ and $b_k$ is a scalar in $\mathbb{R}$. Then the loss function of a logistic regression problem is expressed as

\begin{equation}
    f_i(\xx) = \frac{1}{\vert\cD_i\vert}\sum_{(\aa_k,b_k)\in\cD_i}\frac{1}{1+\exp(b_k-\aa^T_k \xx)}.
\end{equation}

The gradient of this loss function is

\begin{equation}
    \nabla f_i(\xx) = \frac{1}{\vert\cD_i\vert}\sum_{(\aa_k,b_k)\in\cD_i}\frac{\aa_k\exp(b_k-\aa^T_k \xx)}{(1+\exp(b_k-\aa^T_k \xx))^2}.
\end{equation}

Define the scalar $\frac{\exp(b_k-\aa^T_k \xx)}{(1+\exp(b_k-\aa^T_k \xx))^2}$ as $v(\aa_k,b_k,\xx)$, we have $v(\aa_k,b_k,\xx)\in(0,1),\;\forall x, \aa_k,b_k$. Further stack $v(\aa_k,b_k,\xx)$ as $\vv(\cD_i,\xx)$, that is  $\vv(\cD_i,\xx) = [v(\aa_1,b_1,\xx);\dots,;v(\aa_{\vert\cD_i\vert},b_{\vert\cD_i\vert},\xx)]$.
Further we define $A_i$ as the stacked matrix of all $\aa_k\in \cD_i$ (i.e., $A_i=[\aa_1,\dots,\aa_{\vert\cD_i\vert}]$), then we can express $\nabla f_i(\xx)$ as
\begin{equation}
    \nabla f_i(\xx) = \frac{1}{\vert\cD_i\vert}A_i\vv(\cD_i,\xx).
\end{equation}

The difference between the gradients of $f_i$ and $f_j$ is
\begin{equation}\label{eq:gradient_difference}
    \begin{aligned}
    \norm{\nabla f_i(\xx)-\nabla f_j(\xx)} &= \norm{\frac{1}{\vert\cD_i\vert}A_i\vv(\cD_i,\xx) - \frac{1}{\vert\cD_j\vert}A_j\vv(\cD_j,\xx)}\\
    &\leq \frac{1}{\vert\cD_i\vert}\norm{A_i\vv(\cD_i,\xx)} + \frac{1}{\vert\cD_j\vert}\norm{A_j\vv(\cD_j,\xx)}.
    \end{aligned}
\end{equation}
As $v(\aa,b,\xx)\in(0,1)$, we know $\norm{\vv(\cD_i,\xx)}\leq \norm{[1,\dots,1]} = \sqrt{\vert\cD_i\vert}$, which implies: 
$$\norm{A_i}\geq \frac{\norm{A_i\vv(\cD_i,\xx)}}{\norm{\vv(\cD_i,\xx)}}\geq \frac{\norm{A_i\vv(\cD_i,\xx)}}{\sqrt{\vert\cD_i\vert}}.$$ 
Utilizing the above inequality in \eqref{eq:gradient_difference}, we obtain:
\begin{equation}
    \begin{aligned}
    \norm{\nabla f_i(\xx)-\nabla f_j(\xx)} &\leq \frac{1}{\vert\cD_i\vert}\norm{A_i\vv(\cD_i,\xx)} + \frac{1}{\vert\cD_j\vert}\norm{A_j\vv(\cD_j,\xx)}\\
    &\leq \frac{1}{\sqrt{\vert\cD_i\vert}}\norm{A_i} + \frac{1}{\sqrt{\vert\cD_j\vert}}\norm{A_j}.
    \end{aligned}
\end{equation}

So we can define $\delta = \max_{i,j}\left\{\frac{1}{\sqrt{\vert\cD_i\vert}}\norm{A_i} + \frac{1}{\sqrt{\vert\cD_j\vert}}\norm{A_j}\right\}$ which is a finite constant. Note that the above analysis holds true for any $D_i$ and $\xx$. Note that with finer analysis we can obtain better bounds for $\delta$.

{\bf{Hyperbolic Tangent}}

{\black Similar to logistic regression}, we can also show that \asref{as:heterogeneous} holds for hyperbolic tangent function which is commonly used in neural network models. First, notice that the hyperbolic tangent is a rescaled version of logistic regression:
$$\tanh(b_k-\aa^T_k \xx) = \frac{\exp(b_k-\aa^T_k \xx) - \exp(\aa^T_k \xx - b_k)}{\exp(b_k-\aa^T_k \xx) + \exp(\aa^T_k \xx - b_k)} = \frac{2}{1+\exp(2(b_k-\aa^T_k \xx))}-1,$$

Therefore we have

$$\nabla_{\xx}\tanh(b_k-\aa^T_k \xx) = 4 \nabla_{\xx}\frac{1}{1+\exp(2(b_k-\aa^T_k \xx))}.$$

So, $\delta$ for $\tanh$ is $4$ times that applicable to the logistic regression problem. 
Note that this analysis can further cover a wide range of neural network training problems that uses cross entropy loss and sigmoidal activation functions (e.g. MLP, CNN and RNN). 

{\bf Special Case in Linear Regression}

Consider the linear regression problem 

\[f_i(x) = \frac{1}{2}\norm{A_i\xx+\bb_i}^2, i = 1,\dots,N.\]

We have

\[\nabla f_i(\xx) = A_i^TA_i\xx + A_i^T\bb_i.\]

Then if the feature $A_i$'s satisfy $A_i^TA_i = A_j^TA_j, \forall~i\neq j,$ we have

\[\delta = \max_{i,j}\abs{A_i^T\bb_i -A_j^T\bb_j}.\]


\section{Proof of Claim \ref{claim:lower-bound}}\label{app:lower-bound}
The proof is related to techniques developed in classical and recent works that characterize lower bounds for first-order methods in centralized \cite{Nesterov04,carmon2017lower} and decentralized \cite{scaman2017optimal,sun18optimal} settings. Technically, our computational / communication model is {\it different} compared to the aforementioned works, since we allow arbitrary number of local processing iterations, and we have a central aggregator. The difference here is that our goal is {\it not} to show the lower bounds on  the number of total (centralized) gradient access, nor to show the optimal graph dependency. The main point we would like to make is that  there exist constructions of {\it local} functions $f_i$'s such that {\it no matter} how many times that local first-order processing is performed, without {\it communication} and {\it aggregation}, no significant progress can be made in reducing the stationarity gap of the original problem. 
	
For notational simplicity, we will assume that the full local gradients $\{\nabla f_i(x^k_i)\}$ can be evaluated. Later we will comment on how to extend this result to enable access to the sample gradients $\nabla F(x^k_i; \xi_i)$. In particular, we consider the following slightly simplified model for now: 
\begin{subequations}\label{eq:span:2}
	\begin{align}
	x^{t} & ={V^t(\{x^{t-1,Q}_i\}_{i=1}^{N})}, \; x^{t,0}_i = x^t, \;\; \forall~i\in[N], \label{eq:agg:2}\\
	x^{t,q}_i  & \in {W^t_i\left(\{x_i^{r,k}, \left\{\nabla f_i(x^{r,k}_i)\}\right\}^{k=0:q-1}_{r=0:t}\right)}, q\in [Q],\; \forall~i. \label{eq:comp:2}
	\end{align} 
\end{subequations}

\subsection{{Notation.}}
 In this section, we will call each $t$ a ``stage," and call each local iteration $q$ an ``iteration." We use $x$  to denote the variable located at the server. We use  $x_i$  (and sometimes $x_q$) to denote the local variable at node $i$, and use $x_i[j]$ and $x_i[k]$ to denote its $j$th and $k$th elements, respectively. We use $g_i(\cdot)$ and $f_i(\cdot)$ to denote some functions related to node $i$, and $g(\cdot)$ and $f(\cdot)$ to denote the average functions of $g_i$'s and $f_i$'s, respectively. We use $N$ to denote the total number of nodes.

	\subsection{{Main Constructions.}}
	Suppose there are $N$ distributed nodes in the system, and they can all communicate with the server. 
	To begin, we construct the following two non-convex functions
	\begin{align}\label{eq:h}
	{g}(x):=\frac{1}{N}\sum_{i=1}^{N} g_i(x), \quad f(x):=\frac{1}{N}\sum_{i=1}^{N} f_i(x).
	\end{align}
	Here we have $x\in\mathbb{R}^{T+1}$.  We assume $N$ is constant, and $T$ is the total number of stages (a large number and one that can potentially increase). 
	For notational simplicity, and without loss of generality, we assume that $T\ge N$, and it is divisible by $N$. Let us define the component functions $g_i$'s in \eqref{eq:h} as follows. 	
	\begin{align}\label{eq:construct:h}
	g_i(x) = \Theta(x, 1)+\sum_{j=1}^{{T/N}} \Theta(x, (j-1) N + i+1),
	\end{align}
	{where we have defined the following functions
		\begin{subequations}\label{eq:Theta}
			\begin{align}
			\Theta(x, j) & := \Psi(-x[j-1])\Phi(-x[j]) -\Psi(x[j-1])\Phi(x[j]), \; \forall~j= 2, \cdots, T+1,\nonumber\\
			\Theta(x, 1)&:= -\Psi(1)\Phi(x[1]).
			\end{align}
		\end{subequations}
	}
	Clearly, each $\Theta(x, j)$ is only related to two components in $x$, i.e., $x[j-1]$ and $x[j]$. 

	The component functions $\Psi, \Phi :\mathbb{R}\to \mathbb{R}$ are given as below
	\begin{equation*}
	\begin{aligned}
	\Psi(w)
	&:= \begin{cases}
	0 & w \le 0\\
	1-e^{-w^2}  & w>0,
	\end{cases}\\
	\Phi(w)&: =4 \arctan w+2\pi.
	\end{aligned}
	\end{equation*}
	
	By the above definition, the {average} function becomes:
	\begin{align} \label{eq:hbar}
	{g}(x)&:= \frac{1}{M}\sum_{j=1}^{M} g_i(x)  = \Theta(x, 1)+\sum_{j=2}^{T+1} \Theta(x, j) \\
	&= -\Psi(1)\Phi\left(x[1]\right)+ \sum_{j=2}^{T+1}\left[
	\Psi\left(-x[j-1]\right)\Phi\left(-x[j]\right)-
	\Psi\left(x[j-1]\right)\Phi\left(x[j]\right)
	\right].\nonumber
	\end{align}
	
\begin{figure}[bt!]
		\centering
		\includegraphics[width=0.7\linewidth]{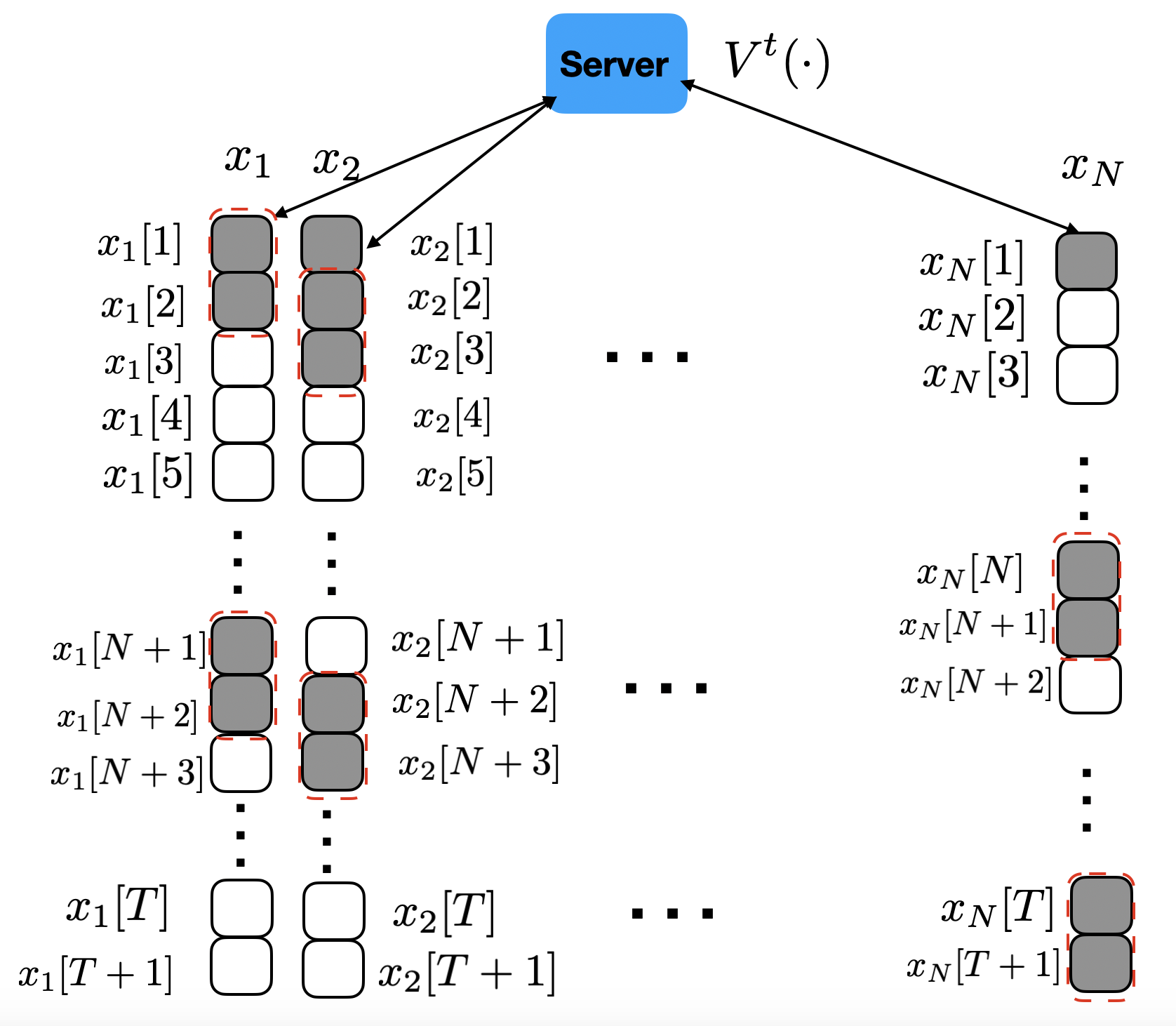}
		\caption{\small The example constructed for proving Claim 2.1. Here each agent has a local length $T+1$ vector $x_i$; each block in the figure represents one dimension of the local vector. If for  agent $i$, its $j$th block is white it means that $f_i$ is not a function of $x_i[j]$, while if $j$th block is shaded means $f_i$ is a function of $x_i[j]$. Each dashed red box contains two variables that are coupled together by a function $\Theta(\cdot)$.}
		\label{fig:construction}
	\end{figure}

	See Fig. \ref{fig:construction} for an illustration of the construction discussed above.

	Further, for a given error constant $\epsilon>0$ and a given the Lipschitz constant $L$,  let us define 
	\begin{equation}\label{eq:f:construct}
	f_i(x) :=\frac{2 \pi \epsilon}{L} {g}_i\left(\frac{x L}{\pi \sqrt{2\epsilon}}\right). 
	\end{equation}
	Therefore, we also have
	\begin{equation}\label{eq:f:construct:2}
	f(x) :=\frac{1}{N} \sum_{i=1}^{N} f_i(x) = \frac{2 \pi \epsilon}{L} {g}\left(\frac{x L}{\pi \sqrt{2\epsilon}}\right). 
	\end{equation}

	\subsection{{Properties.}}
	
	First we present some properties of the component functions $h_i$'s. 
	\begin{lemma}  \label{lem:func_property}
		The functions $\Psi$ and $\Phi$ satisfy the following:  
		\begin{enumerate}
			\item  For all $w \le 0$, $\Psi(w) = 0$, $\Psi'(w) = 0$. 
			\item The following bounds hold for the functions and their first- and second-order derivatives:
			\begin{align*}
			0 \le \Psi(w) < 1,
			~~ 0 \le \Psi'(w) \le \sqrt{\frac{2}{e}},\\
			-\frac{4}{e^{\frac{3}{2}}} \le \Psi''(w) \le 2,	 
			~~ \forall w>0.
			\end{align*}
			\begin{align*}
			~~ 0 < \Phi(w) < 4\pi,
			~~ 0 < \Phi'(w) \le 4,\\
			-\frac{3\sqrt{3}}{2} \le \Phi''(w)  \le \frac{3\sqrt{3}}{2}, 
			~~ \forall w\in \mathbb{R}.
			\end{align*}
			\item  The following key property holds: 
			\begin{align}\label{eq:key:3}
			\Psi(w)\Phi'(v) > 1, \quad \forall~w\ge 1, \; |v|<1. 
			\end{align}	
			\item  The function $h$ is lower bounded as follows:
			$${g_i}(0) - \inf_{x} {g_i}(x) \le {5\pi T /N},$$ $$ {g}(0) - \inf_{x} {g}(x) \le {5\pi T /N}.$$
			\item  The first-order derivative of ${g}$ (respectively, $g_i$) is Lipschitz continuous with constant $\ell = 27\pi$ (respectively, $\ell_i = 27\pi$, $\forall~i$).
		\end{enumerate}
	\end{lemma}
	
	{\bf Proof.} Property 1) is easy to check.  
	
		To prove Property 2), note that following holds for $w>0$:  
		\begin{align}
		\Psi(w) = 1-e^{-w^2},
		~~ \Psi'(w) =  2e^{-w^2}w,
		~~ \Psi''(w) =  2e^{-w^2}-4e^{-w^2}w^2, \; \forall~w>0.
		\end{align}
		Obviously, $\Psi(w)$ is an increasing function over $w>0$, therefore the lower and upper bounds are $\Psi(0) = 0, \Psi(\infty) = 1$; $\Psi'(w)$ is increasing on $[0, \frac{1}{\sqrt{2}}]$ and decreasing on $[\frac{1}{\sqrt{2}}, \infty]$, where $\Psi''(\frac{1}{\sqrt{2}}) = 0$, therefore the lower and upper bounds are $\Psi'(0) = \Psi'(\infty) = 0, \Psi'(\frac{1}{\sqrt{2}}) = \sqrt{\frac{2}{e}}$; $\Psi''(w)$ is decreasing on $(0, \sqrt{\frac{3}{2}}]$ and increasing on $[\sqrt{\frac{3}{2}}, \infty)$ (this can be verified by checking the signs of  $\Psi'''(w) =4e^{-w^2}w(2w^2-3)$ in these intervals). Therefore the lower and upper bounds are $\Psi''(\sqrt{\frac{3}{2}}) = -\frac{4}{e^{\frac{3}{2}}}, \Psi''(0^{+}) = 2$, i.e.,
		\begin{equation*}
		0 \le \Psi(w) < 1,
		~~ 0 \le \Psi'(w) \le \sqrt{\frac{2}{e}},
		~~ -\frac{4}{e^{\frac{3}{2}}} \le \Psi''(w) \le 2,	 
		~~ \forall w>0.
		\end{equation*}
		Further,  for all $w\in\mathbb{R}$, the following holds:
		\begin{align}
		\Phi(w) = 4\arctan w+2 \pi, 
		~~ \Phi'(w) = \frac{4}{w^2+1},
		~~ \Phi''(w) = -\frac{8w}{(w^2+1)^2}.
		\end{align}
		
		Similarly, as above, we can obtain the following bounds:
		\begin{equation*}
		~~ 0 < \Phi(w) < 4\pi,
		~~ 0 < \Phi'(w) \le 4,
		~~ -\frac{3\sqrt{3}}{2} \le \Phi''(w)  \le \frac{3\sqrt{3}}{2}, 
		~~ \forall w\in \mathbb{R}.
		\end{equation*}

		To show Property 3), note that for all $w \ge 1$ and $|v| < 1$, 
		$$\Psi(w)\Phi'(v)> \Psi(1)\Phi'(1) = 2 (1-e^{-1}) > 1$$
		where the first inequality is true because $\Psi(w)$ is strictly increasing and $\Phi'(v)$ is strictly decreasing for all $w>0$ and $v>0$, and that $\Phi'(v) = \Phi'(|v|)$.
	
	Next we show Property 4). Note that $0 \le \Psi(w) < 1$ and $0 < \Phi(w) < 4\pi$. Therefore we have  ${g}(0) = - \Psi(1)\Phi(0) < 0$ and using the construction in \eqref{eq:construct:h} 
	\begin{align}
	\inf_{x} {g_i}(x) &\ge -\Psi(1)\Phi(x[1])- \sum_{j=1}^{{T/N}}  \sup_{w, v} \Psi(w)\Phi(v) \\
	&\ge {-4\pi - 4(T/N)\pi  \ge -5\pi T/N},
	\end{align}
	where the first inequality follows from $\Psi(w)\Phi(v)>0$, the second follows from $\Psi(w)\Phi(v)< 4\pi$, and the last is true because $T/N\ge 1$. 

	Finally, we show Property 5), using the fact that a function is Lipschitz if it is piecewise smooth with bounded derivative. Before proceeding, let us note a few properties of the construction in \eqref{eq:hbar} (also see Fig. \ref{fig:construction}). First, for a given node $q$, its local function $h_q$ is only related to the following $x[j]$'s
	\begin{align*}
	j & = 1+q+\ell \times N \ge 1, \; \ell=0,\cdots, (N-1),\\
	j & = q + \ell \times N \ge 1, \; \ell=0,\cdots, (N-1),
	\end{align*}
	or equivalently
	\begin{align*}
	q & = j-1-\ell \times N \ge 1, \; \ell=0,\cdots, (N-1),\\
	q & = j -\ell \times N \ge 1, \; \ell=0,\cdots, (N-1).
	\end{align*}

	Then the first-order partial derivative of $g_q(y)$  can be expressed below.

	\noindent{\bf Case I)} If $j\ne 1$ we have 
		\begin{equation}  \label{derivative_even}
		\frac{\partial g_q}{\partial x[j]} = \left\{
		\begin{array}{lll}
		&\left( - \Psi\left(-x[j-1]\right)\Phi'\left(-x[j]\right) -\Psi\left(x[j-1]\right)\Phi'\left(x[j]\right) \right), & \\
		& \quad \quad \quad \quad \quad \quad  q = j-1-N(\ell)\ge 1, \; \ell=0,\cdots, \frac{T}{N}-1, j =2, 3, \cdots, T+1\\
		& \left( -\Psi'\left(-x[j]\right)\Phi\left(-x[j+1]\right) -\Psi'\left(x[j]\right)\Phi\left(x[j+1]\right) \right), & \\
		& \quad \quad \quad\quad \quad \quad   q = j-N(\ell)\ge 1, \; \ell=0,\cdots, \frac{T}{N}-1, j =3, 4, \cdots T\\
		& 0 \\
		& \quad \quad \quad\quad \quad \quad  \mbox{otherwise}.
		\end{array}
		\right..
		\end{equation}
	
	\noindent{\bf Case II)} If $j = 1$, we have  
	\begin{equation}  \label{derivative_zero}
	\frac{\partial g_q}{\partial x[1]} = \left\{
	\begin{array}{lll}
	&- \Psi(1)\Phi'(x[1]) + \left( -\Psi'\left(-x[1]\right)\Phi\left(-x[2]\right) -\Psi'\left(x[1]\right)\Phi\left(x[2]\right) \right), &q =1\\
	&- \Psi(1)\Phi'(x[1]),   &q \ne 1 \\
	\end{array}
	\right..
	\end{equation}
	
	From the above derivation, it is clear that for any $j,q$, $\frac{\partial g_q}{\partial x[j]}$ is either  zero or is a piecewise smooth function separated at the non-differentiable point $x[j] = 0$, because the function $\Psi'(\cdot)$ is not differentiable at $0$. 
	
	Further, fix a point $x\in\mathbb{R}^{T+1}$ and a unit vector $v\in\mathbb{R}^{T+1}$ where $\sum_{j=1}^{T+1} v[j]^2 = 1$.  Define 
	$$\ell_{q}(\theta; x,v):= g_q(x+\theta v)$$ 
	to be the directional projection of $g_q$ on to the direction $v$ at point $x$.  We will show that there exists $C>0$ such that $|\ell_{q}{''}(0;x,v)| \le C$ for all $x\ne 0$ (where the second-order derivative is taken with respect to $\theta$). 
	
	First, by noting the fact that each if $x[j]$ appears in $g_q(x)$, then it must also be {\it coupled with} either $x[j+1]$ or $x[j-1]$, but not other $x[k]$'s for $k\ne j-1, j+1$. This means that   $\frac{\partial^2 g_q\left(x\right)}{\partial x[j_1] \partial x[j_2]} =0$, $\forall~j_2\ne \{j_1, j_1+1,j_1-1\}$. 
	Using this fact, we can compute $\ell_{q}{''}(0;x,v)$ as follows:
	\begin{align*}{}
	\ell_{q}^{''}\left(0;x,v\right)
	&=\sum_{j_{1}, j_{2}=1}^{T}
	\frac{\partial^2 g_q\left(x\right)}{\partial x[j_1] \partial x[j_2]} v[j_1] v[i_2]\\
	&=  \sum_{\delta\in\left\{ 0,1, -1\right\}}
	\sum_{j=1}^{T}
	\frac{\partial^2 g_q\left(x\right)}{\partial x[j]\partial x[j+\delta] }
	v[j]v[j+\delta],
	\end{align*}
	where we take $v[0] := 0$ and $v[T+1]:= 0$.

	By using \eqref{derivative_even} and the above facts, the second-order partial derivative of $g_q(x)$ ($\forall x\ne 0$) is given as follows when $j\ne 1$: 
		\begin{equation} \label{eq:1a}
		\frac{\partial^2 g_q}{ \partial x[j]\partial x[j]}  = \left\{
		\begin{array}{lll}
		& \left(  \Psi\left(-x[j-1]\right)\Phi''\left(-x[j]\right) -\Psi\left(x[j-1]\right)\Phi''\left(x[j]\right) \right), &\\
		& \quad \quad \quad \quad 	 \quad \quad \quad \quad \quad   q = j-1-N(\ell)\ge 1, \; \ell=0,\cdots, \frac{T}{N}-1, j =2, 3, \cdots, T +1&\\
		&  \left( \Psi''\left(-x[j]\right)\Phi\left(-x[j+1]\right)  -\Psi''\left(x[j]\right)\Phi\left(x[j+1]\right) \right), &\\
		& \quad \quad \quad \quad \quad \quad  \quad \quad \quad  q = j-N(\ell)\ge 1, \; \ell=0,\cdots, \frac{T}{N}-1, j =3, 4, \cdots, T & \\
		&0,\\
		&\quad \quad \quad\quad \quad \quad \mbox{otherwise}  \\
		\end{array}
		\right.
		\end{equation}
		
		\begin{equation}  \label{eq:1b}
		\frac{\partial^2 g_q}{ \partial x[j]\partial x[j+1]} = \left\{
		\begin{array}{lll}
		& \left( \Psi'\left(-x[j]\right)\Phi'\left(-x[j+1]\right)  -\Psi'\left(x[j]\right)\Phi'\left(x[j+1]\right) \right), \\
		& \quad \quad \quad \quad \quad \quad  \quad \quad \quad  q = j-N(\ell)\ge 1, \; \ell=0,\cdots, \frac{T}{N}-1, j =3, 4, \cdots, T & \\
		&0, \\
		&\quad \quad \quad\quad \quad \quad \mbox{otherwise} 
		\end{array}
		\right.
		\end{equation}
		\begin{align}   \label{eq:1c}
		\frac{\partial^2 g_q}{ \partial x[j]\partial x[j-1]}
		= \left\{
		\begin{array}{lll}
		& \left( \Psi'\left(-x[j-1]\right)\Phi'\left(-x[j]\right)
		-\Psi'\left(x[j-1]\right)\Phi'\left(x[j]\right) \right), \\
		& \quad \quad \quad \quad \quad \quad  \quad \quad \quad  q = j-N(\ell)\ge 1, \; \ell=0,\cdots, \frac{T}{N}-1, j =2, 3, \cdots, T +1& \\
		&0,\\
		&\quad \quad \quad\quad \quad \quad \mbox{otherwise}   \\
		\end{array}
		\right..
		\end{align}
	By applying Lemma \ref{lem:func_property} -- i) [i.e., $\Psi(w)= \Psi'(w)= \Psi''(w)=0$ for $\forall \; w\le0$],  we can obtain that at least one of the terms $\Psi\left(-x[j-1]\right)\Phi''\left(-x[j]\right)$ or $ -\Psi\left(x[j-1]\right)\Phi''\left(x[j]\right)$ is zero. It follows that 
	\begin{align*}
		\Psi\left(-x[j-1]\right)\Phi''\left(-x[j]\right) -\Psi\left(x[j-1]\right)\Phi''\left(x[j]\right)\le \sup_w|\Psi(w)|\sup_v|\Phi''(v)|.
	\end{align*}

	Taking the maximum over equations \eqref{eq:1a} to \eqref{eq:1c} and plug in the above inequalities, we obtain
	\begin{align*}
	\left|\frac{\partial^2 g_q}{ \partial x[j_1]\partial x[j_2]} \right| \nonumber
	&\leq  \max \{\sup_w|\Psi''(w)|\sup_v|\Phi(v)|, \sup_w|\Psi(w)|\sup_v|\Phi''(v)|, \sup_w|\Psi'(w)|\sup_v|\Phi'(v)|\} \\
	&=  \max \left\{8\pi, \frac{3\sqrt{3}}{2}, 4\sqrt{\frac{2}{e}}\right\} < 8\pi, \quad {\forall~j_1\ne 1},
	\end{align*}
	{where the equality comes from Lemma \ref{lem:func_property} -- ii).}
	
	When $j=1$, by using \eqref{derivative_zero},  we have the following:
	\begin{equation*}  
	\frac{\partial^2 g_q(x)}{ \partial x[1]\partial x[1]}  = \left\{
	\begin{array}{lll}
	&- \Psi(1)\Phi''(x[1]) + \left( -\Psi''\left(-x[1]\right)\Phi\left(-x[2]\right) -\Psi''\left(x[1]\right)\Phi\left(x[2]\right) \right), &q =1\\
	&- \Psi(1)\Phi''(x[1]),   &\mbox{otherwise} \\
	\end{array}
	\right.,
	\end{equation*}
	\begin{equation*}  
	\frac{\partial^2 g_q(x)}{ \partial x[1]\partial x[2]}  = \left\{
	\begin{array}{lll}
	&\left( -\Psi'\left(-x[1]\right)\Phi'\left(-x[2]\right) -\Psi'\left(x[1]\right)\Phi'\left(x[2]\right) \right), &q =1\\
	&0,   &\mbox{otherwise} \\
	\end{array}
	\right..
	\end{equation*}
	Again by applying Lemma \ref{lem:func_property}  -- i) and ii), 	
	\begin{align*}
	\left|\frac{\partial^2 g_q(x)}{ \partial {x[1]}\partial x[j_2]} \right| \nonumber
	&\leq  \max \{\sup_w|\Psi(1)\Phi''(w)|+\sup_w|\Psi''(w)|\sup_v|\Phi(v)|, \sup_w|\Psi'(w)|\sup_v|\Phi'(v)|\} \\
	&= \max \left\{ \frac{3\sqrt{3}}{2}(1-e^{-1})+8\pi , 4\sqrt{\frac{2}{e}}\right\}  < 9\pi,\;\forall~j_2.
	\end{align*}
	
	Summarizing the above results, we obtain:
	\begin{align*} 
	|\ell_{q}''\left(0; x,v\right)| &= 
	|\sum_{\delta\in\left\{ 0,1, -1\right\}}
	\sum_{j=1}^{T}
	\frac{\partial^2 g_q\left(y\right)}{\partial x[j]\partial x[j+\delta] }
	v[j]v[j+\delta]|\\
	&\le 9\pi \sum_{\delta\in\left\{ 0,1, -1\right\}}|
	\sum_{j=1}^{T}
	v[j]v[j+\delta]| \\
	& \le 9\pi\left(  | \sum_{j=1}^{T}  v[j]^2| + 2| \sum_{j=1}^{T} v[j]v[j+1]|\right) \\
	&\le 27\pi \sum_{j=1}^{T} | v[j]^2|=  27\pi.
	\end{align*}
	Overall, the first-order derivatives of $h_q$ are Lipsschitz  continuous for any $q$ with constant at most $\ell=27\pi$. \hfill $\blacksquare$

	The following lemma is a simple extension of the previous result. 
	\begin{lemma}\label{lem:property:f}
		We have the following properties for the functions $f$ defined in \eqref{eq:f:construct:2} and \eqref{eq:f:construct}:
		
		\begin{enumerate}
			\item We have $\forall~x\in\mathbb{R}^{T+ 1}$
			\begin{equation*}
			f(0) - \inf_x f(x) \leq \frac{10\pi^2 \epsilon}{L N} T.
			\end{equation*}
			
			\item We have
			\begin{align}
			\norm{\nabla f(x)}
			= \sqrt{2\epsilon}	\norm{\nabla {g}\left(\frac{x L}{\pi \sqrt{2\epsilon}}\right)}, \;  \forall~x\in\mathbb{R}^{T+1}.
			\end{align}
			\item The first-order derivatives of ${f}$ and that for each $f_i, i\in[N]$ are Lipschitz continuous, with the same constant $U>0$. 
		\end{enumerate}
	\end{lemma}
	
	\noindent{\bf Proof.} To show that property 1) is true, note that
	 we have  the following:
	{$$f(0) - \inf_x f(x) = \frac{2 \pi \epsilon}{L} \left({g}(0) - \inf_x {g}(x) \right).$$}
	
	Then by applying Lemma \ref{lem:func_property} we have that for any $T\ge 1$, the following holds
	{$$f(0) - \inf_x f(x) \le \frac{2 \pi \epsilon}{L} \times \frac{5 \pi T}{N}.$$

		Property 2) is true is due to the definition of $f_i$, so that we have:
		$$\nabla f_i(x) = {\sqrt{2\epsilon}}\times \nabla g_i\left(\frac{x L}{\pi \sqrt{2\epsilon}}\right).$$ 
		
		Property 3) is true because the following:
		$$\|\nabla f(z) - \nabla f(y)\| 
		= \sqrt{2\epsilon} \left\| \nabla g\left(\frac{z U}{\pi \sqrt{2\epsilon}}\right) - \nabla g \left(\frac{y U}{\pi \sqrt{2\epsilon}}\right)\right\|
		\le U\|z-y\|$$
		where the last inequality comes from Lemma \ref{lem:func_property}  --  (5). This completes the proof.  \hfill $\blacksquare$}

	Next let us analyze the size of $\nabla {g}$. We have the following result. 
	\begin{lemma} \label{lem:bound_gradient}
		If there exists $k\in[T]$ such that $|x[k]| < 1$,  then
		\begin{equation*}
		\norm{\nabla g(x)} = \norm{  \frac{1}{N}\sum_{i=1}^{N} \nabla g_i(x)}
		\ge \left|   \frac{1}{N}\sum_{i=1}^{N}    \frac{\partial g_i(x)}{\partial x[k]}  \right| > 1/N.
		\end{equation*}
	\end{lemma}
	\noindent{\bf Proof.}
	The first inequality holds for all $k\in[T]$, since {$\frac{1}{N}\sum_{i=1}^{N}    \frac{\partial }{\partial y[k]} g_i(x)$ is one element of $ \frac{1}{N}\sum_{i=1}^{N} \nabla g_i(x)$}. We divide the proof for the second inequality into two cases. 
	
	\noindent{\bf Case 1.} Suppose $|x[j-1]|<1$ for all $2 \le j \le k$.  {Therefore, we have $|x[1]|<1$.}
	Using \eqref{derivative_zero},  we have the following inequalities:
	\begin{equation}  
	\frac{\partial g_i(x)}{\partial x[1]} \stackrel{\rm (i)} \leq -\Psi(1)\Phi'(x[1]) \stackrel{\rm (ii)}<-1, \forall i
	\end{equation}
	{where ${\rm (i)}$ is true because $\Psi'(w), \Phi(w)$ are all non-negative from Lemma \ref{lem:func_property} -(2); ${\rm (ii)}$  is true due to Lemma \ref{lem:func_property}  --  (3). 
		Therefore, we have the following
		\begin{equation*}
		\norm{  \frac{1}{N}\sum_{i=1}^{N} \nabla g_i(x)}
		\ge \left|   \frac{1}{N}\sum_{i=1}^{N}    \frac{\partial }{\partial x[1]} g_i(x) \right| > 1.
		\end{equation*} 
		\noindent{\bf Case 2)} Suppose there exists $2 \le j \le k$  such that $|x[j-1]|\ge 1$.   
		
		We choose $j$ so that $|x[j-1]|\ge 1$ and $|x[j]| < 1$. Therefore, depending on the choices of $(i,j)$ we have three cases: 
			\begin{equation}  
			\frac{\partial g_i(x)}{\partial x[j]} = \left\{
			\begin{array}{lll}
			&\left( - \Psi\left(-x[j-1]\right)\Phi'\left(-x[j]\right) -\Psi\left(x[j-1]\right)\Phi'\left(x[j]\right) \right), \\
			& \quad \quad \quad \quad \quad \quad  i = j-1-N(\ell)\ge 1, \; \ell=0,\cdots, \frac{T}{N}-1, j =2, 3, \cdots, T+1\\
			& \left( -\Psi'\left(-x[j]\right)\Phi\left(-x[j+1]\right) -\Psi'\left(x[j]\right)\Phi\left(x[j+1]\right) \right), \\
			& \quad \quad \quad \quad \quad \quad  i = j-1-N(\ell)\ge 1, \; \ell=0,\cdots, \frac{T}{N}-1, j =3, 4, \cdots, T\\
			& 0 \\
			& \quad \quad \quad \quad \quad \quad \mbox{otherwise}
			\end{array}
			\right..
			\end{equation}

		First, note that $\frac{\partial g_i(x)}{\partial x[j]} \le 0$, for all $i,j$, by checking the definitions of $\Psi(\cdot), \Phi'(\cdot), \Psi'(\cdot), \Phi(\cdot)$.

		Then for $(i,j)$ satisfying the first condition, because $|x[j-1]|\ge 1$ and $|x[j]| < 1$, using Lemma \ref{lem:func_property}  -- (3), and the fact that the negative part is zero for $\Psi$, and $\Phi'$ is even function,  the expression further simplifies to:
		\begin{align}
		-   \Psi(|x[j-1]|)\Phi'\left(|x[j]|\right)  ]\stackrel{\eqref{eq:key:3}}< -1.
		\end{align}
		
		If the second condition holds true,  the expression is obviously non-positive because both $\Psi'$ and $\Phi$ are non-negative. 
		Overall, we have
		$$   \left|   \frac{1}{N}\sum_{i=1}^{N}   \frac{\partial g_i(x)}{\partial x[j]} \right| > \frac{1}{N}.$$ 
		This completes the proof.  \hfill $\blacksquare$}
	
	\begin{lemma}  \label{lem:bound_steps}
		Consider using an algorithm of the form \eqref{eq:span:2} to solve the following problem:
		\begin{align}
		\min_{x\in \mathbb{R}^{T+ 1}} \; g(x)=\frac{1}{N}\sum_{i=1}^{N} g_i(x).
		\end{align}
		Assume  the  initial solution: $x_i= 0, \; \forall~i\in[N]$. Let $\bar{x}=\frac{1}{N}\sum_{i=1}^{N} \alpha_i x_i$ denote some linear combination of local variables, where $\{\alpha_i>0\}$ are the coefficients (possibly time-varying and dependent on $t$).
		Then no matter how many local computation steps \eqref{eq:comp:2} are performed, at least $T$ communication steps \eqref{eq:agg:2} are needed to ensure $\bar{x}[T]\ne 0$.
	\end{lemma}
	
	{\bf Proof.} For a given $j\ge 2$, suppose that  $x_i[j], x_i[j+1],...,x_i[T]=0$, $\forall i$, that is, $\mbox{support}\{x_i\} \subseteq \{1,2,3,...,j-1\}$ for all $i$. Then  $\Psi'\left(x_i[j]\right) = \Psi'\left(-x_i[j]\right) = 0$ for all $i$, and $g_i$ has the following partial derivative  (see \eqref{derivative_even})
	\begin{align}  \label{eq:even:construction}
	& \frac{\partial g_i(x_i)}{\partial x_i[j]} = -    \left( \Psi\left(-x_i[j-1]\right)\Phi'\left(-x_i[j]\right)\right)
	+   \left(\Psi\left(x_i[j-1]\right)\Phi'\left(x_i[j]\right)\right), \\
	& 	\quad \quad \quad \quad \quad \quad  i = j-1-N(\ell)\ge 1, \; \ell=0,\cdots, \frac{T}{N}-1, j =2, 3, \cdots, T+1.
	\end{align}
	
	Clearly, if $x_i[j-1]=0$, then by the definition of $\Psi(\cdot)$, the above partial gradient is also zero. In other words, the above partial gradient is only non-zero if $x_i[j-1]\ne0$.
	
	Recall that we have assumed that the server aggregation is performed using a linear combination $\bar{x}=\frac{1}{N}\sum_{i=1}^{N} \alpha_i x_i$, with the coefficients $\alpha_i$'s possibly depending on the stage $t$ (but such a dependency will be irrelevant for our purpose, as will be see shortly). 
	Therefore, at a given stage $t$, for a given node $i$, when $j\ge 3$, its $j$th element will become {\it nonzero} only if one of the following two cases hold true: 
	\begin{itemize}
		\item If before the aggregation step (i.e., at stage $t-1$), some other node $q$ has $x_q[j]$ being nonzero. 
		\item If $\frac{\partial g_i(x_i)}{\partial x_i[j]}$ is nonzero at stage $t$.
	\end{itemize}
	
	{Now suppose that the initial solution is $x_i[j]=0$ for all $(i,j)$. Then at the first iteration only $\frac{\partial g_i (x_i) }{\partial x_i[1]}$ is non-zero for all $i$, 
		due to the fact that $\frac{\partial g_i (x_i) }{\partial x_i[1]} =  \Psi(1)\Phi'(0) =4(1-e^{-1}) $ for all $i$ from \eqref{derivative_zero}. It is also important to observe that, if all nodes $i\ne 1$ were to perform subsequent local updates \eqref{eq:comp:2}, the local variable $x_j$ will have the same support (i.e., only the first element is non-zero). To see this, suppose $k=2$, then for  $i=2$, we have
		\begin{align}
		\frac{\partial g_i(x_i)}{\partial x_i[2]}  = \left( -\Psi'\left(-x[2]\right)\Phi\left(-x[3]\right) -\Psi'\left(x[2]\right)\Phi\left(x[3]\right) \right) = 0,
		\end{align}
		since $x[2]=0$ implies $\Psi'\left(-x[2]\right)=0$. Similarly reasoning applies when $i=2$, $k\ge 3$. 
		
		If $i\ge 3$, then these local functions are not related to $x_i[2]$, so the partial derivative is also zero. 
		
		Now let us look at node $i=1$. For this node, according to \eqref{eq:even:construction}, we have
		\begin{equation} 
		\frac{\partial g_1(x_1)}{\partial x_1[2]} = -    \left( \Psi\left(-x_1[1]\right)\Phi'\left(-x_1[2]\right)\right)
		+   \left(\Psi\left(x_1[1]\right)\Phi'\left(x_1[2]\right)\right).
		\end{equation}
		Since $x_1[1]$ can be non-zero, then this partial gradient can also be non-zero. Further, with a similar argument as above, we can also confirm that no matter how many local computation steps that node $1$ performs, only the first two elements of $x_1$ can be non-zero. 
		
		So for the first stage $t=1$, we conclude that, no matter how many local computation that the nodes perform  (in the form of the computation step given in \eqref{eq:comp:2}), only $x_1$ can have two non-zero entries, while the rest of the local variables only have one non-zero entries. 
		
		Then suppose that the communication and aggregation step is performed once. It follows that after broadcasting $\bar{x}=\frac{1}{N}\sum_{i=1}^{N} \alpha_i x_i$ to all the nodes, everyone can have two non-zero entries.  
		Then the nodes proceed with local computation, and by the same argument as above, one can show that this time only $x_2$ can have three non-zero entries. 			
		Following the above procedure, it is clear that each aggregation step can advance the non-zero entry of $\bar{x}$ by one, while performing multiple local updates does not advance the non-zero entry. Then we conclude that we need at least $T$ communication steps, and local gradient computation steps, to make $x_i[T]$ possibly non-zero.
		\hfill$\blacksquare$}
	
		\subsection{{Main Result for Claim 2.1.}}
	
		 Below we state and prove a formal version of Claim 2.1 given in the main text. 
 		\begin{theorem}\label{thm:1}
		{\it {Let $\epsilon$ be a positive number}. Let {$x_i^0[j]=0$} for all $i\in[N]$, and all $j=1,\cdots, T+1$. Consider any algorithm obeying the rules given in  \eqref{eq:span}, where the $V^t(\cdot)$ and $W^t_i(\cdot)$'s are {\it linear} operators. Then  regardless of the number of local updates there exists a problem satisfying Assumption \ref{as:smooth} -- \ref{as:lower-bounded}, such that it requires at least the following number of stages $t$ (and equivalently, aggregation and communications rounds in \eqref{eq:agg:2})
		\begin{equation}
		t	\ge   {\frac{\left( f(0) - \inf_x f(x) \right)  LN}{10\pi^2 }\label{eq:lower:bound}
			\epsilon^{-1}}
		\end{equation}
		to achieve the following error
		\begin{align}
		h^*_t= \bigg\|\frac{1}{N}\sum_{i=1}^{N}\nabla f_i(x^t)\bigg\|^2 <\epsilon.
		\end{align} }
 		\end{theorem}
		{\bf Proof of Claim 2.1.} 
	First, let us show that the algorithm obeying the rules given in \eqref{eq:span:2} has the desired property. Note that the difference between two rules is whether the {\it sampled} local gradients are used for the update, or the full local gradients are used. 
	
	By Lemma~\ref{lem:bound_steps} we have $\bar{x}[T] = 0$ for all $t< T$. Then by applying Lemma~\ref{lem:property:f}  --  (2) and  Lemma~\ref{lem:bound_gradient}, we can conclude that the following holds
	\begin{align}
	\norm{\nabla {f}(\bar{x}{[T]})}
	= \sqrt{2\epsilon}	\norm{\nabla {{h}}\left(\frac{\bar{x}{[T]} U}{\pi \sqrt{2\epsilon}}\right)} > \sqrt{2\epsilon}/N,
	\end{align}
	{where the second inequality follows that there exists $k\in [T]$ such that $|\frac{\bar{x}[k] U}{\pi \sqrt{2\epsilon}}|=0< 1$, then we can directly apply Lemma~\ref{lem:bound_gradient}.}

	The third part of  Lemma~\ref{lem:property:f} ensures  that $f_i$'s are $L$-Lipschitz continuous gradient, and the first part shows
	\begin{equation*}
	f(0) - \inf_x f(x)  \leq \frac{10\pi^2 \epsilon}{LN} T,
	\end{equation*}
	Therefore we obtain
	\begin{equation}\label{lowerbound_T}
	T \ge
	{\frac{\left( f(0) - \inf_x f(x) \right)  L N}{10\pi^2}
		\epsilon^{-1}}.
	\end{equation}
	This completes the proof. 
	
	Second, consider the algorithm obeying the rules give in \eqref{eq:span}, in which local {\it sampled} gradients are used. By careful inspection, the result for this case can be trivially extended from the previous case. We only need to consider the following local functions
	\begin{align}
	\hat{f}_i(x) =  \sum_{\xi_i\in D_i} F(x; \xi_i)
	\end{align} 
	where each sampled loss function $F(x; \xi_i)$ is defined as
	\begin{align}
	F(\xx; \xi_i) =  \delta(\xi_i) f_i(x), \; \; \; \mbox{where}~f_i(x)~\mbox{is defined in \eqref{eq:f:construct}}. 
	\end{align}
	where $\delta(\xi_i)$'s satisfy $\delta(\xi_i)>0$ and $\sum_{\xi_i\in D_i}  \delta(\xi_i) = 1$. It is easy to see that, the local sampled gradients have the same dependency on $x$ as their averaged version (by dependency we meant the structure that is depicted in Fig. \ref{fig:construction}). Therefore,  the progression of the non-zero pattern of the average $\bar{x}=\frac{1}{N}\sum_{i=1}^{N}x_i$ is exactly the same as the batch gradient version.  Additionally, since the local function $\hat{f}(x)$ is exactly the same as the previous local function $f(x)$, so other estimates, such as the one that bounds $f(0)-\inf f(x)$, also remain the same.

\section{Proof of Claim \ref{claim:bn}}\label{app:bn}
First let us consider FedAvg with local-GD update \eqref{eq:local:gd}.	We consider the following problem with $N=2$, which satisfies both Assumptions \ref{as:smooth} and \ref{as:lower-bounded}, with $f(\xx) =0, \; \forall~\xx$
	\begin{align}
	f_1(\xx) = \frac{1}{2}\xx^2, \quad f_2(\xx) = -\frac{1}{2}\xx^2. 
	\end{align}
	Each local iteration of the FedAvg is given by
	\begin{align}
	\xx^{r+1}_1 = (1-\eta^{r+1})\xx^r_1, \quad \xx^{r+1}_2 = (1 + \eta^{r+1})\xx^r_2. 
	\end{align}
	For simplicity, let us define $\yy = [\xx_1, \xx_2]^T$, and define the matrix $\mathbf{D} = [1-\eta, 0; 0, 1+\eta]$. Then running $Q$ rounds of the FedAvg algorithm starting with $r= kQ$ for some non-negative integer $k\ge 0$, can be expressed as
	\begin{align}
	\yy^{(k+1)Q} = \mD^{Q-1} \yy^{kQ+1}, \quad \yy^{kQ+1} = \frac{1}{2}\mathbf{1}\mathbf{1}^T\mathbf{D} \yy^{kQ}.
	\end{align}
	Therefore overall we have
	\begin{align}
	\yy^{(k+1)Q} = \frac{1}{2}\mD^{Q-1}\mathbf{1}\mathbf{1}^T\mD \yy^{kQ}.
  \end{align}

  It is easy to show that for any $Q>1$, the eigenvalues of the matrix $\frac{1}{2}\mD^{Q-1}\mathbf{1}\mathbf{1}^T\mD$ are $0$ and $\frac{(1+\eta)^Q+(1-\eta)^Q}{2}>1$.
  
  It follows that the above iteration will diverge for any $Q>1$ starting from any non-zero initial point.	
  
  Moreover, when the sample on one agent are the same (e.g., agent 1 has two samples that both has loss function $x^2$), then using SGD as local update will be identical to the update of GD.

\section{Results showing the role of GB for FedAvg with diminishing stepsizes} \label{sub:FDGD}

This section is used to show the role of \asref{as:bounded} in FedAvg. First we prove that with \asref{as:bounded}, FedAvg can use arbitrary stepsize as long as it diminishes to zero. Next, we show by a simple example that without \asref{as:bounded}, FedAvg will diverge with the same diminishing stepsize choice. 

{\black \begin{claim}{\label{th:FDGD}
		Suppose  \asref{as:smooth}--\asref{as:bounded} hold and the stepsizes satisfy: 1) $\eta^{r,0}=\eta\in (0, 1/L)$ for all $r$;
		2) set $0<\eta^{r,q}\leq \min\{\frac{1}{2(Q-1)L},\frac{\eta}{Q}\},\;\lim_{r\rightarrow\infty}\eta^{r,q}=0,~q \neq 0$. Then the average gradient converges to zero for FedAvg with local-GD update \eqref{eq:local:gd}:
		{\small
			\begin{align*}\label{eq:FDGD_1}
			\frac{1}{T}\sum^{T}_{r=0}\norm{\nabla f(\xx^{r})}^2& \leq\frac{2(f(\xx^{0})-f(\xx^{\star}))}{C_1T} +\frac{2QG^2\eta^2}{C_1T}\sum^{T}_{r=0}\sum^{Q-1}_{q=1}\eta^{r,q},\;\; \mbox{\rm for some $C_1:=\eta(1-L\eta)$.}
			\end{align*}
		}
	}
\end{claim}
\vspace{-0.5cm}
\begin{claim}\label{claim:diverge:bb} Suppose that all the assumptions made in Claim \ref{th:FDGD} hold, except that \asref{as:bounded} does not hold. Then FedAvg with local-GD can diverge for any $Q>1$.
\end{claim}}

Before we prove Claim \ref{th:FDGD}, the following lemma is needed.

\begin{lemma}\label{le:FDGD:descent}
	Under \asref{as:smooth} and \asref{as:bounded}, following the update steps in \algref{alg:FedAvg}, between each outer iterations we have:
	\begin{equation}\label{eq:le:FDGD:descent}
	\small
	\begin{aligned}
	f(\xx^{r+1})-f(\xx^{r})\leq &-(\eta^{r,0}(1-L\eta^{r,0})+\sum^{Q-1}_{q=1}\frac{\eta^{r,q}}{2})\norm{\nabla f(\xx^r)}^2\\
	&-\sum^{Q-1}_{q=1}(\frac{\eta^{r,q}}{2}-2L(Q-1)(\eta^{r,q})^2)\norm{\frac{1}{N}\sum^N_{i=1}\nabla f_i(\xx^{r,q}_i)}^2\\
	&+\frac{QG^2}{2}((\eta^{r,0})^2+\sum^{Q-1}_{q=1}(\eta^{r,q})^2)\sum^{Q-1}_{q=1}\eta^{r,q},
	\end{aligned}
	\end{equation}
	where $r_0+1 \mod Q=0$.
\end{lemma}

\emph{Proof:}{
	By using \asref{as:smooth} we have:
	\begin{equation}\label{eq:le:FDGD:1}
	\small
	\begin{aligned}
	&f(\xx^{r+1})-f(\xx^{r})\\
	&\leq\lin{\nabla f(\xx^{r}),\xx^{r+1}-\xx^{r}}+\frac{L}{2}\norm{\xx^{r+1}-\xx^{r}}^2\\
	&\stackrel{(a)}{=}-\lin{\nabla f(\xx^{r}),\frac{1}{N}\sum^{N}_{i=1}\sum^{Q-1}_{q=0}\eta^{r,q}\nabla f_i(\xx^{r,q}_i)}+\frac{L}{2}\norm{\frac{1}{N}\sum^{N}_{i=1}\sum^{Q-1}_{q=0}\eta^{r,q}\nabla f_i(\xx^{r,q}_i)}^2\\
	&\stackrel{(b)}{\leq}-\sum^{Q-1}_{q=1}\eta^{r,q}\lin{\nabla f(\xx^{r}),\frac{1}{N}\sum^{N}_{i=1}\nabla f_i(\xx^{r,q}_i)}+L(\eta^{r,0})^2\norm{\nabla f(\xx^r)}^2\\
	&\quad+{(Q-1)L}\sum^{Q-1}_{q=1}(\eta^{r,q})^2\norm{\frac{1}{N}\sum^{N}_{i=1}\nabla f_i(\xx^{r,q}_i)}^2\\
	&\stackrel{(c)}{=}-\eta^{r,0}\norm{\nabla f(\xx^{r})}^2-\sum^{Q-1}_{q=1}\eta^{r,q}\lin{\nabla f(\xx^{r}),\frac{1}{N}\sum^{N}_{i=1}\nabla f_i(\xx^{r,q}_i)}\\
	&\quad+L(\eta^{r,0})^2\norm{\nabla f(\xx^{r})}^2+{(Q-1)L}\sum^{Q-1}_{q=1}(\eta^{r,q})^2\norm{\frac{1}{N}\sum^{N}_{i=1}\nabla f_i(\xx^{r,q}_i)}^2,
	\end{aligned}
	\end{equation}
	where $(a)$ comes from the update rule in \algref{alg:FedAvg}, in $(b)$ we use Jensen's inequality and {notice $\xx^{r,0}_i=\xx^{r}$ so in $(c)$ we extract the terms with index $(r,0)$ from the inner product.} 
	
	Note that for any vector $a,b$ of the same length, the equality $2\lin{a,b}=\norm{a}^2+\norm{b}^2-\norm{a-b}^2,$ holds, we have
	\begin{equation}\label{eq:le:FDGD:2}
	\small
	\begin{aligned}
	&-\eta^{r,q}\lin{\nabla f(\xx^{r}),\frac{1}{N}\sum^{N}_{i=1}\nabla f_i(\xx^{r,q}_i)}+{(Q-1)L}(\eta^{r,q})^2\norm{\frac{1}{N}\sum^{N}_{i=1}\nabla f_i(\xx^{r,q}_i)}^2\\
	&= -\frac{\eta^{r,q}}{2}\norm{\nabla f(\xx^{r})}^2-\frac{\eta^{r,q}}{2}\norm{\frac{1}{N}\sum^{N}_{i=1}\nabla f_i(\xx^{r,q}_i)}^2+\frac{\eta^{r,q}}{2}\norm{\nabla f(\xx^{r})-\frac{1}{N}\sum^{N}_{i=1}\nabla f_i(\xx^{r,q}_i)}^2\\
	&\quad +{(Q-1)L}(\eta^{r,q})^2\norm{\frac{1}{N}\sum^{N}_{i=1}\nabla f_i(\xx^{r,q}_i)}^2\\
	&\stackrel{(a)}{\leq}-\frac{\eta^{r,q}}{2}\norm{\nabla f(\xx^{r})}^2+\frac{\eta^{r,q}}{2N}\sum^{N}_{i=1}\norm{\nabla f_i(\xx^{r})-\nabla f_i(\xx^{r,q}_i)}^2\\
	&\qquad-\frac{\eta^{r,q}}{2}((1-2(Q-1)L\eta^{r,q}))\norm{\frac{1}{N}\sum^{N}_{i=1}\nabla f_i(\xx^{r,q}_i)}^2\\
	&\stackrel{(b)}{\leq}-\frac{\eta^{r,q}}{2}\norm{\nabla f(\xx^{r})}^2+\frac{L^2\eta^{r,q}}{2N}\sum^{N}_{i=1}\norm{\xx^{r}-\xx^{r,q}_i}^2-\frac{\eta^{r,q}}{2}((1-2(Q-1)L\eta^{r,q}))\norm{\frac{1}{N}\sum^{N}_{i=1}\nabla f_i(\xx^{r,q}_i)}^2,
	\end{aligned}
	\end{equation}
	where we use Jensen's inequality in $(a)$ and \asref{as:smooth} in $(b)$.
	
	Further note that
	\begin{equation}\label{eq:le:FDGD:3}
	\begin{aligned}
	\norm{\xx^{r}-\xx^{r,q}_i}^2 & =\norm{\xx^{r}-\xx^{r}+\sum^{q-1}_{\tau=0}\eta^{r,\tau}\nabla f_i(\xx^{r,\tau}_i)}^2\\
	&=\norm{\sum^{q-1}_{\tau=0}\eta^{r,\tau}\nabla f_i(\xx^{r,\tau}_i)}^2\\
	&\stackrel{(a)}{\leq} 2(q-1)\sum^{q-1}_{\tau=1}(\eta^{r,\tau})^2\norm{\nabla f_i(\xx^{r,\tau}_i)}^2+ 2(\eta^{r,0})^2\norm{\nabla f_i(\xx^{r,0}_i)}^2\\
	&\stackrel{(b)}{\leq} 2\left((q-1)\sum^{q-1}_{\tau=1}(\eta^{r,\tau})^2+(\eta^{r,0})^2\right) G^2.
	\end{aligned}
	\end{equation}
	{The first equality comes from the update rule of $\xx^{r,q}_i$, which basically performs $q$ steps of updates on $\xx^r$};  $(a)$ comes from Jensen's inequality; in $(b)$ we use \asref{as:bounded}. 
	
	Substitute \eqref{eq:le:FDGD:3} to \eqref{eq:le:FDGD:2} and then to \eqref{eq:le:FDGD:1}, rearrange the terms we obtain \eqref{eq:le:FDGD:descent}, which ends the proof of the lemma. \hfill $\blacksquare$
}

\subsection{Proof of Claim \ref{th:FDGD}}
Next we prove Claim \ref{th:FDGD}

\emph{Proof:}
{
	By choosing $\eta^{r,0}=\eta_1=\in (0,1/L)$ as constant and $\eta^{r,q}\leq 1/(2QL)\;,\forall~q \neq 0$ then applying \leref{le:FDGD:descent} we have 
	\begin{equation}
	\begin{aligned}
	f(\xx^{r+1})&-f(\xx^{r})\leq -(C_1+\sum^{Q-1}_{q=1}\frac{\eta^{r,q}}{2})\norm{\nabla f(\xx^{r})}^2\\
	&+\frac{QG^2}{2}((\eta_1)^2+\sum^{Q-1}_{q=1}(\eta^{r,q})^2)\sum^{Q-1}_{q=1}\eta^{r,q},
	\end{aligned}
	\end{equation}
	where $C_1=\eta_1(1-L\eta_1)>0$.
	Using telescope sum from $r=0$ to $r=T-1$ we have
	\begin{equation}\label{eq:th:FDGD:1}
	\begin{aligned}
	f(\xx^{T})-f(\xx^{0})&\leq -\sum^{T-1}_{r=0}(C_1+\sum^{Q-1}_{q=1}\frac{\eta^{r,q}}{2})\norm{\nabla f(\xx^{r})}^2\\
	&+\frac{QG^2}{2}\sum^{T-1}_{r=0}((\eta_1)^2+\sum^{Q-1}_{q=1}(\eta^{r,q})^2)\sum^{Q-1}_{q=1}\eta^{r,q}.
	\end{aligned}
	\end{equation}
	Rearrange the terms and multiply both side by $2/(TC_1)$, then we have 
	\begin{equation}\label{eq:th:FDGD:2}
	\begin{aligned}
	(\frac{1}{T}+\frac{\sum^{T-1}_{r=0}\sum^{Q-1}_{q=1}\eta^{r,q}}{TC_1})\sum^{T}_{r=0}\norm{\nabla f(\xx^{r})}^2&\leq\frac{2(f(\xx^{0})-f(\xx^{\star}))}{C_1T}\\
	&\quad+\frac{QG^2}{C_1T}\sum^{T-1}_{r=0}((\eta_1)^2+\sum^{Q-1}_{q=1}(\eta^{r,q})^2)\sum^{Q-1}_{q=1}\eta^{r,q}.
	\end{aligned}
	\end{equation}
	Choose $\eta^{r,q}\leq\eta_1/Q$, then $(\eta_1)^2+\sum^{Q-1}_{q=1}(\eta^{r,q})^2\leq 2(\eta_1)^2$. Choose $\{\eta^{r,q}\}$ as a sequence that diminishes to $0$, then for all $q \neq 0$, as $T\rightarrow \infty$, $\frac{2\eta_1Q^2G^2}{C_1}\frac{1}{QT}\sum^{T-1}_{r=0}\sum^{Q-1}_{q=1}\eta^{r,q}\rightarrow0$. Therefore the right hand side converges to 0, Claim \ref{th:FDGD} is proved.
}

\section{Proof of Claim \ref{claim:diverge:bb}}\label{app:diminish}
\noindent{\bf Proof.} We consider the following problem with $N=2$, which satisfies both Assumptions \ref{as:smooth} and \ref{as:lower-bounded}, with $f(\xx) =0, \; \forall~\xx$
	\begin{align}
	f_1(\xx) = \xx^2, \quad f_2(\xx) = -\xx^2. 
	\end{align}
	Each local iteration of the FedAvg is given by
	\begin{align}
	\xx^{r+1}_1 = (1-\eta^r)\xx^r_1, \quad \xx^{r+1}_2 = (1 + \eta^r)\xx^r_2. 
	\end{align}
	For simplicity, let us define $\yy = [\xx_1, \xx_2]^T$, and define the matrix $\mathbf{D}_r = [1-\eta^r, 0; 0, 1+\eta^r]$. Then running $Q$ rounds of the FedAvg algorithm starting with $r= kQ$ for some non-negative integer $k\ge 0$, can be expressed as
	\begin{align}
	\yy^{(k+1)Q} = \prod_{r= kQ+1}^{(k+1)Q-1}\mathbf{D}_r \yy^{kQ+1}, \quad \yy^{kQ+1} = \frac{1}{2}\mathbf{1}\mathbf{1}^T\mathbf{D}_{kQ} \yy^{kQ}.
	\end{align}
	Therefore overall we have
	\begin{align}
	\yy^{(k+1)Q} = \frac{1}{2}\prod_{r= kQ+1}^{(k+1)Q-1}\mathbf{D}_r \mathbf{1}\mathbf{1}^T\mathbf{D}_{kQ} \yy^{kQ}.
  \end{align}
  
  In particular, we pick $\eta^r=\frac{1}{\sqrt{r}}$ when $r \neq kQ+1$ and $\eta^{kQ+1}=1/2$. Then for $Q>1$, it is easy to compute the eigenvalues of the  matrix $ \frac{1}{2}\prod_{r= kQ+1}^{(k+1)Q-1}\mathbf{D}_r \mathbf{1}\mathbf{1}^T\mathbf{D}_{kQ} $ to be: 
   $$\lambda_1=0,~\lambda_2=\frac{1}{4}\prod^{(k+1)Q-1}_{r=kQ+2}(1-\frac{1}{\sqrt{r}})(1-\frac{1}{\sqrt{kQ}})+\frac{3}{4}\prod^{(k+1)Q-1}_{r=kQ+2}(1+\frac{1}{\sqrt{r}})(1+\frac{1}{\sqrt{kQ}}).$$
  It is clear that $\lambda_2$ is strictly larger than one which indicates that the algorithm will diverge. \hfill{$\blacksquare$}

\section{Proofs for Results in Section \ref{sec:fedpd}}\label{app:convergence}

\subsection{Proof of \thref{th:FedPD_Comm}}
First let us prove \thref{th:FedPD_Comm} about the FedPD algorithm with Oracle I. 
	
Towards this end, let us first introduce some notations. First recall that when Oracle I is used, the local problem is solved such that the following holds true: 
	\begin{align}\label{eq:local:accuracy}
	\norm{\nabla_{\xx_i}\cL(\xx^{r+1}_i,\xx^r_0,\lambda^r_i)}^2\leq \epsilon_1.
	\end{align}
Note that if SGD is applied in Oracle I to solve the local problem, then this condition \eqref{eq:local:accuracy} is replaced with the following
    \begin{align}\label{eq:local:accuracy:sto}
	\E[\norm{\nabla_{\xx_i}\cL(\xx^{r+1}_i,\xx^r_0,\lambda^r_i)}^2]\leq \epsilon_1.
	\end{align}
The difference does not significantly change the proofs and the results. So throughout the proof of \thref{th:FedPD_Comm}, we use \eqref{eq:local:accuracy} as the condition. 

Throughout the proof, we denote the expectation taken on the communication $r^\mathrm{th}$ iteration to the ${r+1}^{\mathrm{th}}$ iteration conditioning on all the previous knowledge as $\E_{r+1}$.
	
Then we define the error between different nodes as 
	\begin{align}\label{eq:error:node}
	\triangle^r\bydef [\triangle\xx^r_0;\triangle\xx^r], \; \mbox{with}\; \triangle\xx^r_0\bydef \max_{i,j}\norm{\xx^r_{0,i}-\xx^r_{0,j}}, \; \triangle\xx^r\bydef \max_{i,j}\norm{\xx^r_{i}-\xx^r_{j}}. 
	\end{align}
	Here, $\triangle\xx^r_0$ denotes the maximum difference of estimated center model among all the nodes and $\triangle\xx^r$ denotes the maximum difference of local models among all nodes. 

	{From the  termination condition that generates $\xx^{r+1}_i$ (given in \eqref{eq:local:accuracy}), we have 
		\begin{align}\label{eq:sub:opt}
		\nabla f_i(\xx^{r+1}_i)+\lambda^{r+1}_i=\nabla f_i(\xx^{r+1}_i)+\lambda^r_i+\frac{1}{\eta}(\xx^{r+1}_i-\xx^{r}_{0,i})=\ee^{r+1}_i,\;\; \mbox{where} ~\norm{\ee^{r+1}_i}^2\leq \epsilon_1,
		\end{align}
		where the first equality holds because of the update rule of $\lambda_i$. 
		 Furthermore, from the update step of $\lambda^{r+1}_i$, we can explicitly write down the following expression 
		$$\lambda^{r+1}_i = \lambda^{r}_i+\frac{1}{\eta}(\xx^{r+1}_i-\xx^r_{0,i}) = -\nabla f_i(\xx^{r+1}_i) + \ee^{r+1}_i.$$}

The main lemmas that we need are outlined below. Their proofs can be found in Sec. \ref{app:sub:1}-- \ref{app:sub:4}.

The first lemma shows the sufficient descent of the local AL function. 
\begin{lemma}\label{le:FedPD_descent}
Suppose \asref{as:smooth} holds true. Consider FedPD with \algref{alg:FedPD_v2} (Oracle I) as the update rule. When the local problem is solved such that \eqref{eq:local:accuracy} is satisfied, the difference of the local augmented Lagrangian is bounded by
\begin{equation}\label{eq:le_PD_decent}
\begin{aligned}
    &\cL_i(\xx^{r+1}_i,\xx^{r+}_{0,i},\lambda^{r+1}_i)-\cL_i(\xx^{r}_i,\xx^{r}_{0,i},\lambda^{r}_i)\\
    &\leq -\frac{1-2L\eta}{2\eta}\norm{\xx^{r+1}_i-\xx^r_i}^2-\frac{1}{2\eta}\norm{\xx^{r+}_{0,i}-\xx^{r}_{0,i}}^2+\eta\norm{\lambda^{r+1}_i-\lambda^r_i}^2 + \frac{\epsilon_1}{2L}.
    \end{aligned}
\end{equation}
\end{lemma}

Then we derive a key lemma about how the error propagates if the communication step is skipped. 
\begin{lemma}\label{le:FedPD_prop}
Suppose \asref{as:smooth} and \asref{as:heterogeneous} hold. Consider FedPD with \algref{alg:FedPD_v2} (Oracle I) as the update rule. When the local problem is solved such that \eqref{eq:local:accuracy} is satisfied, the difference between the {local models $\xx^r_i$'s and the difference between local copies of the global models $\xx^r_{0,i}$'s} are bounded by
\begin{equation}
    \begin{aligned}
        {\black \E_{r+1}}\triangle^{r+1} &\leq \frac{1}{1-L\eta}(A\triangle^r + \eta B(\delta+2\sqrt{\epsilon_1}).
    \end{aligned}
\end{equation}
where 

\begin{equation*}
    A = \left[\begin{array}{cc}
         p(1+L\eta)& L\eta(1-L\eta) \\
         1&L\eta 
    \end{array}\right]
\end{equation*}
is a rank $1$ matrix with eigenvalues $(0, L\eta+ p(1+L\eta))$ and $B = [p(3+L\eta),2]^{\T}$.
\end{lemma}

We define a virtual sequence $\{\overline{\xx}^r_{0}\}$ where $\overline{\xx}^r_0\bydef \frac{1}{N}\sum^N_{i=1}\xx^r_{0,i}$ which is the average of the local $\xx^r_{0,i}$ and we know that $\xx^r_{0,i}=\xx^r_0$ when $r \mod R=1$, that is, when the communication and aggregation step is performed.
Next,  we bound the error between the local AL and the global AL evaluated at the virtual sequence. 

\begin{lemma}\label{le:FedPD_err}
Suppose \asref{as:smooth} holds. Consider FedPD with \algref{alg:FedPD_v2} (Oracle I) as the update rule. When the local problem is solved such that \eqref{eq:local:accuracy} is satisfied, the difference between local AL and the global AL is bounded as below:
    \begin{equation}
        \frac{1}{N}\sum^N_{i=1}[\cL_i(\xx^{r+1}_i,\xx^{r+}_{0,i},\lambda^{r+1}_i)-cL_i(\xx^{r+1}_i,\overline{\xx}^{r+1}_{0},\lambda^{r+1}_i)]\geq-\frac{(N-1)}{2N\eta}(\triangle\xx^{r+1}_0)^2.
    \end{equation}
\end{lemma}

Lastly we bound the original objective function using the global AL.

\begin{lemma}\label{le:FedPD_gradient}
Under \asref{as:smooth} and \asref{as:lower-bounded}, when the local problem is solved to $\epsilon_1$ accuracy, the difference between the original loss and the augmented Lagrangian is bounded.
    \begin{equation}
        f(\xx^r_0)\leq \cL(\xx^{r}_0,\xx^{r}_1,\dots,\xx^{r}_N,\lambda^{r}_1,\dots,\lambda^{r}_N)-\frac{1-2L\eta}{N\eta}\sum^N_{i=1}\norm{\xx^r_i-\xx^r_0}^2+\frac{\epsilon_1}{2L}.
    \end{equation}
\end{lemma}

Using the previous lemmas, we can then prove \thref{th:FedPD_Comm}.

\subsubsection{Proof of \leref{le:FedPD_descent}}\label{app:sub:1}

We divide the left hand side (LHS) of \eqref{eq:le_PD_decent}, i.e.,  $\cL_i(\xx^{r+1}_i,\xx^{r+}_{0,i},\lambda^{r+1}_i)-\cL_i(\xx^{r}_i,\xx^{r}_{0,i},\lambda^{r}_i)$, into the sum of three parts: 
\begin{equation}\label{eq:three}
\begin{aligned}
\cL_i(\xx^{r+1}_i,\xx^{r+}_{0,i},\lambda^{r+1}_i)-\cL_i(\xx^{r}_i,\xx^{r}_{0,i},\lambda^{r}_i)&=\cL_i(\xx^{r+1}_i,\xx^{r}_{0,i},\lambda^{r}_i)-\cL_i(\xx^{r}_i,\xx^{r}_{0,i}\lambda^{r}_i)\\
&+\cL_i(\xx^{r+1}_i,\xx^{r}_{0,i},\lambda^{r+1}_i)-\cL_i(\xx^{r+1}_i,\xx^{r}_{0,i},\lambda^{r}_i)\\
 &+\cL_i(\xx^{r+1}_i,\xx^{r+}_{0,i},\lambda^{r+1}_i)-\cL_i(\xx^{r+1}_i,\xx^{r}_{0,i},\lambda^{r+1}_i),
\end{aligned}
\end{equation} 
which correspond to the three steps in the algorithm's update steps. 

We bound the first difference by first applying \asref{as:smooth} to $-f(\cdot)$: $$-f_i(\xx^r_i)\leq -f_i(\xx^{r+1}_i) + \lin{-\nabla f_i(\xx^{r+1}_i),\xx^{r}_i-\xx^{r+1}_i} + \frac{L}{2}\norm{\xx^{r}_i-\xx^{r+1}_i}^2,$$ and obtain the following series of inequalities:
\begin{equation}\label{eq:le_PD_descent_1}
    \begin{aligned}
        \cL_i(\xx^{r+1}_i,\xx^{r}_{0,i},\lambda^{r}_i)-&\cL_i(\xx^{r}_i,\xx^{r}_{0,i},\lambda^{r}_i)\leq \lin{\nabla f_i(\xx^{r+1}_i),\xx^{r+1}_i-\xx^{r}_i} + \frac{L}{2}\norm{\xx^{r+1}_i-\xx^r_i}^2\\
        &+ \lin{\lambda^r_i, \xx^{r+1}_i-\xx^{r}_i}+ \frac{1}{2\eta}\norm{\xx^{r+1}_i-\xx^{r}_{0,i}}^2-\frac{1}{2\eta}\norm{\xx^{r}_i-\xx^{r}_{0,i}}^2\\
        \stackrel{(a)}{=}~& \lin{\nabla f_i(\xx^{r+1}_i)+\lambda^r_i,\xx^{r+1}_i-\xx^{r}_i} + \frac{L}{2}\norm{\xx^{r+1}_i-\xx^r_i}^2 \\
        &+ \frac{1}{2\eta}\lin{\xx^{r+1}_i + \xx^{r}_i -2\xx^{r}_{0,i}, \xx^{r+1}_i-\xx^{r}_{i}}\\
        \stackrel{(b)}{=}~& \lin{\nabla f_i(\xx^{r+1}_i)+\lambda^r_i+\frac{1}{\eta}(\xx^{r+1}_i-\xx^{r}_{0,i}),\xx^{r+1}_i-\xx^{r}_i} + \frac{L}{2}\norm{\xx^{r+1}_i-\xx^r_i}^2 \\
        &- \frac{1}{2\eta}\norm{\xx^{r+1}_i - \xx^{r}_i}^2\\
        \stackrel{(c)}{\leq}~& \frac{1}{2L}\norm{\nabla f_i(\xx^{r+1}_i)+\lambda^r_i+\frac{1}{\eta}(\xx^{r+1}_i-\xx^{r}_{0,i})}^2+\frac{L}{2}\norm{\xx^{r+1}_i-\xx^{r}_i}^2\\
        &- \frac{1-L\eta}{2\eta}\norm{\xx^{r+1}_i-\xx^r_i}^2\\
        \stackrel{(d)}{\leq}~& -\frac{1-2L\eta}{2\eta}\norm{\xx^{r+1}_i-\xx^r_i}^2+\frac{\epsilon_1}{2L}.
    \end{aligned}
\end{equation}

In the above equation, in $(a)$ we use the fact that $\norm{a}^2 - \norm{b}^2 = \lin{a+b,a-b}$ when vector $a,b$ has the same length to the last two terms; in $(b)$ we split the last term into $2\xx^{r+1}_i-2\xx^{r}_{0,i}$ and $-\xx^{r+1}_i+\xx^{r}_{i}$;  {in $(c)$ we use the fact that $\lin{a,b}\leq\frac{L}{2}\norm{a}^2+\frac{1}{2L}\norm{b}^2)$}; in $(d)$ we apply the fact that $\xx^{r+1}_i$ is the inexact solution; see \eqref{eq:sub:opt}.

Then we bound the second difference in  \eqref{eq:three} by the following:

\begin{equation}\label{eq:le_PD_descent_2}
    \begin{aligned}
        \cL_i(\xx^{r+1}_i,\xx^{r}_{0,i},\lambda^{r+1}_i)-\cL_i(\xx^{r+1}_i,\xx^{r}_{0,i},\lambda^{r}_i) &= \lin{\lambda^{r+1}_i-\lambda^{r}_i, \xx^{r+1}_i-\xx^{r}_{0,i}}\\
        & \stackrel{(a)}{=} \lin{\lambda^{r+1}_i-\lambda^{r}_i, \eta(\lambda^{r+1}_i-\lambda^{r}_i)}\\
        & = \eta\norm{\lambda^{r+1}_i-\lambda^{r}_i}^2,
    \end{aligned}
\end{equation}
where $(a)$ directly comes from the update rule of $\lambda^{r+1}_i$.

Further we bound the third difference in  \eqref{eq:three} by the following:
\begin{equation}\label{eq:le_PD_descent_3}
    \begin{aligned}
        &\cL_i(\xx^{r+1}_i,\xx^{r+}_{0,i},\lambda^{r+1}_i)-\cL_i(\xx^{r+1}_i,\xx^{r}_{0,i},\lambda^{r+1}_i) \\
        & = \lin{\lambda^{r+1}_i, \xx^{r+1}_i-\xx^{r+}_{0,i}}-\lin{\lambda^{r+1}_i, \xx^{r+1}_i-\xx^{r}_{0,i}}  + \frac{1}{2\eta}\norm{\xx^{r+1}_i-\xx^{r+}_{0,i}}^2- \frac{1}{2\eta}\norm{\xx^{r+1}_i-\xx^{r}_{0,i}}^2\\
        & \stackrel{(a)}{=} \lin{\lambda^{r+1}_i, \xx^{r}_{0,i}-\xx^{r+}_{0,i}} + \frac{1}{2\eta}\lin{2\xx^{r+1}_i-2\xx^{r+}_{0,i}+\xx^{r+}_{0,i}-\xx^{r}_{0,i},\xx^{r}_{0,i}-\xx^{r+}_{0,i}}\\
        & = \lin{\frac{1}{\eta}(\eta\lambda^{r+1}_i+\xx^{r+1}_i-\xx^{r+}_{0,i}), \xx^{r}_{0,i}-\xx^{r+}_{0,i}} - \frac{1}{2\eta}\norm{\xx^{r+}_{0,i}-\xx^{r}_{0,i}}^2\\
        &\stackrel{(b)}{=} - \frac{1}{2\eta}\norm{\xx^{r+}_{0,i}-\xx^{r}_{0,i}}^2,
    \end{aligned}
\end{equation}
where, in $(a)$, we use the same reasoning as in \eqref{eq:le_PD_descent_1} $(a)$ and $(b)$; in $(b)$ we apply the update rule of $\xx^{r+}_{0,i}$ in the FedPD algorithm, which implies that the first term becomes zero.

Finally we sum up \eqref{eq:le_PD_descent_1}, \eqref{eq:le_PD_descent_2}, \eqref{eq:le_PD_descent_3} and \leref{le:FedPD_descent} is proved.

\subsubsection{Proof of \leref{le:FedPD_prop}}

First we derive the relation between $\norm{\xx^{r+1}_i-\xx^{r+1}_j}$ for arbitrary $i\neq j$ and $\triangle^r$ by using the definition of $\epsilon_1$ \eqref{eq:sub:opt}:
\begin{equation}\label{eq:le_PD_prop_1}
    \begin{aligned}
        \norm{\xx^{r+1}_i-\xx^{r+1}_j} \stackrel{\eqref{eq:sub:opt}}=&~ \norm{\xx^{r}_{0,i}-\xx^{r}_{0,j} -\eta(\nabla f_i(\xx^{r+1}_i)+\lambda^r_i-\ee^{r+1}_i-\nabla f_j(\xx^{r+1}_j)-\lambda^r_j+\ee^{r+1}_j)}\\
        \leq &~ \norm{\xx^{r}_{0,i}-\xx^{r}_{0,j}} + \eta\norm{\nabla f_i(\xx^{r+1}_i)-\nabla f_j(\xx^{r+1}_j)}\\
        &\quad+\eta\norm{\lambda^r_i-\lambda^r_j} +\eta(\norm{\ee^{r+1}_i}+\norm{\ee^{r+1}_j})\\
        \stackrel{(a)}{\leq} &~ \triangle\xx^r_0 + \eta\norm{\nabla f_i(\xx^{r+1}_i)-\nabla f_i(\xx^{r+1}_j)+\nabla f_i(\xx^{r+1}_j)-\nabla f_j(\xx^{r+1}_j)}\\
        &\quad+\eta\norm{\lambda^r_i-\lambda^r_j} +2\eta\sqrt{\epsilon_1}\\
        \stackrel{(b)}{\leq} &~ \triangle\xx^r_0 + L\eta\norm{\xx^{r+1}_i-\xx^{r+1}_j}+\eta \norm{\nabla f_i(\xx^{r+1}_j)-\nabla f_j(\xx^{r+1}_j)}\\
        &\quad +\eta\norm{\lambda^r_i-\lambda^r_j} +2\eta\sqrt{\epsilon_1}\\
        \stackrel{(c)}{\leq} &~ \triangle\xx^r_0 + L\eta\norm{\xx^{r+1}_i-\xx^{r+1}_j}+\eta \delta +\eta\norm{\lambda^r_i-\lambda^r_j} +2\eta\sqrt{\epsilon_1}\\
        \stackrel{(d)}{=} &~ \frac{1}{1-L\eta}\triangle\xx^r_0 + \frac{\eta}{1-L\eta} \delta +\frac{\eta}{1-L\eta}\norm{\lambda^r_i-\lambda^r_j} +\frac{2\eta}{1-L\eta}\sqrt{\epsilon_1},
    \end{aligned}
\end{equation}
where in $(a)$ we plug the definition of $\triangle\xx^r_0$ and $\ee^{r+1}_i$; in $(b)$ we use \asref{as:smooth}; $(c)$ comes form \asref{as:heterogeneous}; in $(d)$ we move the second term to the left and divide both side by $1-L\eta$.

Then we bound the difference $\norm{\lambda^r_i-\lambda^r_j}$ by plugging in the expression of $\lambda^{r}_i$ in \eqref{eq:sub:opt}, and note that $\lambda^r_i+\frac{1}{\eta}(\xx^{r+1}_i-\xx^{r}_{0,i}) = \lambda^{r+1}_i$:
\begin{equation}\label{eq:le_PD_prop_2}
    \begin{aligned}
        \norm{\lambda^r_i-\lambda^r_j} =&~ \norm{-\nabla f_i(\xx^{r}_i) + \ee^{r}_i +\nabla f_j(\xx^{r}_j) - \ee^{r}_j}\\
        \stackrel{(a)}{\leq} &~ \norm{\nabla f_i(\xx^{r}_i) - \nabla f_i(\xx^{r}_j)}+ \norm{\nabla f_i(\xx^{r}_j)-\nabla f_j(\xx^{r}_j)} + 2\sqrt{\epsilon_1}\\
        \stackrel{(b)}{\leq} &~ L\norm{\xx^{r}_i - \xx^{r}_j}+\delta + 2\sqrt{\epsilon_1}\\
        \stackrel{(c)}{\leq} &~ L\triangle\xx^r+\delta + 2\sqrt{\epsilon_1},\\
    \end{aligned}
\end{equation}
where $(a)$ and $(b)$ follow the same argument in $(a)$, $(b)$ and $(c)$ of \eqref{eq:le_PD_prop_1} ; in $(c)$ we plug in the definition of $\triangle\xx^r$.

Next we bound the difference $\norm{\xx^{r+1}_{0,i}-\xx^{r+1}_{0,j}}$. With probability $1-p$ the aggregation step has just been done at iteration $r$, $\xx^{r+1}_{0,i}=\xx^{r+1}_{0,j}$.{\black With probability $p$, they are not equal, then we take expectation with communication probability $p$}, and get
\begin{equation}\label{eq:le_PD_prop_3}\black
    \begin{aligned}
        \E_{r+1}\norm{\xx^{r+1}_{0,i}-\xx^{r+1}_{0,j}} =&~ p\norm{\xx^{r+1}_i-\xx^{r+1}_j + \eta(\lambda^{r+1}_i-\lambda^{r+1}_j)}\\
        \leq &~ p\norm{\xx^{r+1}_i-\xx^{r+1}_j} + p\eta\norm{\lambda^{r+1}_i-\lambda^{r+1}_j}\\
        \stackrel{(a)}{\leq} &~ p(1+L\eta)\triangle\xx^{r+1} + p\eta(\delta +2\sqrt{\epsilon_1}),
    \end{aligned}
\end{equation}
where in $(a)$ we plug in the definition of $\triangle\xx^{r+1}$ and \eqref{eq:le_PD_prop_2}. {As these relations hold true for arbitrary $(i,j)$ pairs, they are also true for the maximum of $\norm{\xx^{r+1}_i-\xx^{r+1}_j}$ and $\norm{\xx^{r+1}_{0,i}-\xx^{r+1}_{0,j}}$.}

Therefore stacking \eqref{eq:le_PD_prop_1} and \eqref{eq:le_PD_prop_3} and plug in \eqref{eq:le_PD_prop_2}, we have

\begin{equation}
    \begin{aligned}
        \triangle\xx^{r+1}\leq & \frac{1}{1-L\eta}(L\eta\triangle\xx^r+\triangle\xx^r_0) + \frac{2\eta}{1-L\eta}(\delta + 2 \sqrt{\epsilon_1}),\\
        {\E_{r+1}}\triangle\xx^{r+1}_0\leq & {\black p\frac{1+L\eta}{1-L\eta}(L\eta\triangle\xx^r+\triangle\xx^r_0) + p\frac{\eta(3+L\eta)}{1-L\eta}(\delta + 2 \sqrt{\epsilon_1})}.
    \end{aligned}
\end{equation}

Rewrite it into matrix form then we complete the proof of \leref{le:FedPD_prop}.

\subsubsection{Proof of \leref{le:FedPD_err}}
Let us first recall that the definition of local AL is given below: $$\cL_i(\xx_i,\xx_0,\lambda_i)\bydef~f_i(\xx_i)+\lin{\lambda_i,\xx_i-\xx_0}+\frac{1}{2\eta}\norm{\xx_i-\xx_0}^2.$$
Similar to \eqref{eq:le_PD_descent_3}, we have 

\begin{equation}\label{eq:le_PD_err_1}
    \begin{aligned}
        \cL_i(\xx^{r+1}_i,\xx^{r+}_{0,i},\lambda^{r+1}_i)&-\cL_i(\xx^{r+1}_i,\overline{\xx}^{r+1}_{0},\lambda^{r+1}_i) = \lin{\lambda^{r+1}_i, \xx^{r+1}_i-\xx^{r+}_{0,i}}-\lin{\lambda^{r+1}_i, \xx^{r+1}_i-\overbar{\xx}^{r+1}_{0}} \\
        &\quad + \frac{1}{2\eta}\norm{\xx^{r+1}_i-\xx^{r+}_{0,i}}^2- \frac{1}{2\eta}\norm{\xx^{r+1}_i-\overline{\xx}^{r+1}_{0}}^2\\
        & \stackrel{(a)}{=} -\frac{1}{2\eta}\norm{\xx^{r+}_{0,i}-\overline{\xx}^{r+1}_{0}}^2\\
        & \stackrel{(b)}{=} -\frac{1}{2\eta}\norm{\xx^{r+}_{0,i}-\frac{1}{N}\sum^{N}_{j=1}\xx^{r+}_{0,j}}^2\\
        & = -\frac{1}{2\eta}\norm{\frac{1}{N}\sum^{N}_{j=1}(\xx^{r+}_{0,i}-\xx^{r+}_{0,j})}^2\\
        &\stackrel{(c)}{\geq} -\frac{1}{2\eta N}\sum_{j\neq i}\norm{\xx^{r+}_{0,i}-\xx^{r+}_{0,j}}^2\\
        &\stackrel{(d)}{\geq} -\frac{N-1}{2\eta N}(\triangle\xx^{r+1}_0)^2,\\
    \end{aligned}
\end{equation}
where $(a)$ follows the same argument in \eqref{eq:le_PD_descent_3}; in $(b)$,we plug in the definition of $\overline{\xx}^{r+1}_{0}$;  in $(c)$ we use Jensen's inequality and we bound the term with $\triangle\xx^{r+1}_0$. Then the lemma is proved. 
\subsubsection{Proof of \leref{le:FedPD_gradient}}\label{app:sub:4}

Applying \asref{as:smooth}, we have

\begin{equation}\label{eq:le_PD_gradient_1}
    \begin{aligned}
        f_i(\xx^{r}_0)\leq &~ f_i(\xx^r_i)+\lin{\nabla f_i(\xx^r_i),\xx^r_0-\xx^r_i}+\frac{L}{2}\norm{\xx^r_0-\xx^r_i}^2\\
        \stackrel{\eqref{eq:sub:opt}}= &~ \cL_i(\xx^r_i, \xx^r_0, \lambda^r_i) - \lin{\ee^r_i, \xx^r_0-\xx^r_i} - \frac{1-L\eta}{2\eta}\norm{\xx^r_0-\xx^r_i}^2\\
        \leq &~ \cL_i(\xx^r_i, \xx^r_0, \lambda^r_i) + \frac{\epsilon_1}{2L} - \frac{1-2L\eta}{2\eta}\norm{\xx^r_0-\xx^r_i}^2.\\
    \end{aligned}
\end{equation}

Taking an average over $N$ agents we are able to prove \leref{le:FedPD_gradient}.

\subsubsection{Proof of \thref{th:FedPD_Comm}}

First notice that from the optimality condition \eqref{eq:sub:opt}, the following holds:
\begin{equation}\label{eq:th_comm_0}
    \begin{aligned}
        \norm{\lambda^r_i-\lambda^{r-1}_i}^2&\leq 2L^2\norm{\xx^r_i-\xx^{r-1}_i}^2+4\epsilon_1.
    \end{aligned}
\end{equation}

Then we bound the gradients of $\cL(\xx^{r}_i,\xx^{r}_{0,i},\lambda^{r}_i)$.

\begin{equation}\label{eq:th_comm_1}
    \begin{aligned}
        &\norm{\nabla_{\xx_i}\cL_i(\xx^r_i,\xx^{r}_{0,i},\lambda^r_i)}=\norm{\nabla f_i(\xx^r_i)+\lambda^r_i +\frac{1}{\eta}(\xx^r_i-\xx^{r}_{0,i})}\\
        &\stackrel{\eqref{eq:sub:opt}}=\norm{\nabla f_i(\xx^r_i)+\lambda^r_i +\frac{1}{\eta}(\xx^r_i-\xx^{r}_{0,i})-\nabla f_i(\xx^{r+1}_i)-\lambda^r_i -\frac{1}{\eta}(\xx^{r+1}_i-\xx^{r}_{0,i})+\ee^{r+1}_i}\\
        &\leq \frac{1+L\eta}{\eta}\norm{\xx^{r+1}_i-\xx^{r}_i} +\sqrt{\epsilon_1}.\\
    \end{aligned}
\end{equation}

Further, we note that, {when no aggregation has been performed at iteration $r$, then $\xx^{r}_{0,i} = \xx^r_i+ \eta \lambda^r_i$, so the following holds}
\begin{equation}\label{eq:th_comm_2_1}
    \begin{aligned}
        \norm{\nabla_{\xx_0}\cL_i(\xx^r_i,\xx^{r}_{0,i},\lambda^r_i)}=&~\norm{\lambda^r_i +\frac{1}{\eta}(\xx^r_i-\xx^{r}_{0,i})}=0.
    \end{aligned}
\end{equation}

When there the aggregation has been performed at iteration $r$, then $\xx^{r}_{0,i} =\frac{1}{N} \sum^{N}_{j=1}(\xx^r_j+ \eta \lambda^r_j),~\forall i$, so we have
\begin{equation}\label{eq:th_comm_2_2}
    \begin{aligned}
        \norm{\nabla_{\xx_0}\cL(\xx^{r}_0,\xx^{r}_1,\dots,\xx^{r}_N,\lambda^{r}_1,\dots,\lambda^{r}_N)}=&~\norm{\frac{1}{N}\sum^N_{i=1}(\lambda^r_i +\frac{1}{\eta}(\xx^r_i-\xx^{r}_{0,i}))}=0.
    \end{aligned}
\end{equation}

Further we have:
\begin{equation}\label{eq:th_comm_3}
    \begin{aligned}
        \norm{\nabla_{\lambda_i}\cL_i(\xx^r_i,\xx^{r}_{0,i},\lambda^r_i)}=&~\norm{\xx^r_i-\xx^{r}_{0,i}}\\
        \leq &~ \norm{\xx^r_i-\xx^{r-1}_{0,i}}+\norm{\xx^{r-1}_{0,i}-\xx^{r}_{0,i}}\\
        \leq &~ \eta\norm{\lambda^r_i-\lambda^{r-1}_i}+\norm{\xx^{r-1}_{0,i}-\xx^{r}_{0,i}}\\
        \leq &~ \eta(L\norm{\xx^r_i-\xx^{r-1}_i}+2\sqrt{\epsilon_1})+\norm{\xx^{r-1}_{0,i}-\xx^{r}_{0,i}}.
    \end{aligned}
\end{equation}

Summing \eqref{eq:th_comm_1} and \eqref{eq:th_comm_3}, denote $\norm{\nabla_{\xx_i}\cL_i(\xx^r_i,\xx^{r}_{0,i},\lambda^r_i)}+\norm{\nabla_{\lambda_i}\cL_i(\xx^r_i,\xx^{r}_{0,i},\lambda^r_i)}$ as $\norm{\nabla\cL_i(\xx^r_i,\xx^{r}_{0,i},\lambda^r_i)}$ we have

\begin{equation}\label{eq:th_comm_4}
    \begin{aligned}
        \norm{\nabla\cL_i(\xx^r_i,\xx^{r}_{0,i},\lambda^r_i)}\leq \norm{\xx^{r-1}_{0,i}-\xx^{r}_{0,i}} + \frac{1+L\eta}{\eta}\norm{\xx^{r+1}_i-\xx^{r}_i} + L\eta\norm{\xx^r_i-\xx^{r-1}_i} + (1+2\eta)\sqrt{\epsilon_1}.
    \end{aligned}
\end{equation}

Squaring both sides of the above inequality, we obtain: 
\begin{equation}\label{eq:th_comm_5}
    \begin{aligned}
        \norm{\nabla\cL_i(\xx^r_i,\xx^{r}_{0,i},\lambda^r_i)}^2\leq C_6\left(\norm{\xx^{r-1}_{0,i}-\xx^{r}_{0,i}}^2 + \norm{\xx^{r+1}_i-\xx^{r}_i}^2 + \norm{\xx^r_i-\xx^{r-1}_i}^2 + \epsilon_1\right),
    \end{aligned}
\end{equation}
where $C_6\geq\max\{(\frac{1+L\eta}{\eta})^2, (1+2\eta)^2, L^2\eta^2\}$.

Apply \eqref{eq:th_comm_0} to \leref{le:FedPD_descent} we have

\begin{equation}\label{eq:th_comm_6}
    \begin{aligned}
        \frac{1-2L\eta-4L^2\eta^2}{2\eta}\norm{\xx^{r+1}_i-\xx^r_i}^2+&\frac{1}{2\eta}\norm{\xx^{r+}_{0,i}-\xx^{r}_{0,i}}^2 + \frac{1+8L\eta}{2L}\epsilon_1\\
        &\leq \cL_i(\xx^{r}_i,\xx^{r}_{0,i},\lambda^{r}_i)-\cL_i(\xx^{r+1}_i,\xx^{r+}_{0,i},\lambda^{r+1}_i) + \frac{1+8L\eta}{L}\epsilon_1.
    \end{aligned}
\end{equation}
Notice that when communication is not performed $\norm{\xx^{r}_{0,i}-\xx^{r+1}_{0,i}}^2{\le} \norm{\xx^{r}_{0,i}-\xx^{r+}_{0,i}}^2$, and when communication is performed
\begin{equation}\label{eq:th_comm_6_2_0}
    \begin{aligned}
        \frac{1}{N}\sum^N_{i=1}\norm{\xx^{r}_{0,i}-\xx^{r+1}_{0,i}}^2 = \frac{2}{N}\sum^N_{i=1}\norm{\xx^{r}_{0,i}-\xx^{r+}_{0,i}}^2 +\frac{2}{N}\sum^N_{i=1}\norm{\xx^{r+}_{0,i}-\xx^{r+1}_{0,i}}^2\\
        \leq \frac{2}{N}\sum^N_{i=1}\norm{\xx^{r}_{0,i}-\xx^{r+}_{0,i}}^2 + \frac{N-1}{\eta N}(\triangle\xx^{r+1}_0)^2,
    \end{aligned}
\end{equation}
{where the last inequality holds due to the use of Jensen's inequality, and the definition of $\triangle\xx^{r+1}_0$ in \eqref{eq:error:node}. It follows that summing both sides of  \eqref{eq:th_comm_6} over $i$, we have 
\begin{equation}\label{eq:th_comm_6_2}
    \begin{aligned}
        & \frac{1-2L\eta-4L^2\eta^2}{2\eta}\sum_{i=1}^{N}\norm{\xx^{r+1}_i-\xx^r_i}^2+\sum_{i=1}^{N}(\frac{1}{4\eta}\norm{\xx^{r+1}_{0,i}-\xx^{r}_{0,i}}^2 - \frac{N-1}{4\eta }(\triangle\xx^{r+1}_0)^2) + \frac{N(1+8L\eta)}{2L}\epsilon_1\\
        &\leq \sum_{i=1}^{N}\left(\cL_i(\xx^{r}_i,\xx^{r}_{0,i},\lambda^{r}_i)-\cL_i(\xx^{r+1}_i,\xx^{r+}_{0,i},\lambda^{r+1}_i)\right) + \frac{N(1+8L\eta)}{L}\epsilon_1.
    \end{aligned}
\end{equation}}

Taking the expectation over the randomness in $p$, conditioning on the information before the communication the successive difference of $\mathcal{L}_i$,  
\begin{equation}\label{eq:th_comm_6_1}
    \begin{aligned}
        \E_{r+1}\frac{1}{N}\sum^N_{i=1}&[\cL_i(\xx^{r}_i,\xx^{r}_{0,i},\lambda^{r}_i) -\cL_i(\xx^{r+1}_i,\xx^{r+}_{0,i},\lambda^{r+1}_i)] = \frac{1}{N}\sum^N_{i=1}[\cL_i(\xx^{r}_i,\xx^{r}_{0,i},\lambda^{r}_i) - \cL_i(\xx^{r+1}_i,\xx^{r+1}_{0,i},\lambda^{r+1}_i)]\\
        &+\E_{r+1}\frac{1}{N}\sum^N_{i=1}[\cL_i(\xx^{r+1}_i,\xx^{r+1}_{0,i},\lambda^{r+1}_i) - \cL_i(\xx^{r+1}_i,\xx^{r+}_{0,i},\lambda^{r+1}_i)] \\
        & \stackrel{(a)}= \frac{1}{N}\sum^N_{i=1}[\cL_i(\xx^{r}_i,\xx^{r}_{0,i},\lambda^{r}_i) - \cL_i(\xx^{r+1}_i,\xx^{r+1}_{0,i},\lambda^{r+1}_i)]\\
        &+\frac{1}{N}\sum^N_{i=1}[p\cL_i(\xx^{r+1}_i,\overline{\xx}^{r+1}_{0},\lambda^{r+1}_i) +(1-p)\cL_i(\xx^{r+1}_i,\xx^{r+}_{0,i},\lambda^{r+1}_i) - \cL_i(\xx^{r+1}_i,\xx^{r+}_{0,i},\lambda^{r+1}_i)] \\
        & =\frac{1}{N}\sum^N_{i=1}[\cL_i(\xx^{r}_i,\xx^{r}_{0,i},\lambda^{r}_i) - \cL_i(\xx^{r+1}_i,\xx^{r+1}_{0,i},\lambda^{r+1}_i)]\\
        & +p\frac{1}{N}\sum^N_{i=1}[\cL_i(\xx^{r+1}_i,\overline{\xx}^{r+1}_{0},\lambda^{r+1}_i) - \cL_i(\xx^{r+1}_i,\xx^{r+}_{0,i},\lambda^{r+1}_i) ] \\
        & \stackrel{(b)}{\leq}\frac{1}{N}\sum^N_{i=1}[\cL_i(\xx^{r}_i,\xx^{r}_{0,i},\lambda^{r}_i) - \cL_i(\xx^{r+1}_i,\xx^{r+1}_{0,i},\lambda^{r+1}_i)] + p\frac{N-1}{2\eta N}(\triangle\xx^{r+1}_0)^2,
    \end{aligned}
\end{equation}
where {(a) expands the expectation on $p$, and use the fact that with probability $p$, $\xx^{r+1}_{0,i}=\xx^{r+}_{0,i}$, and with probability $(1-p)$ $\xx^{r+1}_0$ will be updated; in (b) we apply \leref{le:FedPD_err} to the last term.}


{Combine \eqref{eq:th_comm_6_2}} and \eqref{eq:th_comm_6_1}, we have

\begin{equation}\label{eq:th_comm_6_3}
    \begin{aligned}
        &\min\{\frac{1-2L\eta-4L^2\eta^2}{2\eta}, \frac{1}{2\eta}, \frac{1+8L\eta}{2L}\}\frac{1}{N}\sum^N_{i=1}\E_{r+1}\left[\norm{\xx^{r+1}_i-\xx^r_i}^2+\norm{\xx^{r+1}_{0,i}-\xx^{r}_{0,i}}^2 + \epsilon_1\right]\\
        &\leq \frac{1}{N}\sum^N_{i=1}\left[\cL_i(\xx^{r}_i,\xx^{r}_{0,i},\lambda^{r}_i)-\cL_i(\xx^{r+1}_i,\xx^{r+1}_{0,i},\lambda^{r+1}_i)\right] + \frac{1+8L\eta}{L}\epsilon_1 + p\frac{(N-1)}{\eta N}(\triangle\xx^{r+1}_0)^2.
    \end{aligned}
\end{equation}

Combining \eqref{eq:th_comm_5}, \eqref{eq:th_comm_6_2} and \eqref{eq:th_comm_6_3}, define $C_7=2C_6/\min\{\frac{1-2L\eta-4L^2\eta^2}{2\eta},\frac{1}{2\eta},\frac{1+8L\eta}{2L}\}$ and sum up the iterations, we have
\begin{equation}\label{eq:th_comm_7}
    \begin{aligned}
        &\frac{1}{N}\sum^{N}_{i=1}\sum^T_{r=0}{\black \E}\norm{\nabla\cL_i(\xx^r_i,\xx^{r}_{0,i},\lambda^r_i)}^2\\
        &\stackrel{\eqref{eq:th_comm_5}\eqref{eq:th_comm_6_2}}{\leq} \frac{2C_6}{N}\sum^{N}_{i=1}\sum^T_{r=0}\E\left[\norm{\xx^{r}_{0,i}-\xx^{r+}_{0,i}}^2 + \norm{\xx^r_i-\xx^{r+1}_i}^2 +\frac{(N-1)}{\eta N}(\triangle\xx^{r+1}_0)^2 + \epsilon_1 \right]\\
        &\stackrel{\eqref{eq:th_comm_6_3}}{\leq} C_7\sum^T_{r=0}\left(\frac{1}{N}\sum^N_{i=1}(\cL_i(\xx^{r}_i,\xx^{r}_{0,i},\lambda^{r}_i)-\cL_i(\xx^{r+1}_i,\xx^{r+1}_{0,i},\lambda^{r+1}_i))+ \frac{1+8L\eta}{L}\epsilon_1\right)\\
        &\quad+ {\black pC_7\sum^T_{r=0} }\frac{N-1}{N\eta}\E(\triangle\xx^{r+1}_0)^2.
    \end{aligned}
\end{equation}

Next we bound the last term. 
By iteratively applying \leref{le:FedPD_prop} from {\black$\tau=0$ to $r$} and use the fact that $\Delta^0 = 0$, we have

\begin{equation}\label{eq:th_comm_8_0}
    \begin{aligned}
        \E\triangle\xx^{r+1}_0 &\leq [1,0]\sum^{r}_{\tau=0}(\frac{A}{1-L\eta})^{\tau}\eta {\black\frac{[p(3+L\eta),2]^T}{1-L\eta}}(\delta+2\sqrt{\epsilon_1}).
    \end{aligned}
\end{equation}

From \leref{le:FedPD_prop} we have: 
{\black $$\lambda_{\max}\left(\frac{1}{1-L\eta}A\right)=\frac{{p(1+L\eta)+L\eta}}{1-L\eta}\triangleq C_8.$$ }
So by taking norm square on both side of \eqref{eq:th_comm_8_0},  we have

\begin{equation}\label{eq:th_comm_8}\black
    \begin{aligned}
        \E(\triangle\xx^{r+1}_0)^2&\leq \norm{[1,0]\sum^{r}_{\tau=1}(\frac{A}{1-L\eta})^{\tau}\eta \frac{[p(3+L\eta),2]^T}{1-L\eta}(\delta+2\sqrt{\epsilon_1})}^2\\
        &\leq \left(\sum^{r}_{\tau=1}C_8^{\tau}\right)^2\frac{8\eta^2(p^2(3+L\eta)^2+4)}{(1-L\eta)^2}(\delta^2+\epsilon_1)\\
        &\leq \frac{(1-C_8^{r+1})^2\times8\eta^2(p^2(3+L\eta)^2+4)}{(1-C_8)^2(1-L\eta)^2}(\delta^2+\epsilon_1).
    \end{aligned}
\end{equation}

Substitute \eqref{eq:th_comm_8} into \eqref{eq:th_comm_7} and divide both side by $T$ we have

\begin{equation}\label{eq:th_comm_9}
    \begin{aligned}
        &\frac{1}{N}\sum^{N}_{i=1}\frac{1}{T}\sum^T_{r=0}\E\norm{\nabla\cL_i(\xx^r_i,\xx^{r}_{0,i},\lambda^r_i)}^2\nonumber\\
        \leq&~ \frac{C_7}{T}\left(\cL(\xx^{0}_{0}, \xx^{0}_i,\lambda^{0}_i)-\cL(\xx^{T}_i,\xx^{T}_{0,i},\lambda^{T}_i)\right)+ \frac{C_7(1+8L\eta)}{L}\epsilon_1\\
        &+ \frac{\black 8p\eta C_7(N-1)(1-C_8^{1/(1-p)})^2(p^2(3+L\eta)^2+4)}{N(1-C_8)^2(1-L\eta)^2}(\delta^2+\epsilon_1).
    \end{aligned}
\end{equation}

From the initial conditions we have $\cL(\xx^{0}_{0}, \xx^{0}_i,\lambda^{0}_i)=f(\xx^0_0)$ and apply \leref{le:FedPD_gradient} we obtain

\begin{equation}\label{eq:th_comm_10}
    \begin{aligned}
        &\frac{1}{NT}\sum^{N}_{i=1}\sum^T_{r=0}\E\norm{\nabla\cL_i(\xx^r_i,\xx^{r}_{0,i},\lambda^r_i)}^2\nonumber\\
        \leq&~ \frac{C_7(f(\xx^0_0)-f(\xx^T_0))}{T}+ \frac{C_7(1+8L\eta)}{L}\epsilon_1\\
        &+ \frac{8p\eta C_7(N-1)(1-C_8^{1/(1-p)})^2(p^2(3+L\eta)^2+4)}{N(1-C_8)^2(1-L\eta)^2}(\delta^2+\epsilon_1).
    \end{aligned}
\end{equation}

Finally we bound $\norm{\nabla f(\xx^r_0)}^2$ by 

\begin{equation}\label{eq:th_comm_11}
    \begin{aligned}
        \norm{\nabla f(\xx^r_0)}^2 \leq&~ 2\norm{\nabla f(\xx^r_0)-\frac{1}{N}\sum^N_{i=1}\nabla_{\xx_i}\cL_i(\xx^r_i,\xx^{r}_{0,i},\lambda^r_i)}^2+\frac{2}{N}\sum^N_{i=1}\norm{\nabla_{\xx_i}\cL_i(\xx^r_i,\xx^{r}_{0,i},\lambda^r_i)}^2\\
        \leq &~  \frac{4}{N}\sum^N_{i=1}\norm{\nabla f_i(\xx^r_0)-\nabla f_i(\xx^r_i)}^2+4\norm{\frac{1}{N\eta}\sum^{N}_{i=1}(\eta\lambda^r_i+\xx^r_i-\xx^r_{0,i})}^2\\
        &+\frac{2}{N}\sum^N_{i=1}\norm{\nabla_{\xx_i}\cL_i(\xx^r_i,\xx^{r}_{0,i},\lambda^r_i)}^2\\
        \stackrel{(a)}{\leq} &~ \frac{4L^2}{N}\sum^N_{i=1}\norm{\xx^r_0-\xx^r_i}^2+\frac{2}{N}\sum^N_{i=1}\norm{\nabla_{\xx_i}\cL_i(\xx^r_i,\xx^{r}_{0,i},\lambda^r_i)}^2\\
        = &~ \frac{4L^2}{N}\sum^N_{i=1}\norm{\nabla_{\lambda_i}\cL_i(\xx^r_i,\xx^{r}_{0,i},\lambda^r_i)}^2+\frac{2}{N}\sum^N_{i=1}\norm{\nabla_{\xx_i}\cL_i(\xx^r_i,\xx^{r}_{0,i},\lambda^r_i)}^2\\
        \leq &~ \frac{4L^2}{N}\sum^N_{i=1}\norm{\nabla\cL_i(\xx^r_i,\xx^{r}_{0,i},\lambda^r_i)}^2,\\
    \end{aligned}
\end{equation}
where in $(a)$ we use the same argument in \eqref{eq:th_comm_2_1} and \eqref{eq:th_comm_2_2}.

Therefore \thref{th:FedPD_Comm} is proved. During the proof, we need all $C_2,\dots,C_7, C_8>0$, therefore, $0<\eta<\frac{\sqrt{5}-1}{4L}$.

Finally, let us note that if the local problems are solved with SGD, then the local problem needs to be solved such that the condition \eqref{eq:local:accuracy:sto} holds true. As no other information of the local solvers except error term $\ee^r_i$ is used in the proof, the proofs and results of FedPD with SGD as local solver will not change much, except that all the results hold in expectation.  Therefore we skip the proof for the SGD version. 

\subsubsection{Constants used in the proofs}

In this subsection we list all the constants $C_2,\dots,C_8$ used in the proof of \thref{th:FedPD_Comm}.
\begin{align*}
    C_2&\geq 4L^2C_7, \qquad C_3=C_8, \qquad C_4\geq \frac{C_2(1+8L\eta)}{L},\\
    C_5&=8C_2,\qquad C_6\geq\max\{(\frac{1+L\eta}{\eta})^2, (1+2\eta)^2, L^2\eta^2\},\\
    C_7&=2C_6/\min\{\frac{1-2L\eta-4L^2\eta^2}{2\eta},\frac{1}{2\eta},\frac{1+8L\eta}{2L}\},\\
    C_8&={\black \frac{{p(1+L\eta)+L\eta}}{1-L\eta}},
\end{align*}
we can see that when $0<\eta<\frac{\sqrt{5}-1}{4L}$, all the terms are positive.

\section{FedPD with Variance Reduction}\label{app:o2}
In this section we provided an alternative oracle for FedPD which has a lower sample complexity. 

\subsection{Algorithm description}

Alternatively, when instantiating the local oracle using \algref{alg:FedPD_v2}, the original local problems are not required to solve to $\epsilon_1$ accuracy. Instead, we successively optimize a linearized AL function: 
\[\Tilde{\cL}^r_i(\xx_i)
\bydef \Tilde{f}_i(\xx_i;\xx^{r,q}_i)+\lin{\lambda^i_r,\xx_i-\xx^r_{0,i}}+\frac{1}{2\eta}\norm{\xx_i-\xx^r_{0,i}}^2.\]
In the above expression, we linearize $f_i(\xx_i)$ at inner iteration $\xx^{r,q}_i$ as ($\gamma$ is a constant)
\begin{align*}
\Tilde{f}^r_i(\xx_i;\xx^{r,q}_i)
\bydef f(\xx^{r,q}_i)+\lin{g^{r,q}_i,\xx_i-\xx^r_i}+\frac{1}{2\gamma}\norm{\xx_i-\xx^{r,q}_i}^2,
\end{align*}
where $g^{r,q}_i$ is an approximation of $\nabla f_i(\xx^{r,q}_i)$. The optimizer has a closed-form expression:
\begin{equation*}
    \xx^{r,q+1}_i = \frac{\eta}{\eta+\gamma} \xx^{r,q}_i+\frac{\gamma}{\eta+\gamma}\xx^r_{0,i}-\frac{\eta\gamma}{\eta+\gamma}(g^{r,q}_i+\lambda^{r}_i).
\end{equation*}

\begin{minipage}[t]{0.59\textwidth}{\small
\begin{algorithm}[H]
  \begin{algorithmic}
		\STATE {\bfseries Input:} $\cL_i(\xx^r_i,\xx^r_{0,i},\lambda^r_i), Q, I, B$\\
		\STATE {Initialize: $\xx^{r,0}_i = \xx^r_i$,}\\
		{\bf if}\; {$r \mod I = 0$} {\bf then} $g^{r,0}_i= \nabla f_i(\xx^{r,0}_i)$\\
		{\bf else}  $g^{r,0}_i= g^{r-1,Q}_i$ \\
		{\bf end if}
		\FOR{$q=0,\dots,Q-1$}
			\STATE \hspace{-0.4cm}$\xx^{r,q+1}_i= \argmin_{\xx_i}\Tilde{\cL}_i(\xx_i,\xx^r_{0,i},\lambda^r_i;\xx^{r,q}_i, g^{r,q}_i)$\\
			\STATE \hspace{-0.6cm} $g^{r,q+1}_i= g^{r,q}_i+ \frac{1}{B}\sum^{B}_{b=1}(h_i(\xx^{r,q+1}_i;\xi^{r,q}_{i,b}) - h_i(\xx^{r,q}_i;\xi^{r,q}_{i,b}))$
		\ENDFOR
		\STATE {{\bfseries Output:} $\xx^{r+1}_i\triangleq \xx^{r,Q}_i$, $g^{r,Q}_i$}
	\end{algorithmic}
	\caption{Oracle Choice II}
	\label{alg:FedPD_v2}  
\end{algorithm}}
\end{minipage}

{\black In Oracle II, an agent $i$ first decides whether to compute the full gradient $\nabla f_i(\xx^{r,0}_i)$, or to keep using the previous estimate $g^{r-1,Q}_i$. Then $Q$ local steps are performed, 
each requires $B$ local data samples. In this scheme, $Q$ can be chosen as {\it any} positive integer.}

It is important to note that this oracle does not simply apply the VR technique (such as F-SVRG) to solve the subproblem of optimizing $\cL_i(\xx_i,\xx^{r}_{0,i},\lambda^r_i)$. That is, it is {\it not} a variation of Oracle I. Instead, the VR technique is applied to the entire primal-dual iteration, and the full gradient evaluation $\nabla f_i(\xx^{r,0}_i)$ is only needed every $I$ iteration $r$. Later we will see that if $I$ is large enough, then there is an $\cO(\sqrt{M})$ {\black reduction} of sample complexity.

\subsection{Algorithm Convergence and Complexity}
The convergence result of FedPD with Oracle II is given as follows:
\begin{theorem}\label{th:FedPD_VR}
    Suppose  \asref{as:smooth}--\asref{as:lower-bounded} hold, and consider FedPD  with Oracle II. Choose {\black $p=0$}, $\eta\in \big(0, \frac{1}{3(Q+\sqrt{QI/B})L}\big)$, and $\gamma > \frac{5\eta}{B\sqrt{L}}$. Then, the following holds (where $C_9>0$ is a constant):
    \begin{equation}
        \frac{1}{T}\sum^{T}_{r=0}\E\norm{\nabla f(\xx^r_0)}^2\leq \frac{C_9}{T}(f(\xx^0_0)-f(\xx^\star)).
    \end{equation}
\end{theorem}

\remark{{\bf (Communication complexity)}: As $p=0$, the communication round to achieve $\epsilon$ accuracy is $T=\cO(1/\epsilon)$ which is independent of $Q$.} 


\remark{{\bf (Computation complexity)}: {Note that the total number full gradient evaluation is $T/I+1$, each uses $M$ samples. Meanwhile,  the total number of mini-batch stochastic gradient evaluation is $TQ$, each uses $2B$ samples per node. So the total sample complexity is $\cO(M+MT/I+2TQBN)$.} In order to keep the same convergence speed, we need stepsize $\eta$ to be unchanged. Therefore, we choose $I=\sqrt{M}, B=I/QN=\sqrt{M}/QN$, then the SC of \algref{alg:FedPD_v2} is $\cO(M+\frac{\sqrt{M}}{\epsilon})$.}

\subsection{Proof of \thref{th:FedPD_VR}}

Following the similar proof of \thref{th:FedPD_Comm}, we first analyze the descent between each outer iteration. Notice throughout the proof, we assume that $p=0$, that is, there is no delayed communication. 
It follows that the following holds: 
$$\xx^{r+1}_{0,i}=\frac{1}{N}\sum^N_{j=1}\xx^{r+}_{0,j},\; \; \forall i=1,\dots,N.$$

We also recall that $r$ is the (outer) stage index, and $q$ is the local update index. First we provide a series of lemmas. 

\begin{lemma}\label{le:VR_descent}
Under Assumption~\ref{as:smooth}, consider FedPD with \algref{alg:FedPD_v2} (Oracle II) as the update rule. The difference of the local AL is bounded by:
\begin{equation}\label{eq:le_VR_descent}
    \begin{aligned}
        &\cL_i(\xx^{r+1}_i,\xx^{r+1}_{0,i},\lambda^{r+1}_i)-\cL_i(\xx^{r}_i,\xx^{r}_{0,i},\lambda^{r}_i)\nonumber\\
        &\leq -\frac{1}{2\eta}\norm{\xx^{r+1}_{0,i}-\xx^{r}_{0,i}}^2-\left(\frac{1}{2\eta}+\frac{1}{\gamma}-L-\frac{3\eta}{\gamma^2}\right)\norm{\xx^{r,Q}_i-\xx^{r,Q-1}_i}^2\\
        &-(\frac{1}{2\eta}+\frac{1}{\gamma}-L-9Q^2L^2\eta)\sum^{Q-1}_{q=1}\norm{\xx^{r,q}_i-\xx^{r,q-1}_i}^2\\
        & + \left(9Q^2L^2\eta+\frac{3\eta}{\gamma^2}\right)\norm{\xx^{r-1,Q}_i-\xx^{r-1,Q-1}_i}^2+\frac{1}{2L}\sum^{Q-2}_{q=0}\norm{\nabla f_i(\xx^{r,q}_i)-g^{r,q}_i}^2\\
        & +(\frac{1}{2L}+9\eta)\norm{g^{r,Q-1}_i-\nabla f_i(\xx^{r,Q-1}_i)}^2 + 9\eta \norm{g^{r-1,Q-1}_i-\nabla f_i(\xx^{r-1,Q-1}_i)}^2\\
        & +\lin{\lambda^{r+1}_i+\frac{1}{\eta}(\xx^{r+1}_i-\xx^{r+1}_{0,i}),\xx^{r+1}_{0,i}-\xx^{r}_{0,i}}.
    \end{aligned}
\end{equation}
\end{lemma}

Then we deal with the variance of the stochastic gradient estimations.
\begin{lemma}\label{le:VR_var}
{Suppose ~\asref{as:smooth} holds true and the samples are randomly sampled according to \eqref{eq:sample}, consider FedPD with \algref{alg:FedPD_v2} (Oracle II) as the update rule. The expected norm square of the difference between $g^{r,q+1}_i$ and $\nabla f_i(\xx^{r,q+1}_i)$ is bounded by}
    \begin{equation}
        \begin{aligned}
            \E\norm{g^{r,q+1}_i-\nabla f_i(\xx^{r,q+1}_i)}^2 \leq \frac{L^2}{B}\sum^{\{r,q+1\}}_{\tau=\{r_0,1\}} \E\norm{\xx^{\tau}_i-\xx^{\tau-1}_i}^2.
        \end{aligned}
    \end{equation}
\end{lemma}

Lastly we upper bound the original loss function.

\begin{lemma}\label{le:VR_gradient}
Under \asref{as:smooth} and \asref{as:lower-bounded}, the difference between the original loss and the AL is bounded as below:
    \begin{equation}
    \begin{aligned}
        \E f(\xx^r_0)\leq &~ \E\cL(\xx^{r}_0,\xx^{r}_1,\dots,\xx^{r}_N,\lambda^{r}_1,\dots,\lambda^{r}_N)-\frac{1-3L\eta}{2N\eta}\sum^N_{i=1}\E\norm{\xx^r_i-\xx^r_0}^2\\
        &+\frac{(1+L\gamma)^2+L^2\gamma^2}{4L\gamma^2}\left[\frac{1}{B}\sum^{\{r-1,Q-1\}}_{\tau=\{r_0,1\}} \E\norm{\xx^{\tau}_i-\xx^{\tau-1}_i}^2
        +\E\norm{\xx^{r-1,Q}_i-\xx^{r-1,Q-1}_i}^2\right].
    \end{aligned}
    \end{equation}
\end{lemma}

\subsubsection{Proof of \leref{le:VR_descent}}
Let us first express the difference of the local AL as following:
\begin{align}
&\cL_i(\xx^{r+1}_i,\xx^{r+1}_{0,i},\lambda^{r+1}_i)-\cL_i(\xx^{r}_i,\xx^{r}_{0,i},\lambda^{r}_i)\label{eq:L:difference}\\
& = \cL_i(\xx^{r+1}_i,\xx^{r}_{0,i},\lambda^{r}_i)-\cL_i(\xx^{r}_i,\xx^{r}_{0,i},\lambda^{r}_i) + \cL_i(\xx^{r+1}_i,\xx^{r}_{0,i},\lambda^{r+1}_i)-\cL_i(\xx^{r+1}_i,\xx^{r}_{0,i},\lambda^{r}_i) \nonumber\\
& \quad \quad + \cL_i(\xx^{r+1}_i,\xx^{r+1}_{0,i},\lambda^{r+1}_i)-\cL_i(\xx^{r+1}_i,\xx^{r}_{0,i},\lambda^{r+1}_i), \nonumber
\end{align}
where the above three differences respectively correspond to the three steps in the algorithm's update steps.

{Let us bound the above three differences one by one. First, note that we have the following decomposition (by using the fact that $\xx^{r,Q+1}_i = \xx^{r+1}_i$ and $\xx^{r,1}_i = \xx^{r}_i$):}
\begin{align}
\cL_i(\xx^{r+1}_i,\xx^{r}_{0,i},\lambda^{r}_i)-\cL_i(\xx^{r}_i,\xx^{r}_{0,i},\lambda^{r}_i) = \sum_{q=1}^{Q}\left(\cL_i(\xx^{r,q+1}_i,\xx^{r}_{0,i},\lambda^{r}_i)-\cL_i(\xx^{r,q}_i,\xx^{r}_{0,i},\lambda^{r}_i)\right).
\end{align}

Each term on the right hand side (RHS) of the above equality can be bounded by (see a similar arguments in \eqref{eq:le_PD_descent_1}):
\begin{equation}\label{eq:le_VR_descent_1}
    \begin{aligned}
        \cL_i(\xx^{r,q+1}_i,\xx^{r}_{0,i},\lambda^{r}_i)-&\cL_i(\xx^{r,q}_i,\xx^{r}_{0,i},\lambda^{r}_i)\leq \lin{\nabla f_i(\xx^{r,q}_i)+\lambda^r_i+\frac{1}{\eta}(\xx^{r,q+1}-\xx^r_{0,i}),\xx^{r,q+1}_i-\xx^{r,q}_i}\\
        &- \frac{1-L\eta}{2\eta}\norm{\xx^{r,q+1}_i-\xx^{r,q}_i}^2\\
        \stackrel{(a)}{=}& \lin{\nabla f_i(\xx^{r,q}_i)-g^{r,q}_i-\frac{1}{\gamma}(\xx^{r,q+1}-\xx^{r,q}_{i}),\xx^{r,q+1}_i-\xx^{r,q}_i}\\
        &- (\frac{1}{2\eta}-\frac{L}{2})\norm{\xx^{r,q+1}_i-\xx^{r,q}_i}^2\\
        = & ~\lin{\nabla f_i(\xx^{r,q}_i)-g^{r,q}_i,\xx^{r,q+1}_i-\xx^{r,q}_i}- (\frac{1}{2\eta}+\frac{1}{\gamma}-\frac{L}{2})\norm{\xx^{r,q+1}_i-\xx^{r,q}_i}^2\\
        \stackrel{(b)}{\leq} & ~\frac{1}{2L}\norm{\nabla f_i(\xx^{r,q}_i)-g^{r,q}_i}^2 - (\frac{1}{2\eta}+\frac{1}{\gamma}-L)\norm{\xx^{r,q+1}_i-\xx^{r,q}_i}^2,\\
    \end{aligned}
\end{equation}
where in $(a)$ we use the optimal condition that $\nabla_{\xx_i}\Tilde{\cL}_i(\xx^{r,q+1}_i,\xx^r_{0,i},\lambda^r_i;\xx^{r,q}_i, g^{r,q}_i)=0$ which gives us the following relation
\begin{equation}\label{eq:le_VR_descent_1_1}
    \lambda^{r}_i+\frac{1}{\eta}(\xx^{r,q+1}_i-\xx^r_{0,i})+g^{r,q}_i+\frac{1}{\gamma}(\xx^{r,q+1}_i-\xx^{r,q}_i)=0;
\end{equation}
in $(b)$ we use the fact that $2\lin{a,b}\leq L\norm{a}^2+\frac{1}{L}\norm{b}^2$. 
Therefore, {the first difference in the RHS of \eqref{eq:L:difference} is given by}
\begin{align}
&\cL_i(\xx^{r+1}_i,\xx^{r}_{0,i},\lambda^{r}_i)-\cL_i(\xx^{r}_i,\xx^{r}_{0,i},\lambda^{r}_i) \nonumber\\
&\leq \frac{1}{2L}\sum_{q=1}^{Q}\norm{\nabla f_i(\xx^{r,q}_i)-g^{r,q}_i}^2 - (\frac{1}{2\eta}+\frac{1}{\gamma}-L)\sum_{q=1}^{Q}\norm{\xx^{r,q+1}_i-\xx^{r,q}_i}^2.
\end{align}

The other two differences in \eqref{eq:L:difference} can be explicitly expressed as:
\begin{align}
        & \cL_i(\xx^{r+1}_i,\xx^{r}_{0,i},\lambda^{r+1}_i)-\cL_i(\xx^{r+1}_i,\xx^{r}_{0,i},\lambda^{r}_i)=\eta\norm{\lambda^{r+1}_i-\lambda^{r}_i}^2,\label{eq:le_VR_descent_2}\\
        & \cL_i(\xx^{r+1}_i,\xx^{r+1}_{0,i},\lambda^{r+1}_i)-\cL_i(\xx^{r+1}_i,\xx^{r}_{0,i},\lambda^{r+1}_i) \nonumber\\
        & = -\frac{1}{2\eta}\norm{\xx^{r+1}_{0,i}-\xx^{r}_{0,i}}^2+\lin{\lambda^{r+1}_i+\frac{1}{\eta}(\xx^{r+1}_i-\xx^{r+1}_{0,i}),\xx^{r+1}_{0,i}-\xx^{r}_{0,i}}. \label{eq:le_VR_descent_3}
\end{align}
Next we bound $\norm{\lambda^{r+1}_i-\lambda^{r}_i}^2$. Notice that the from the update rule the following holds: 
\begin{equation}\label{eq:le_VR_descent_3_1}
    \lambda^{r+1}_i = \lambda^r_i+\frac{1}{\eta}(\xx^{r,Q}_i-\xx^{r}_{0,i}) \stackrel{\eqref{eq:le_VR_descent_1_1}}{=} -\frac{1}{\gamma}(\xx^{r,Q}_i-\xx^{r,Q-1}_i)-g^{r,Q-1}.
\end{equation}
Using the above property, we have
\begin{equation}\label{eq:le_VR_descent_4}
    \begin{aligned}
        \norm{\lambda^{r+1}_i-\lambda^{r}_i}^2 =&~ \norm{\frac{1}{\gamma}(\xx^{r,Q}_i-\xx^{r,Q-1}_i)+g^{r,Q-1}_i-\frac{1}{\gamma}(\xx^{r-1,Q}_i-\xx^{r-1,Q-1}_i)-g^{r-1,Q-1}_i}^2\\
        \stackrel{(a)}{\leq}&~ 3\norm{g^{r,Q-1}_i-g^{r-1,Q-1}_i}^2 + \frac{3}{\gamma^2}\norm{\xx^{r,Q}_i-\xx^{r,Q-1}_i}^2+ \frac{3}{\gamma^2}\norm{\xx^{r-1,Q}_i-\xx^{r-1,Q-1}_i}^2.
    \end{aligned}
\end{equation}
where in $(a)$ we apply Cauchy-Schwarz inequality. Next we bound $\norm{g^{r,Q-1}_i-g^{r-1,Q-1}_i}^2$ by 

\begin{equation}\label{eq:le_VR_descent_5}
    \begin{aligned}
        &\norm{g^{r,Q-1}_i-g^{r-1,Q-1}_i}^2 \nonumber\\
        & =~ \norm{g^{r,Q-1}_i-\nabla f_i(\xx^{r,Q-1}_i)+ \nabla f_i(\xx^{r,Q-1}_i)-\nabla f_i(\xx^{r-1,Q-1}_i)+ \nabla f_i(\xx^{r-1,Q-1}_i) -g^{r-1,Q-1}_i}^2\\
        &\stackrel{(a)}{\leq}~ 3\norm{g^{r,Q-1}_i-\nabla f_i(\xx^{r,Q-1}_i)}^2 + 3\norm{g^{r-1,Q-1}_i-\nabla f_i(\xx^{r-1,Q-1}_i)}^2+ 3L^2\norm{\xx^{r,Q-1}_i-\xx^{r-1,Q-1}_i}^2\\
        &\stackrel{(b)}{\leq}~ 3\norm{g^{r,Q-1}_i-\nabla f_i(\xx^{r,Q-1}_i)}^2 + 3\norm{g^{r-1,Q-1}_i-\nabla f_i(\xx^{r-1,Q-1}_i)}^2\\
        &+ 3Q^2L^2\sum^{Q-1}_{q=1}\norm{\xx^{r,q}_i-\xx^{r,q-1}_i}^2+ 3Q^2L^2\norm{\xx^{r-1,Q}_i-\xx^{r-1,Q-1}_i}^2,\\
    \end{aligned}
\end{equation}
where in $(a)$ and $(b)$ we both apply Cauchy-Schwarz inequality, in $(a)$ we use \asref{as:smooth} to the last term and in $(b)$ we notice $\xx^{r-1,Q}_i=\xx^{r,0}_i$.

Substitute \eqref{eq:le_VR_descent_5} to \eqref{eq:le_VR_descent_4} and sum the three parts, we have

\begin{equation}
    \begin{aligned}
        &\cL_i(\xx^{r+1}_i,\xx^{r+1}_{0,i},\lambda^{r+1}_i)-\cL_i(\xx^{r}_i,\xx^{r}_{0,i},\lambda^{r}_i)\nonumber\\
        &\leq -\frac{1}{2\eta}\norm{\xx^{r+1}_{0,i}-\xx^{r}_{0,i}}^2-(\frac{1}{2\eta}+\frac{1}{\gamma}-L-\frac{3\eta}{\gamma^2})\norm{\xx^{r,Q}_i-\xx^{r,Q-1}_i}^2\\
        &-(\frac{1}{2\eta}+\frac{1}{\gamma}-L-9Q^2L^2\eta)\sum^{Q-1}_{q=1}\norm{\xx^{r,q}_i-\xx^{r,q-1}_i}^2\\
        & + (9Q^2L^2\eta+\frac{3\eta}{\gamma^2})\norm{\xx^{r-1,Q}_i-\xx^{r-1,Q-1}_i}^2+\frac{1}{2L}\sum^{Q-2}_{q=0}\norm{\nabla f_i(\xx^{r,q}_i)-g^{r,q}_i}^2\\
        & +(\frac{1}{2L}+9\eta)\norm{g^{r,Q-1}_i-\nabla f_i(\xx^{r,Q-1}_i)}^2 + 9\eta \norm{g^{r-1,Q-1}_i-\nabla f_i(\xx^{r-1,Q-1}_i)}^2\\
        & +\lin{\lambda^{r+1}_i+\frac{1}{\eta}(\xx^{r+1}_i-\xx^{r+1}_{0,i}),\xx^{r+1}_{0,i}-\xx^{r}_{0,i}},
    \end{aligned}
\end{equation}
which complete the proof of \leref{le:VR_descent}.

\subsubsection{Proof of \leref{le:VR_var}}

To study $\E\norm{g^{r,q}_i-\nabla f_i(\xx^{r,q}_i)}^2$, we denote the latest iteration before $r$ that computes full gradients as $r_0$.  That is, in $r_0$ we have $g^{r_0,0}_i=\nabla f_i(\xx^{r_0,0}_i)$. 
By the description of the algorithm we know 
$$r_0=kI, \; \; k\in \mathbb{N}, \; \; rQ+q-r_0Q\leq IQ.$$ 
{That is, $r_0$ is a multiple of $I$ and there is no more than $IQ$ local update steps between step $\{r_0,0\}$ and step $\{r,q\}$.} By the update rule of $g^{r,q}_i$, we have
\begin{equation}\label{eq:le_VR_var_0}
    \begin{aligned}
        g^{r,q+1}_i-\nabla f_i(\xx^{r,q+1}_i) =& g^{r,q}_i-\nabla f_i(\xx^{r,q+1}_i)+\frac{1}{B}\sum^{B}_{b=1}(h_i(\xx^{r,q+1}_i;\xi^{r,q}_{i,b}) - h_i(\xx^{r,q}_i;\xi^{r,q}_{i,b})).
    \end{aligned}
\end{equation}
{Take expectation on both sides, we have 
\begin{equation}\label{eq:le_VR_var_0_1}
    \begin{aligned}
        &\E_{\{\xi^{r,q}_{i,b}\}^B_{b=1}} [g^{r,q+1}_i-\nabla f_i(\xx^{r,q+1}_i)] \nonumber\\
        & = g^{r,q}_i-\nabla f_i(\xx^{r,q+1}_i)+ \E_{\{\xi^{r,q}_{i,b}\}^B_{b=1}}[\frac{1}{B}\sum^{B}_{b=1}(h_i(\xx^{r,q+1}_i;\xi^{r,q}_{i,b}) - h_i(\xx^{r,q}_i;\xi^{r,q}_{i,b}))]\\
        &=~ g^{r,q}_i-\nabla f_i(\xx^{r,q+1}_i)+ \nabla f_i(\xx^{r,q+1}_i) -\nabla f_i(\xx^{r,q}_i)\\
        &=~ g^{r,q}_i -\nabla f_i(\xx^{r,q}_i).
    \end{aligned}
\end{equation}
}

By using the fact that $\E[X^2]=[\E X]^2+\E[[X-\E X]^2]$ and substitute \eqref{eq:le_VR_var_0_1} we have
\begin{equation}\label{eq:le_VR_var_1}
    \begin{aligned}
        & \E_{\{\xi^{r,q}_{i,b}\}^B_{b=1}}\norm{g^{r,q+1}_i-\nabla f_i(\xx^{r,q+1}_i)}^2 \nonumber\\
        & = \norm{\E_{\{\xi^{r,q}_{i,b}\}^B_{b=1}}[g^{r,q+1}_i-\nabla f_i(\xx^{r,q+1}_i)]}^2\\
        &\quad + \E_{\{\xi^{r,q}_{i,b}\}^B_{b=1}}\norm{g^{r,q+1}_i-\nabla f_i(\xx^{r,q+1}_i)-\E_{\{\xi^{r,q}_{i,b}\}^B_{b=1}}[g^{r,q+1}_i-\nabla f_i(\xx^{r,q+1}_i)]}^2\\
        & \stackrel{\eqref{eq:le_VR_var_0_1}}{=} \norm{g^{r,q}_i-\nabla f_i(\xx^{r,q}_i)}^2 \\
        &\quad+ \E_{\{\xi^{r,q}_{i,b}\}^B_{b=1}}\norm{\frac{1}{B}\sum^{B}_{b=1}(h_i(\xx^{r,q+1}_i;\xi^{r,q}_{i,b} - h_i(\xx^{r,q}_i;\xi^{r,q}_{i,b})) -\nabla f_i(\xx^{r,q+1}_i) +\nabla f_i(\xx^{r,q}_i)}^2\\
        & \stackrel{(a)}{\leq}\norm{g^{r,q}_i-\nabla f_i(\xx^{r,q}_i)}^2 +\frac{1}{B^2}\sum^{B}_{b=1}\E_{\{\xi^{r,q}_{i,b}\}^B_{b=1}}\norm{h_i(\xx^{r,q+1}_i;\xi^{r,q}_{i,b}) - h_i(\xx^{r,q}_i;\xi^{r,q}_{i,b}))}^2\\
        & \stackrel{(b)}{\leq}\norm{g^{r,q}_i-\nabla f_i(\xx^{r,q}_i)}^2 +\frac{L^2}{B}\norm{\xx^{r,q+1}_i-\xx^{r,q}_i}^2,
    \end{aligned}
\end{equation}
where $(a)$ comes form the fact that we view $h_i(\xx^{r,q+1}_i;\xi^{r,q}_{i,b}) - h_i(\xx^{r,q}_i;\xi^{r,q}_{i,b})$ as $X$ and by identically random sampling strategy we have $\E X = \nabla f_i(\xx^{r,q+1}_i) -\nabla f_i(\xx^{r,q}_i)$ and $\E[[X-\E X]^2\leq \E[X]^2$, in $(b)$ we use \asref{as:smooth}.

Iteratively taking expectation until $\{r,q\}=\{r_0,0\}$, we have

\begin{equation}\label{eq:le_VR_var_2}
    \begin{aligned}
        \E\norm{g^{r,q+1}_i-\nabla f_i(\xx^{r,q+1}_i)}^2 \leq \frac{L^2}{B}\sum^{\{r,q+1\}}_{\tau=\{r_0,1\}} \E\norm{\xx^{\tau}_i-\xx^{\tau-1}_i}^2,
    \end{aligned}
\end{equation}

which completes the proof.

\subsubsection{Proof of \leref{le:VR_gradient}}

Applying \asref{as:smooth}, we have

\begin{equation}\label{eq:le_VR_gradient_1}
    \begin{aligned}
        f_i(\xx^{r}_0)\leq &~ f_i(\xx^r_i)+\lin{\nabla f_i(\xx^r_i),\xx^r_0-\xx^r_i}+\frac{L}{2}\norm{\xx^r_0-\xx^r_i}^2\\
        = &~ \cL_i(\xx^r_i, \xx^r_0, \lambda^r_i) - \lin{\nabla f_i(\xx^r_i)+\lambda^r_i, \xx^r_0-\xx^r_i} - \frac{1-L\eta}{2\eta}\norm{\xx^r_0-\xx^r_i}^2\\
        \leq &~ \cL_i(\xx^r_i, \xx^r_0, \lambda^r_i) + \frac{1}{4L}\norm{\nabla f_i(\xx^r_i)+\lambda^r_i}^2 - \frac{1-3L\eta}{2\eta}\norm{\xx^r_0-\xx^r_i}^2.\\
    \end{aligned}
\end{equation}
Then notice $\xx^r_i=\xx^{r-1,Q}_i$ and apply \eqref{eq:le_VR_descent_3_1}, we can bound $\E\norm{\nabla f_i(\xx^r_i)+\lambda^r_i}^2$ by the following:
\begin{equation}\label{eq:le_VR_gradient_2}
    \begin{aligned}
        \E\norm{\nabla f_i(\xx^r_i)+\lambda^r_i}^2 \stackrel{\eqref{eq:le_VR_descent_3_1}}{=}&~ \E\norm{\nabla f_i(\xx^{r-1,Q}_i)-g^{r-1,Q-1}_i-\frac{1}{\gamma}(\xx^{r-1,Q}_i-\xx^{r-1,Q-1}_i)}^2\\
        \stackrel{(a)}{\leq}&~ (1+\frac{(1+L\gamma)^2}{L^2\gamma^2})\E\norm{\nabla f_i(\xx^{r-1,Q-1}_i)-g^{r-1,Q-1}_i}^2\\
        & +(1+\frac{L^2\gamma^2}{(1+L\gamma)^2})(1+\frac{1}{L\gamma})\E\norm{\nabla f_i(\xx^{r-1,Q}_i)-\nabla f_i(\xx^{r-1,Q-1}_i)}^2\\
        &+\frac{(1+\frac{L^2\gamma^2}{(1+L\gamma)^2})(1+L\gamma)}{\gamma^2}\E\norm{\xx^{r-1,Q}_i-\xx^{r-1,Q-1}_i}^2\\
        \stackrel{(b)}{\leq}&~ \frac{(1+L\gamma)^2+L^2\gamma^2}{B\gamma^2}\sum^{\{r-1,Q-1\}}_{\tau=\{r_0,1\}} \E\norm{\xx^{\tau}_i-\xx^{\tau-1}_i}^2\\
        &+(1+\frac{L^2\gamma^2}{(1+L\gamma)^2})\left((1+\frac{1}{L\gamma})L^2+\frac{1+L\gamma}{\gamma^2}\right)\E\norm{\xx^{r-1,Q}_i-\xx^{r-1,Q-1}_i}^2\\
        =&~ \frac{(1+L\gamma)^2+L^2\gamma^2}{B\gamma^2}\sum^{\{r-1,Q-1\}}_{\tau=\{r_0,1\}} \E\norm{\xx^{\tau}_i-\xx^{\tau-1}_i}^2\\
        &+\frac{(1+L\gamma)^2+L^2\gamma^2}{\gamma^2}\E\norm{\xx^{r-1,Q}_i-\xx^{r-1,Q-1}_i}^2,\\
    \end{aligned}
\end{equation}
where in $(a)$ we apply Cauchy-Schwarz inequality twice,  that is $$\norm{x+y+z}^2\leq(1+\frac{1}{a})\norm{x}^2+(1+a)\norm{y+z}^2\leq(1+\frac{1}{a})\norm{x}^2+(1+a)(1+b)\norm{y}^2+(1+a)(1+\frac{1}{b})\norm{z}^2;$$ 
in $(b)$ we apply \leref{le:VR_var} to the first term and apply \asref{as:smooth} to the second term. 

Substitute \eqref{eq:le_VR_gradient_2} to \eqref{eq:le_VR_gradient_1} and average over the agents, \leref{le:VR_gradient} is proved. 

\subsubsection{Proof of \thref{th:FedPD_VR}}

By the update step of $\xx^r_0$,  following~\eqref{eq:th_comm_2_1} we have \[\norm{\frac{1}{N}\sum^N_{i=1}\nabla_{\xx_{0,i}}\cL_i(\xx^r_i,\xx^{r}_{0,i},\lambda^r_i)}=\norm{\frac{1}{N}\sum^N_{i=1}(\frac{1}{\eta}(\xx^r_i-\xx^{r}_{0,i})+\lambda^r_i)}=0.\] 
We also have
\begin{equation}\label{eq:th_VR_0}
    \begin{aligned}
        &\norm{\nabla\cL_i(\xx^r_i,\xx^{r}_{0,i},\lambda^r_i)}^2 \nonumber\\
        &=~ \norm{\nabla_{\xx_i}\cL_i(\xx^r_i,\xx^{r}_{0,i},\lambda^r_i)}^2 + \norm{\nabla_{\lambda_i}\cL_i(\xx^r_i,\xx^{r}_{0,i},\lambda^r_i)}^2\\
         &=~ \norm{\nabla f_i(\xx^r_i)+\lambda^r_i +\frac{1}{\eta}(\xx^r_i-\xx^r_{0,i})}^2 + \norm{\xx^r_i-\xx^r_{0,i}}^2 \\
         &\stackrel{(a)}{=}~ \norm{\nabla f_i(\xx^r_i)-g^{r,0}_i-\frac{\eta+\gamma}{\eta\gamma}(\xx^{r,1}_i-\xx^{r}_i)}^2 + \norm{\xx^r_i-\xx^r_{0,i}+\xx^{r-1}_{0,i}-\xx^{r-1}_{0,i}}^2\\
         &\leq~ \norm{\nabla f_i(\xx^r_i)-g^{r,0}_i-\frac{\eta+\gamma}{\eta\gamma}(\xx^{r,1}_i-\xx^{r}_i)}^2 + 2\norm{\xx^r_i-\xx^{r-1}_{0,i}}^2 +2\norm{\xx^{r}_{0,i}-\xx^{r-1}_{0,i}}^2\\
         &\leq~ 2\norm{\nabla f_i(\xx^r_i)-g^{r,0}_i}^2+2(\frac{\eta+\gamma}{\eta\gamma})^2\norm{\xx^{r,1}_i-\xx^{r}_i}^2 + 2\eta^2\norm{\lambda^r_i-\lambda^{r-1}_i}^2+2\norm{\xx^{r}_{0,i}-\xx^{r-1}_{0,i}}^2.
    \end{aligned}
\end{equation}
 where in $(a)$, the first term is obtained by plugging in \eqref{eq:le_VR_descent_3_1} given below $$\lambda^r_i = -g^{r,0}_i -\frac{1}{\gamma}(\xx^{r,1}_i-\xx^{r}_i)-\frac{1}{\eta}(\xx^{r,1}_i-\xx^{r}_{0,i}).$$

{Next we take expectation and substitute \eqref{eq:le_VR_descent_4}, \eqref{eq:le_VR_descent_5},
\begin{equation}\label{eq:th_VR_0_1}
    \begin{aligned}
        &\E\norm{\nabla\cL_i(\xx^r_i,\xx^{r}_{0,i},\lambda^r_i)}^2 \nonumber\\
        &\leq 2\E\norm{\nabla f_i(\xx^r_i)-g^{r,0}_i}^2 +2(\frac{\eta+\gamma}{\eta\gamma})^2\E\norm{\xx^{r,1}_i-\xx^{r}_i}^2+ 2\E\norm{\xx^{r}_{0,i}-\xx^{r-1}_{0,i}}^2\\
        &+ \frac{6\eta^2}{\gamma^2}(\gamma^2\E\norm{g^{r-1,Q-1}_i-g^{r-2,Q-1}_i}^2 + \E\norm{\xx^{r-1,Q}_i-\xx^{r-1,Q-1}_i}^2+ E\norm{\xx^{r-2,Q}_i-\xx^{r-2,Q-1}_i}^2)\\
        & \stackrel{(a)}{\leq} \frac{2L^2}{B}\sum^{\{r,0\}}_{\tau =\{r_0,1\}}\E\norm{\xx^\tau_i-\xx^{\tau-1}_i}^2 + 2(\frac{\eta+\gamma}{\eta\gamma})^2\E\norm{\xx^{r,1}_i-\xx^{r}_i}^2 + 2\E\norm{\xx^{r}_{0,i}-\xx^{r-1}_{0,i}}^2 \\
        &+ \frac{6\eta^2}{\gamma^2}(\E\norm{\xx^{r-1,Q}_i-\xx^{r-1,Q-1}_i}^2+ \E\norm{\xx^{r-2,Q}_i-\xx^{r-2,Q-1}_i}^2)\\
        &+ 18\eta^2\left(\E\norm{g^{r-1,Q-1}_i-\nabla f_i(\xx^{r-1,Q-1}_i)}^2 + \E\norm{g^{r-2,Q-1}_i-\nabla f_i(\xx^{r-2,Q-1}_i)}^2\right)\\
        &+ 18\eta^2Q^2L^2\left(\sum^{Q-1}_{q=1}\E\norm{\xx^{r-1,q}_i-\xx^{r-1,q-1}_i}^2+ \E\norm{\xx^{r-2,Q}_i-\xx^{r-2,Q-1}_i}^2\right),
    \end{aligned}
\end{equation}
where we substitute \leref{le:VR_var} and \eqref{eq:le_VR_descent_5} in $(a)$.}

Taking expectation of \eqref{eq:le_VR_descent}, summing over $r=0$ to $r=T-1$ and average over the agents, we have the following
\begin{equation}
    \begin{aligned}\label{eq:th_VR_1}
        \frac{1}{N}\sum^N_{i=1}\E[\cL_i(\xx^{T}_i,\xx^{T}_{0,i},\lambda^{T}_i)&-\cL_i(\xx^{0}_i,\xx^{0}_{0,i},\lambda^{0}_i)]\leq -\frac{1}{2\eta}\sum^{T-1}_{r=0}\E\norm{\xx^{r+1}_{0}-\xx^{r}_{0}}^2\\
        & -(\frac{1}{2\eta}+\frac{1}{\gamma}-L-\frac{6\eta}{\gamma^2}-9Q^2L^2\eta)\frac{1}{N}\sum^{N}_{i=1}\sum^{Q-1}_{q=0}\sum^{T-1}_{r=0}\E\norm{\xx^{r,q+1}_i-\xx^{r,q-1}_i}^2\\
        &+ (\frac{1}{2L}+18\eta)\frac{1}{N}\sum^{N}_{i=1}\sum^{T-1}_{r=0}\sum^{Q-1}_{q=0}\E\norm{\nabla f_i(\xx^{r,q}_i)-g^{r,q}_i}^2\\
        &+\sum^{T-1}_{r=0}\frac{1}{N}\E\lin{\sum^{N}_{i=1}(\lambda^{r+1}_i+\frac{1}{\eta}(\xx^{r+1}_i-\xx^{r+1}_{0,i})),\xx^{r+1}_{0,i}-\xx^{r}_{0,i}}\\
        \stackrel{(a)}{\leq} &~ -(\frac{1}{2\eta}+\frac{1}{\gamma}-L-\frac{6\eta}{\gamma^2}-9Q^2L^2\eta)\frac{1}{N}\sum^{N}_{i=1}\sum^{Q-1}_{q=0}\sum^{T-1}_{r=0}\E\norm{\xx^{r,q+1}_i-\xx^{r,q-1}_i}^2\\
        & - \frac{1}{2\eta}\sum^{T-1}_{r=0}\E\norm{\xx^{r+1}_{0}-\xx^{r}_{0}}^2\\
        & + \frac{(1+18L\eta)LIQ}{2B}\frac{1}{N}\sum^{N}_{i=1}\sum^{T-1}_{r=0}\sum^{Q-1}_{q=0}\E\norm{\xx^{r,q+1}_i-\xx^{r,q-1}_i}^2\\
        = &~ -\frac{C_{10}}{N}\sum^{N}_{i=1}\sum^{Q-1}_{q=0}\sum^{T-1}_{r=0}\E\norm{\xx^{r,q+1}_i-\xx^{r,q-1}_i}^2-\frac{1}{2\eta}\sum^{T-1}_{r=0}\E\norm{\xx^{r+1}_{0}-\xx^{r}_{0}}^2,
    \end{aligned}
\end{equation}
where in $(a)$ we apply \leref{le:VR_var} and \eqref{eq:th_comm_2_1}.

Finally, in the last equation of \eqref{eq:th_VR_1}, we have defined the constant $C_{10}$ as 
$$C_{10}:=\frac{1}{2\eta}+\frac{1}{\gamma}-L-\frac{6\eta}{\gamma^2}-9Q^2L^2\eta-\frac{(1+18L\eta)LIQ}{2B}.$$ 

Then by taking expectation and applying \leref{le:VR_gradient}, we obtain
\begin{equation}
    \begin{aligned}\label{eq:th_VR_2}
        &\E[f(\xx^T_0)-f(\xx^0_0)]\nonumber\\
        &\leq ~ -\frac{C_{10}-\frac{(1+L\gamma)^2+L^2\gamma^2}{4BL\gamma^2}}{N}\sum^{N}_{i=1}\sum^{Q-1}_{q=0}\sum^{T-1}_{r=0}\E\norm{\xx^{r,q+1}_i-\xx^{r,q-1}_i}^2-\frac{1}{2\eta}\sum^{T-1}_{r=0}\E\norm{\xx^{r+1}_{0}-\xx^{r}_{0}}^2,
    \end{aligned}
\end{equation}
where by the initialization that $\xx^0_i = \xx^0_0$ we have $f_(\xx^0_0)=\frac{1}{N}\sum^N_{i=1}\cL_i(\xx^{0}_i,\xx^{0}_{0,i},\lambda^{0}_i).$

Combine \eqref{eq:th_VR_0_1} and \eqref{eq:th_VR_2}, we can find a positive constant $C_{11}$ satisfying 
{\black \begin{align}
    C_{11} \leq\min\bigg\{C_{12}/C_{13},1/(4\eta)\bigg\},
\end{align}
where we have defined
\begin{align}
    C_{12}&\triangleq {C_{10}-\frac{(1+L\gamma)^2+L^2\gamma^2}{4BL\gamma^2}},\nonumber\\
    C_{13}&\triangleq Q\bigg(2(\frac{\eta+\gamma}{\eta\gamma})^2 +\frac{2I(1+18\eta^2)L^2}{B}+\frac{3L(1+9L\eta)\eta^2}{2B\gamma^2}+18Q^2L^2\eta^2\bigg)
\end{align}}

so that the following holds
\begin{equation}
    \begin{aligned}\label{eq:th_VR_3}
        \frac{C_{11}}{NT}\sum^{T}_{r=0}\sum^N_{i=1}\E\norm{\nabla\cL_i(\xx^r_i,\xx^{r}_{0,i},\lambda^r_i)}^2\leq&~\frac{C_{10}-\frac{(1+L\gamma)^2+L^2\gamma^2}{4BL\gamma^2}}{NT}\sum^{N}_{i=1}\sum^{Q-1}_{q=0}\sum^{T-1}_{r=0}\E\norm{\xx^{r,q+1}_i-\xx^{r,q-1}_i}^2\\
        &+\frac{1}{2\eta T}\sum^{T-1}_{r=0}\E\norm{\xx^{r+1}_{0}-\xx^{r}_{0}}^2\\
        \leq&~ \frac{1}{T}(f(\xx^0_0)-\E f(\xx^T_0))\leq \frac{1}{T}(f(\xx^0_0)-f(\xx^\star)).
    \end{aligned}
\end{equation}

Similar to the proof of \thref{th:FedPD_Comm}, we can bound $\norm{\nabla f(\xx^r_0)}^2$ by $\frac{1}{N}\sum^{N}_{i=1}\norm{\nabla \cL_i(\xx^r_i,\xx^r_0,\lambda^r_i)}^2,$ therefore \thref{th:FedPD_VR} is proved.

During the prove we need
$$C_9 = 4L^2/C_{11},\;\quad C_{10}=\frac{1}{2\eta}+\frac{1}{\gamma}-L-\frac{6\eta}{\gamma^2}-9Q^2L^2\eta-\frac{(1+18L\eta)LIQ}{2B},$$ $$C_{11}\leq\min\left\{\frac{\left({C_{10}-\frac{(1+L\gamma)^2+L^2\gamma^2}{4BL\gamma^2}}\right)}{Q\left(2(\frac{\eta+\gamma}{\eta\gamma})^2+\frac{2I(1+18\eta^2)L^2}{B}+\frac{3L(1+9L\eta)\eta^2}{2B\gamma^2}+18Q^2L^2\eta^2\right)},\frac{1}{4\eta}\right\}$$
to be positive constant. By selecting $\gamma>\frac{5}{B\sqrt{L}}\eta$, and $0<\eta<\frac{1}{3(Q+\sqrt{QI/B})L}$, this is guaranteed.

\section{Additional Numerical Results}\label{app:numerical}

\subsection{Penalized Logistic Regression}

In this experiment, we consider the penalized regression problem \cite{antoniadis2011penalized}, whose loss function evaluated on a single sample $(\aa,b)=\xi$ is given by: \begin{equation}
    F(\xx;(\aa, b)) = \log(1+\exp(-b\xx^T\aa)) + \sum^{D}_{d=1}\frac{\beta\alpha(\xx[d])^2}{1+\alpha(\xx[d])^2}.
\end{equation}
Here $\xx[d]$ denotes the $d^{th}$ component of $\xx$. The feature vector and model parameter $\aa, \xx\in \mathbb{R}^D$ have dimension $D$ and $b\in\{-1,1\}$ is the label corresponding to the feature. During the simulation, we set the constants to be $\alpha = 1$ and $\beta = 0.1$. 

In the experiment, we use two ways to generate the data. In the first case (referred to as the ``weakly non-i.i.d" case), the features and the labels on the agents are randomly generated, so the local data sets are not very  non-i.i.d. {In the second case (referred to as the ``strong non-i.i.d." case), we first generate the feature vector $\mathbf{a}$'s  following the standard Normal distribution, then we generate the local model $\xx_i$ on the $i^{th}$ agent by using uniform distribution in the range of $[-10,10]$ for each component. Then we compute the label $b$'s according to the local models and the features and add some uniform noise. In this case, the data distribution on the agents are more non-i.i.d. compared to the first case.} In both cases, there are $400$ samples on each agent with total $100$ agents.

The total number of iterations $T$ is set as $600$ for all algorithms. We choose the stepsize to be $\eta = 4$ for FedAvg-GD with local update number $Q=8$ and for FedAvg-SGD we use diminishing stepsize $\eta = 4/\sqrt{Qr+q+1}$ with $Q=600$. For FedProx we use VR algorithm as the local solver and set $Q=8$, $\rho = 1$ and stepsize $\eta = 4$. For FedPD, we also use the same stepsize $\eta = 4$ with $Q=8$ with local GD. For FedPD-SGD,  we also set $\eta=4$ and uses local step size $\eta_1 = \frac{1}{Q}$ with inner iteration number $Q=600$. Lastly for FedPD with VR, we set the parameters to be $\eta = 4$, $\gamma = 4$, $I=100$, $Q=2$ and $B=1$. The choice of the stepsize is the same among all the algorithms. We used grid search on stepsizes $\eta\in\{5, 2, 1, 0.1, 0.01\}$ and the relative performance of the algorithms are similar to what we will show shortly. 

\begin{figure}[h]
    \centering
    \begin{subfigure}[t]{0.40\linewidth}
        \centering
        \includegraphics[width=\linewidth]{Matlab_v5.eps}
        \caption{The stationary gap of FedAvg, FedProx and FedPD with respect to the number of communication rounds.}
    \end{subfigure}
    \hfill
    \begin{subfigure}[t]{0.40\linewidth}
        \centering
        \includegraphics[width=\linewidth]{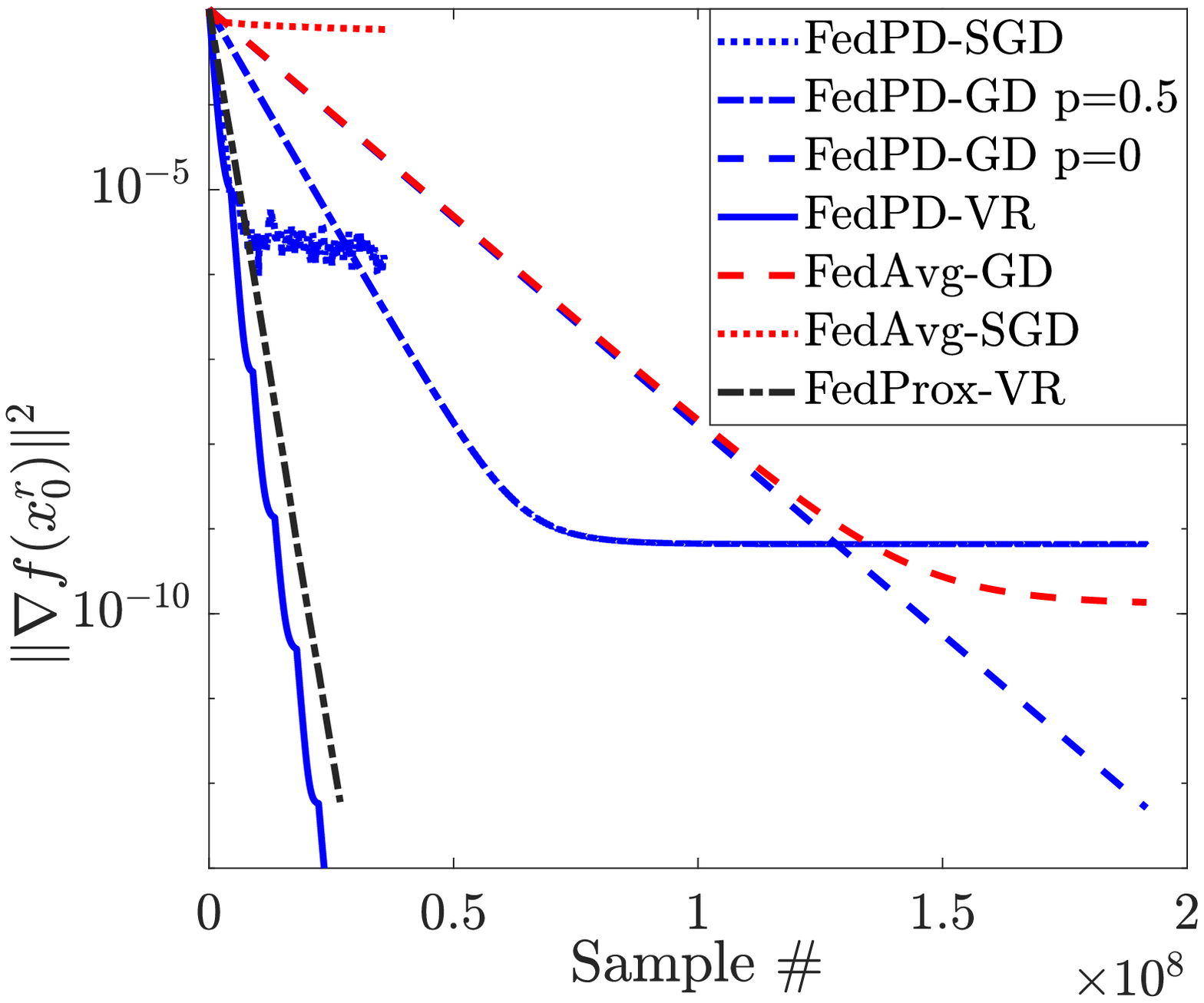}
        \caption{The stationary gap of of FedAvg, FedProx and FedPD with respect to the number of samples.}
    \end{subfigure}
    \caption{The convergence result of the algorithms on penalized logistic regression with weakly non-i.i.d. data.}
    \label{fig:exp_logit_app}
    \vspace{0.5cm}
\end{figure}

\begin{figure}[htb]
    \centering
    \begin{subfigure}[t]{0.40\linewidth}
        \centering
        \includegraphics[width=\linewidth]{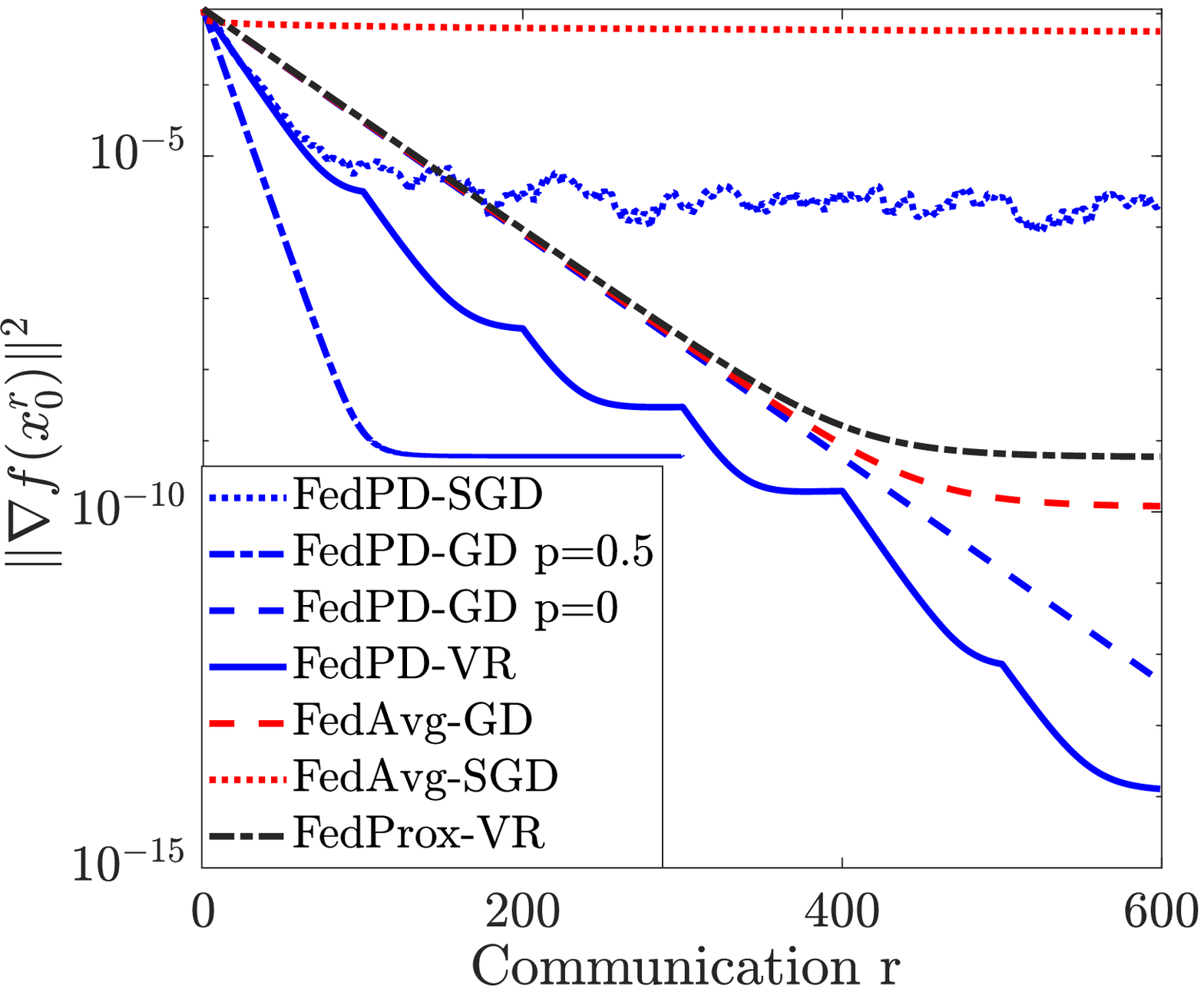}
        \caption{The stationary gap of FedAvg, FedProx and FedPD with respect to the number of communication rounds.}
    \end{subfigure}
    \hfill
    \begin{subfigure}[t]{0.40\linewidth}
        \centering
        \includegraphics[width=\linewidth]{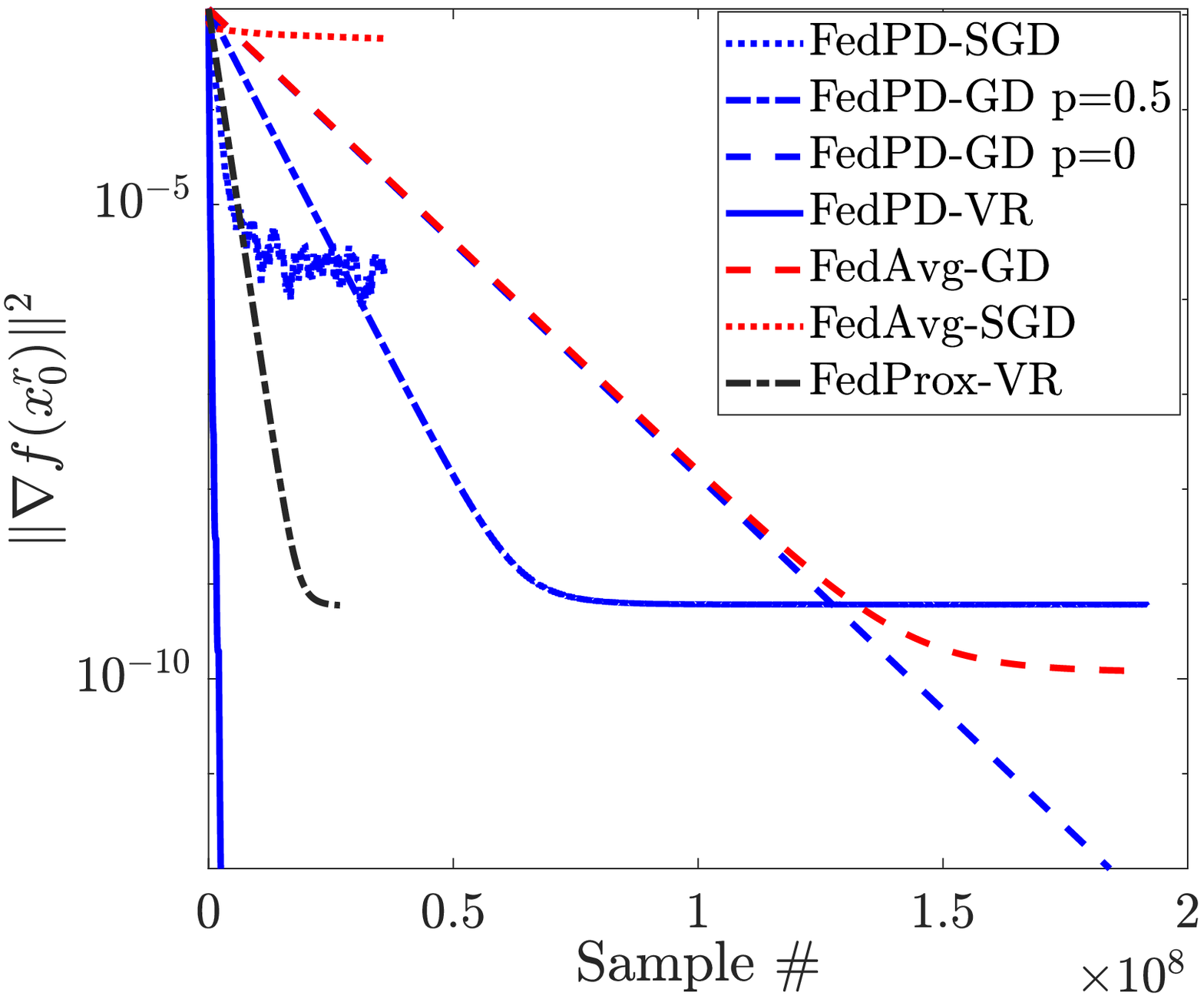}
        \caption{The stationary gap of of FedAvg, FedProx and FedPD with respect to the number of samples.}
    \end{subfigure}
    \caption{The convergence result of the algorithms on penalized logistic regression with strongly non-i.i.d. data.}
    \label{fig:exp_logit_app_1}
\end{figure}

Fig.~\ref{fig:exp_logit_app} shows the convergence results of the penalized logistic regression problem with the first data set. In Fig.~\ref{fig:exp_logit_app}(a), we compare the convergence of the tested algorithms w.r.t the communication rounds. It is clear that FedProx and  FedPD with $R=1$ (i.e., no communication skipping) are comparable. 
Meanwhile, FedAvg with local GD will not converge to the stationary point with a constant stepsize when local update step $Q>1$. By skipping half of the communication, FedPD with local GD can still achieve a similar error as FedAvg, but using fewer communication rounds. In Fig.~\ref{fig:exp_logit_app}(b), we compare the sample complexity of different algorithms. It can be shown that when using the same number of samples for computation, FedPD with Oracle II (FedPD-VR) converges the fastest among all the algorithms. FedProx uses VR to solve the inner problem and converges the second fastest. 
Fig~\ref{fig:exp_logit_app_1} shows the convergence results with the strongly non-i.i.d. data set. We can see that the algorithms using stochastic solvers become less stable compared with the case when the data sets are weakly non-i.i.d. Further, FedPD-VR and FedPD-GD with $R=1$ are able still to converge to the global stationary point while FedProx will achieve a similar error as the FedAvg with local GD.

\subsection{Handwritten Character Classification}

In the second experiment, we compare FedPD with FedAvg and FedProx on the FEMNIST data set~\cite{caldas2018leaf}. The FEMNIST data set collects the handwritten characters, including numbers 1--10 and the upper- and lower-case letters A--Z and a--z, from different writers and is separated by the writers, therefore the data set naturally preserves non-i.i.d-ness.

The entire data set contains 805,000 samples collected from 3,550 writers. In our experiments, we use the data collected from 100 writers with an average of 300 samples per writer and the size of the whole data set is 29,214. We set the number of agent $N=90$, the first ten agents are assigned with data from two writers, and the rest of the agents are assigned with data form one writer. Therefore, the data distribution is neither i.i.d. nor balanced. We use the neural network given in~\cite{caldas2018leaf} as the training model, which consists of 2 convolutional layers and two fully connected layers. The output layer has 62 neurons that matches the number of classes in the FEMNIST data set.

\begin{figure}[tb!]
	\centering
	\begin{subfigure}[t]{0.48\linewidth}
		\centering
		\includegraphics[width=\linewidth]{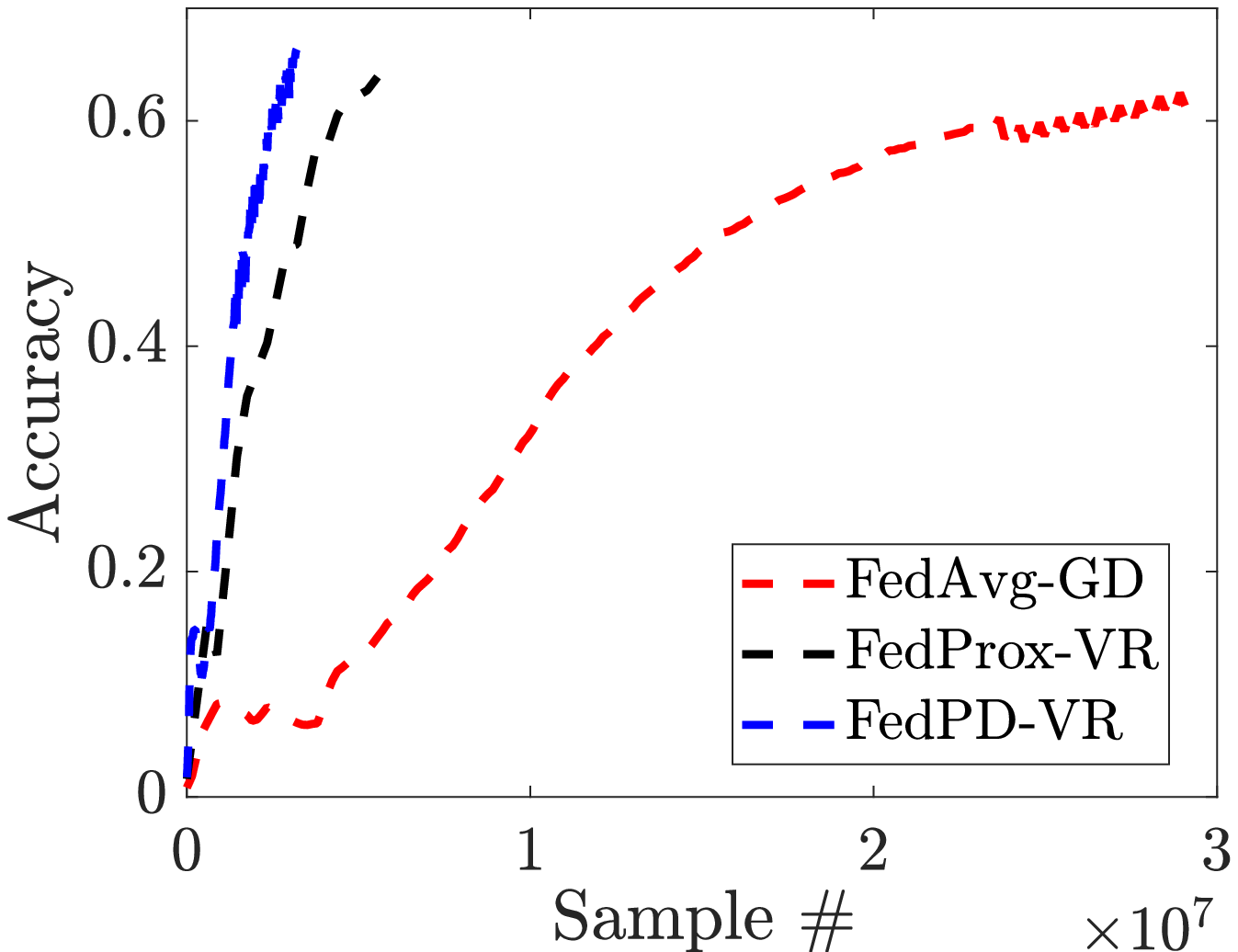}
		\caption{The testing accuracy of FedAvg-GD, FedProx-VR and FedPD-VR with respect to the number of samples.}
	\end{subfigure}
	\hfill
	\begin{subfigure}[t]{0.48\linewidth}
		\centering
		\includegraphics[width=\linewidth]{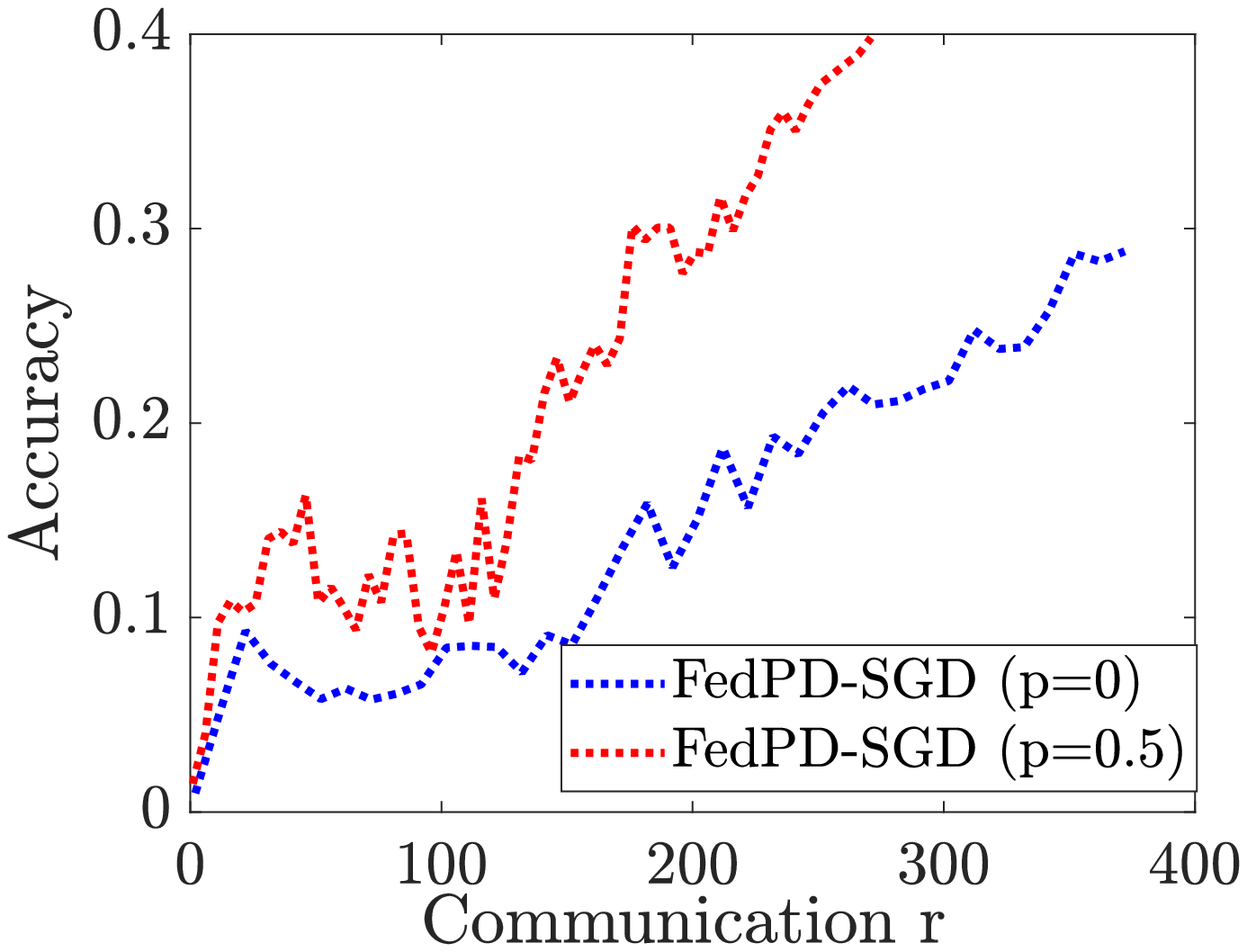}
		\caption{The testing accuracy of FedPD-SGD with $R=1$ and $R=2$ with respect to the number of communications.}
	\end{subfigure}
	\caption{The convergence result of the algorithms on training neural network for handwriting character classification.}
	\label{fig:exp_NN}
\end{figure}

The numerical results shown in Fig.~\ref{fig:exp_NN} in the main text were generated by running  MATLAB codes on Amazon Web Services (AWS), with Intel Xeon E5-2686 v4 CPUs. In the training phase, we train the CNN model with FedAvg, FedProx and FedPD. In Fig.~\ref{fig:exp_NN}(a), for FedAvg, we use gradient descent for $Q=8$ local update steps between each communication rounds; to solve the local problem for FedProx, we use SARAH with $Q=20$ local steps; we use FedPD with Oracle II, computing full gradient every $I=20$ communication rounds and perform $Q=2$ local steps between two communication rounds. The hyper-parameters we use for FedAvg is $\eta = 0.005$; for FedProx we use $\rho=1$ and stepsize $\eta = 0.01$; for FedPD we use $\eta = 100$ and $\gamma = 400$. In Fig.~\ref{fig:exp_NN}(b), we use FedPD with Oracle I, with $Q=20$, $\eta = 100$ and $\gamma = 400$ and the mini-batch size $2$. We set the communication saving to $p=0$ and $p=0.5$.

\begin{figure}[bt!]
	\centering
	\begin{subfigure}[t]{0.40\linewidth}
		\centering
		\includegraphics[width=\linewidth]{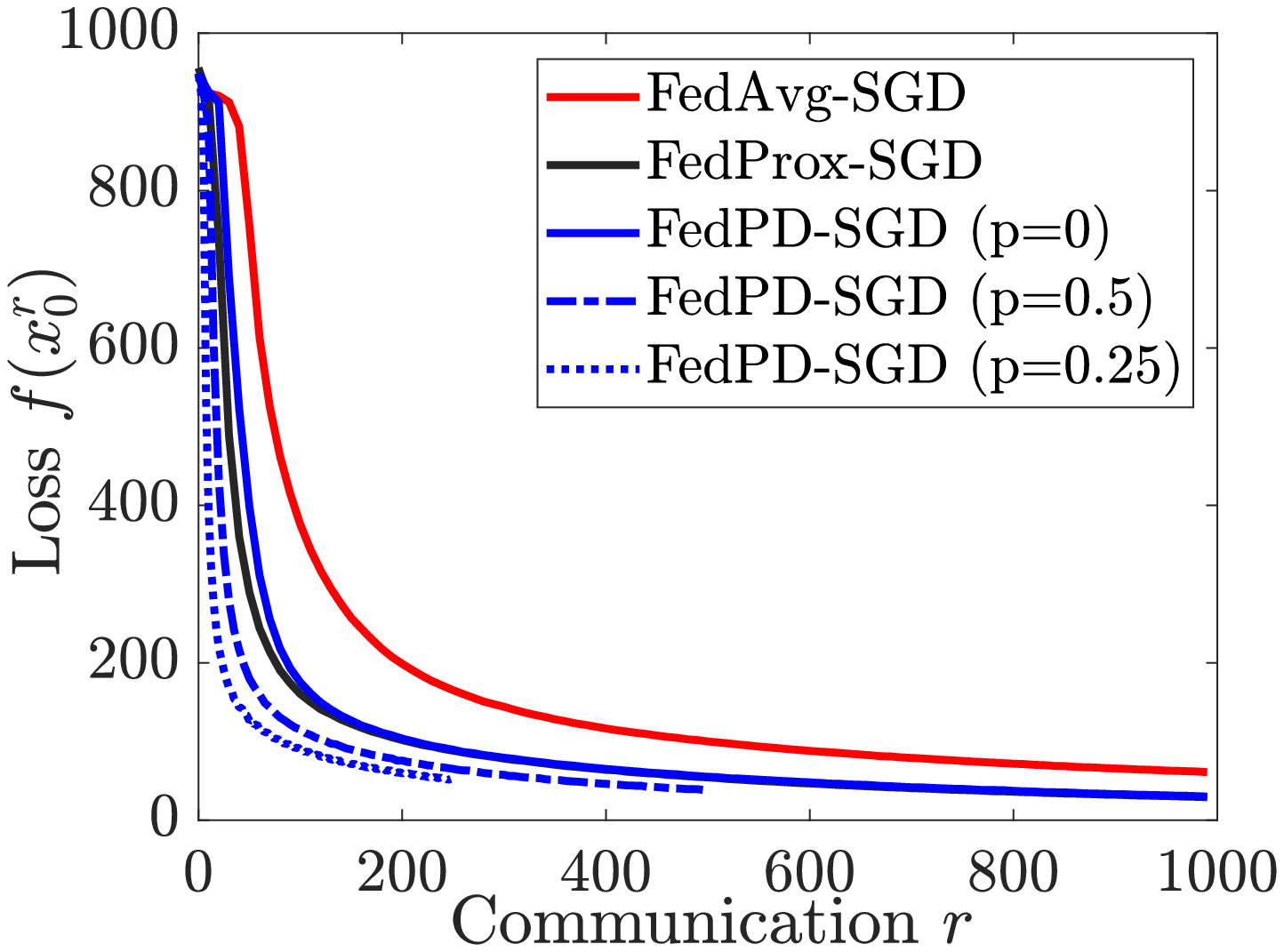}
		\caption{The loss value of FedAvg-SGD, FedProx-SGD and FedPD-SGD with respect to the number of communication rounds.}
	\end{subfigure}
	\hfill
	\begin{subfigure}[t]{0.40\linewidth}
		\centering
		\includegraphics[width=\linewidth]{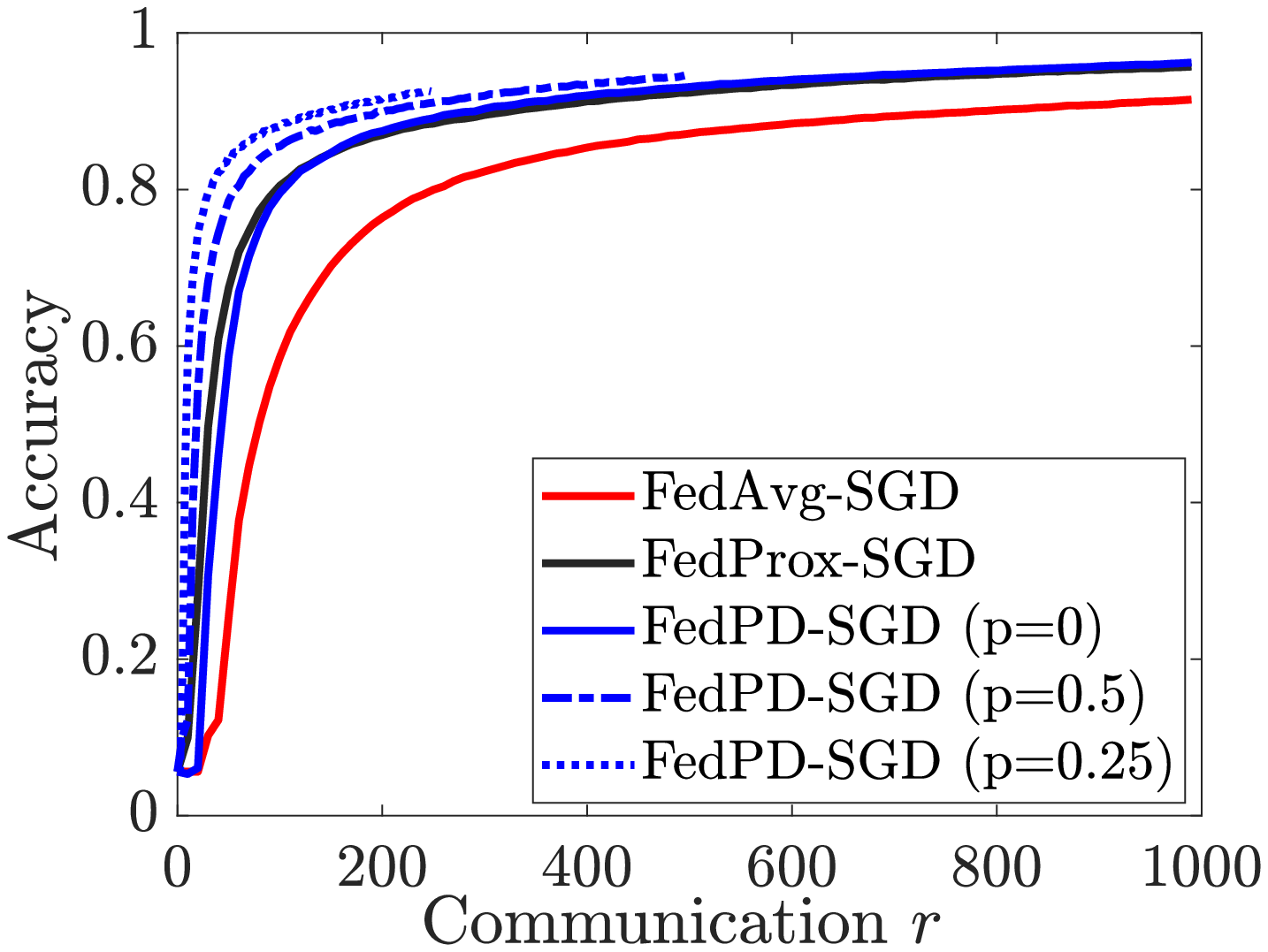}
		\caption{The training accuracy of of FedAvg-SGD, FedProx-SGD and FedPD-SGD with respect to the number of communication rounds.}
	\end{subfigure}
	\caption{The convergence results of the algorithms on training neural networks on the federated handwritten characters classification problem.}
	\label{fig:exp_NN_app}
	\vspace{0.3cm}
\end{figure}

The results shown in Fig.~\ref{fig:exp_NN_app} were generated by running Python codes (using the the PyTorch package~\footnote{PyTorch: An Imperative Style, High-Performance Deep Learning Library, \url{https://pytorch.org/}}) with AMD EPYC 7702 CPUs and an NVIDIA V100 GPU. 

\begin{figure}[bt!]
	\centering
	\begin{subfigure}[t]{0.40\linewidth}
		\centering
		\includegraphics[width=\linewidth]{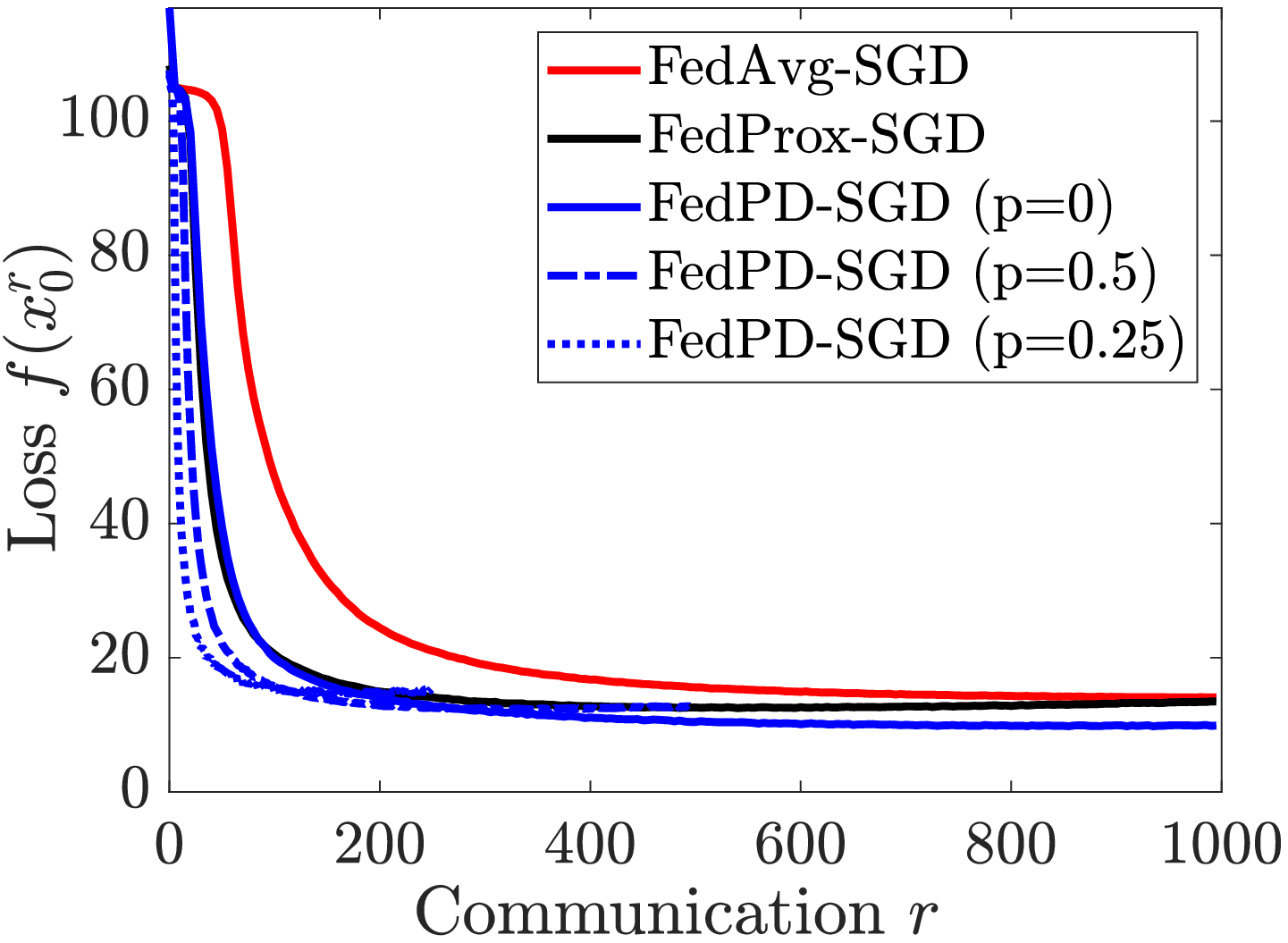}
		\caption{The testing loss value of FedAvg-SGD, FedProx-SGD and FedPD-SGD with respect to the number of communication rounds.}
	\end{subfigure}
	\hfill
	\begin{subfigure}[t]{0.40\linewidth}
		\centering
		\includegraphics[width=\linewidth]{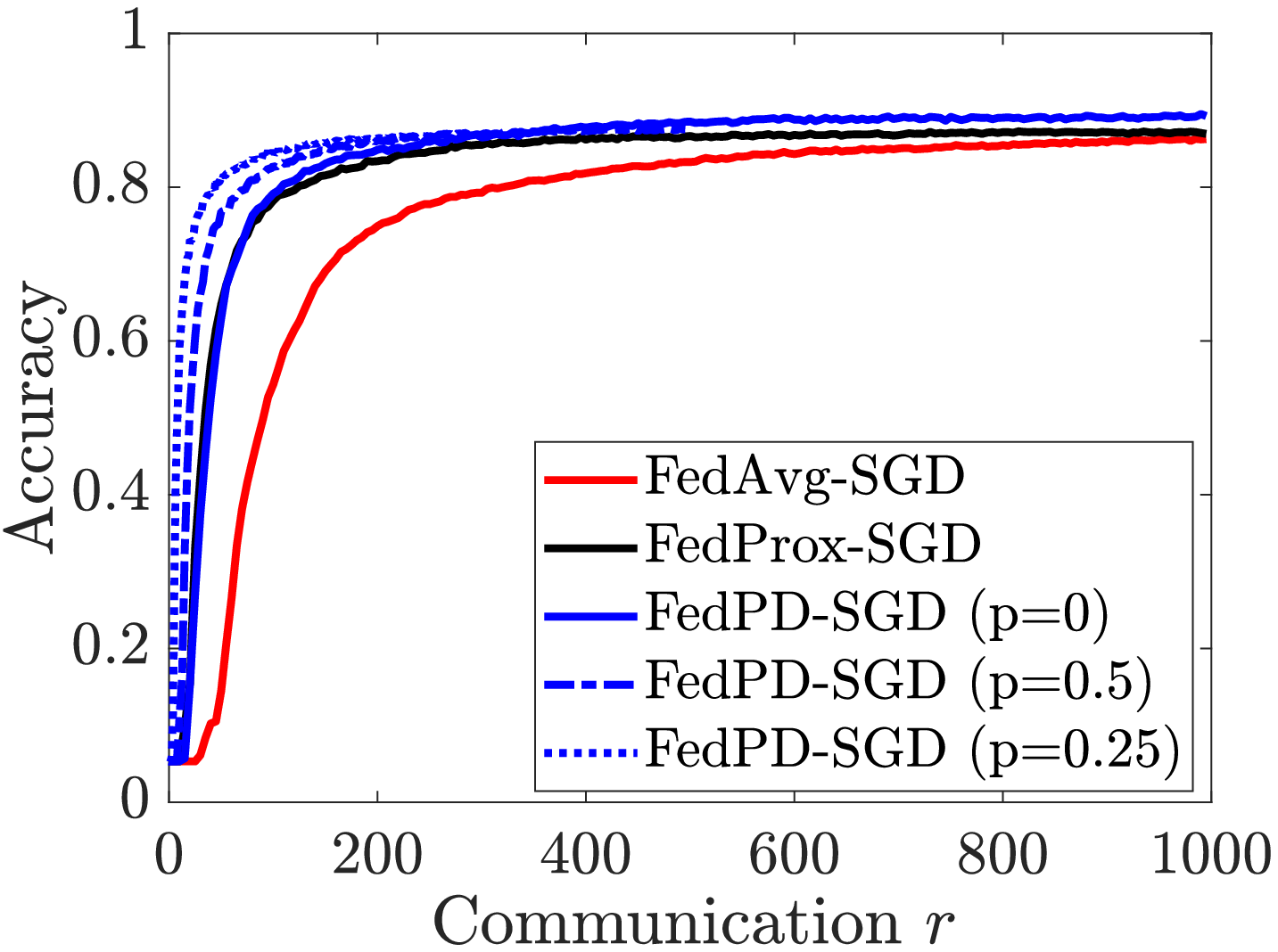}
		\caption{The testing accuracy of of FedAvg-SGD, FedProx-SGD and FedPD-SGD with respect to the number of communication rounds.}
	\end{subfigure}
	\caption{The convergence results of the algorithms on training neural networks on the federated handwritten characters classification problem with test data set.}
	\label{fig:exp_NN_app_test}
	\vspace{0.3cm}
\end{figure}

In the training phase, we train with FedProx, FedAvg and FedPD with a total $T=1000$ outer iterations. The local problems are solved with SGD for $Q=300$ local iterations and the mini-batch size in evaluating the stochastic gradient is $2$. The stepsize choice for FedAvg, FedProx and FedPD are $0.001$, $0.01$ and $0.01$, the hyper-parameter of FedProx is $\rho=1$ and for FedPD $\eta = 1$. In the experiment, we set the communication saving for FedPD to be $p=0$, $p=0.5$ and $p=0.25$. Note that we also tested FedAvg with larger stepsize $0.01$, but the algorithm becomes unstable, and its performance degrages significantly. As shown in Fig.~\ref{fig:exp_NN_app} and \ref{fig:exp_NN_app_test}, {FedAvg is slower than FedPD and FedProx, while FedProx has similar performance as FedPD when $R=1$.}
Further, we can see that as the frequency of communication of FedPD decreases, the final accuracy decreases and the final loss increases. However, the drop of accuracy is not significant, so FedPD is able to achieve a better performance with the same number of communication rounds. 

\vspace{-0.4cm}


\end{document}